\documentclass{article}

\usepackage[english]{babel}

\usepackage[letterpaper,top=2cm,bottom=2cm,left=3cm,right=3cm,marginparwidth=1.75cm]{geometry}
\usepackage{amsmath,amssymb,amsthm}
\usepackage{bbm}
\usepackage{graphicx}
\usepackage{booktabs}
\usepackage{xcolor}
\usepackage{algorithm}
\usepackage[noend]{algpseudocode}
\usepackage{tikz}
\usetikzlibrary{decorations.pathreplacing, calc}
\usepackage{natbib}
\usepackage[colorlinks=true, allcolors=blue]{hyperref}
\usepackage{authblk}

\newcommand{\bw}{\boldsymbol{w}}
\newcommand{\bx}{\boldsymbol{x}}
\newcommand{\bk}{\boldsymbol{k}}
\newcommand{\be}{\boldsymbol{e}}

\newcommand{\sD}{\mathcal{D}}

\newcommand{\sP}{\mathcal{P}}
\newcommand{\sF}{\mathcal{F}}
\newcommand{\sG}{\mathcal{G}}
\newcommand{\sN}{\mathcal{N}}
\newcommand{\sO}{\mathcal{O}}

\newcommand{\R}{\mathbb{R}}
\newcommand{\simplex}{\mathcal{W}}

\let\originalleft\left
\let\originalright\right
\renewcommand{\left}{\mathopen{}\mathclose\bgroup\originalleft}
\renewcommand{\right}{\aftergroup\egroup\originalright}

\newcommand{\abs}[1]{\left|#1\right|}
\newcommand{\inner}[2]{\left\langle#1, #2\right\rangle}

\DeclareMathOperator{\Wealth}{Wealth}
\DeclareMathOperator{\Dirichlet}{Dirichlet}
\DeclareMathOperator{\Beta}{Beta}
\DeclareMathOperator{\KL}{KL}
\DeclareMathOperator{\Var}{Var}
\DeclareMathOperator{\sign}{sign}
\newcommand{\Prob}{\mathbb{P}}
\newcommand{\E}{\mathbb{E}}
\newcommand{\ind}{\mathbbm{1}}
\newcommand{\iid}{\overset{\text{i.i.d.}}{\sim}}

\theoremstyle{plain}
\newtheorem{theorem}{Theorem}[section]
\newtheorem{proposition}[theorem]{Proposition}
\newtheorem{lemma}[theorem]{Lemma}
\newtheorem{corollary}[theorem]{Corollary}
\theoremstyle{definition}
\newtheorem{definition}[theorem]{Definition}

\theoremstyle{remark}

\makeatletter
\AtBeginDocument{%
  \let\orig@caption\caption
  \renewcommand{\caption}{\ifhmode\unskip\fi\orig@caption}%
}
\makeatother

\title{Monitoring for Shifts at Unknown Times: Parameter-Free Conditional Conformal Martingales}

\author[1]{Tuo Liu\thanks{\texttt{tuo.liu@kaust.edu.sa}}}
\author[2]{Drew Prinster\thanks{\texttt{drew@cs.jhu.edu}}}
\author[1]{Francesco Orabona\thanks{\texttt{francesco@orabona.com}}}

\affil[1]{King Abdullah University of Science and Technology, Saudi Arabia}
\affil[2]{Johns Hopkins University, USA}

\begin{document}
\maketitle

\begin{abstract}
  Conditional conformal test martingales provide sequential tests for detecting distribution shifts, using a fixed reference dataset to avoid contamination from test-time samples.
  Prior work used a penalized betting function via online optimization with smoothing, clipping, and other hyperparameters, which can adapt slowly when a shift occurs after a long stationary period.
  In this work, we introduce an approach that requires no hyperparameters and provably adapts quickly to shifts occurring at unknown times.
  Our key idea is to frame the penalized betting problem as portfolio selection in a three-asset market.
  Then, we use a restarting universal portfolio, with a heavy-tailed wealth allocation over time.
  We establish anytime false-alarm control and derive optimal finite-sample bounds on detection delay under detectable alternatives.
  We also develop an FFT-based implementation that computes the portfolio sequence over $T$ rounds in $\mathcal{O}(T^2\ln T)$ time, instead of $\mathcal{O}(T^3)$ of the naive implementation.
  Synthetic and real-world experiments suggest that the proposed approach adapts rapidly after long periods of null samples, with detection delays depending substantially less on shift time than prior baselines.
\end{abstract}

\section{Introduction}
\label{sec:intro}

A deployed machine learning model is typically only reliable when the data distribution it was trained on is consistent with the distribution it currently sees.
In reality, this consistency can break in many ways: a hospital replaces an imaging scanner, a camera lens degrades, or the questions a user asks a large language model evolve over time.
Labels usually arrive late or not at all, so the resulting drop in accuracy can go unnoticed for a long time.
Monitoring the inputs for distribution shift gives an early warning, which can trigger model retraining, additional data collection, or human review.
A common approach is to monitor a scalar score of each incoming input, such as the confidence of a classifier~\citep{hendrycks2017baseline} or the likelihood under a density model~\citep{ren2019likelihood}, and to raise an alarm once the scores no longer resemble those of in-distribution data.
In particular, a useful monitor should meet at least two key requirements.
First, false alarms must stay minimal no matter how long the monitor runs, so the probability of rejecting a true null should be \emph{anytime valid}.
Second, shifts rarely happen at the start of deployment; a model usually operates normally for a long and unknown period before, say, a sensor degrades or incoming images become corrupted. So, a good monitor should detect a shift \emph{quickly}, \emph{whenever} it happens.

The framework of testing by betting~\citep{shafer2021testing,ramdas2023game} allows one to design algorithms that meet the first requirement by construction.
In this framework, a gambler starts with unit wealth and repeatedly places bets that are fair under the null hypothesis of no shift, and statistical evidence for distribution shift is accumulated by the wealth process.
As no bet is favorable under the null, the wealth is a nonnegative supermartingale, and by Ville's inequality~\citep{ville1939etude} it exceeds level $1/\alpha$ \emph{at any time} with probability at most $\alpha$.
Conformal test martingales (CTMs)~\citep{vovk2003testing,vovk2021testing} apply this idea to monitoring for distribution shift by betting on conformal $p$-values, which rank each new score among all observations seen so far.
Because the ranking uses test data, however, new scores collected after a shift gradually enter the reference pool; post-shift observations then look typical, weakening the evidence against the null, an effect known as test-time contamination~\citep{podkopaev2022tracking,shaer2026testing}.
To tackle this issue, \citet{shaer2026testing} proposed \emph{conditional} CTMs (CCTMs), which instead compare every test score with a \emph{fixed} reference dataset $\sD_0$ and add a penalty that absorbs the estimation error of the reference distribution.

Crucially, the additional penalty term, $\epsilon |\eta|$, for a bet size of $\eta$ is not differentiable at $\eta = 0$, and \citet{shaer2026testing} learn the bet with the online Newton step (ONS)~\citep{hazan2007logarithmic} as proposed by \citet{cutkosky2018black}.
The resulting detector carries several hyperparameters: a smoothing constant to smooth $|\eta|$, a bound on the bet size, a clipping threshold that sets small bets to zero, and a warm-up period, with different clipping thresholds and warm-up lengths used for synthetic and image data. A good value will depend on how long the null period lasts, which is exactly what is \emph{unknown}.
More fundamentally, ONS is designed to compete with the best \emph{constant} bet in hindsight, which is a poor target when the shift comes late.
Before the shift, the optimal bet is precisely zero, since systematic wealth growth is impossible under the null, and every nonzero constant bet will cause wealth decay at a rate proportional to its size.
Therefore, every nonzero constant bet needs a number of post-shift observations that grows \emph{linearly} with the length of the null period to recover its expected wealth loss, so competing with the best constant bet guarantees little for late shifts.
Empirically, we measured that the median delay of the CCTM grows roughly 9 times for a shift after $t_0 = 4000$ null samples (Section~\ref{sec:experiments}).

In this paper, we address both problems: hyperparameter tuning and the slow response to late shifts.
These are our contributions:
\begin{itemize}
    \item \textbf{An equivalent reformulation of conditional CTMs.} Our key observation is that bet selection for the penalized betting function of~\citet{shaer2026testing} is equivalent to investing in a three-asset market (Section~\ref{sec:market}). So, the nonsmooth penalty becomes a linear function of the portfolio weights.
    The test martingale keeps the original betting function and is bounded below by the market wealth, so the reformulation needs no smoothing and costs no power.
    \item \textbf{A parameter-free detector.}
    With the new formulation, betting strategies can be built with the tools from online portfolio selection, which offers algorithms with wealth guarantees on \emph{every} sequence of returns~\citep{cover1991universal,OrabonaJ21,orabona2019modern}.
    To adapt to an unknown shift time, we propose a restarting universal portfolio with a heavy-tailed allocation of the initial wealth among the start times (Section~\ref{sec:restarting}).
    Each portfolio thus holds cash until its start time, so the one started right after the shift loses no wealth during the null period and then competes with the best constant-rebalanced portfolio (CRP) on the post-shift data alone.
    Apart from the test level and the confidence level of the reference band, it has no hyperparameters.
    \item \textbf{Adaptivity to an unknown shift time.}
    In Section~\ref{sec:guarantees}, we show that false alarms are controlled anytime. For a detectable shift after $t_0$ null observations, the cost of not knowing $t_0$ \emph{a priori} is $\ln t_0 + \sO(\ln \ln t_0)$, measured in log-wealth. So, the delay grows only logarithmically with $t_0$, compared with the linear growth for constant bets.
    We also prove that $\ln t_0$ is unavoidable: under anytime false-alarm control, no procedure detects shifts with a delay of smaller order than $\ln t_0$ uniformly over all $t_0$, even if the pre- and post-shift distributions are known.
    \item \textbf{Efficient implementation.}
    Despite this adaptivity, restarting a universal portfolio at every round costs $\sO(T^3)$ over a horizon of $T$ when computed directly.
    We give a divide-and-conquer FFT-based algorithm that computes \emph{exactly} the portfolio sequence in $\sO(T^2 \ln T)$ time (Section~\ref{sec:implementation}).
\end{itemize}
On synthetic and ImageNet-C data, our detector's delay is nearly flat in shift time, while those of conditional/standard CTMs grow steadily; after 4000--8000 null samples, we detect shifts two to five times faster.

\paragraph{Abbreviated Related Work.}
Sequential change detection has a long history, starting with the CUSUM procedure~\citep{page1954continuous}.
However, classical theory ~\citep{lorden1971procedures,lai1998information,tartakovsky2014sequential} generally assumes simple and parametric settings with known pre- and post-change distributions, and its guarantees generally control the average run length (ARL) to a false alarm under the null, which bounds how soon a false alarm is expected but allows one to occur eventually with probability one.
E-detectors~\citep{shin2022detectors} extend CUSUM-type and Shiryaev--Roberts (SR)-type procedures to certain composite and nonparametric settings by starting a new e-process (a generalization of a test supermartingale) at every round and combining them, with a guarantee on the ARL to false alarm.
For ARL control under the null of exchangeability, \citet{vovk2021testing} similarly studies CUSUM- and SR-type standard CTMs.
Our detector also starts a new bettor at every round, but it combines them with weights that sum to one, so that the total wealth itself is a valid test supermartingale.
This yields a stronger, unbounded-horizon guarantee: given a reference dataset whose confidence band holds, the probability of raising a false alarm is at most $\alpha$ \emph{anytime}. Under this criterion, a delay growing logarithmically with the shift time is unavoidable (Section~\ref{sec:guarantees}).
For additional related work on sequential change detection, see Appendix \ref{app:extended_related_work}.

Universal portfolios~\citep{cover1991universal,cover1996universal} achieve logarithmic regret against the best constant-rebalanced portfolio without tuning, and coin betting connects portfolio selection with parameter-free online learning~\citep{orabona2016coin,orabona2019modern}.
To compete with comparators that change over time, one can mix over sequences of experts~\citep{herbster1998tracking,singer1997switching,kozat2008universal,hazan2009efficient,daniely2015strongly,jun2017improved}.
The use of a universal portfolio as a testing-by-betting algorithm appeared for the first time in \citet{OrabonaJ21}, for the specific case of confidence sequences.

\section{Penalized Betting as Portfolio Selection}
\label{sec:market}

This section sets up the testing problem, reviews the conditional CTM (CCTM) of~\citet{shaer2026testing}, and recasts choosing the bets for the CCTM's penalized betting function as portfolio selection in a three-asset market.
Detailed proofs are given in Appendix~\ref{app:market}.

\subsection{Problem Setup}
\label{sec:setup}

Let $\sD_0 = \{X_1^0, \ldots, X_n^0\}$ be a reference dataset drawn i.i.d.\ from an unknown distribution $P$ on $\R$ with a continuous cumulative distribution function (CDF) $F$, and let $X_1, X_2, \ldots$ be test observations, independent of $\sD_0$, that arrive one at a time.
In applications, each observation is a score computed from a possibly high-dimensional input by a fixed function (e.g., entropy of a pretrained classifier), so a change in the score distribution signals a shift in the inputs.
Given $\sD_0$, we test the null hypothesis
\begin{equation}
    \label{eq:null}
    H_0: X_t \iid P \text{ for all } t \geq 1~.
\end{equation}
The alternative hypothesis of interest is a shift to a distribution $Q \neq P$ at an unknown time $t_0 \geq 0$, i.e., the scores are independent with
\begin{equation}
    \label{eq:change-point}
    H_1: X_t \sim P \text{ for } t \leq t_0 \text{ and } X_t \sim Q \text{ for } t > t_0,
\end{equation}
so that $t_0$ is the last round in which $H_0$ holds.
We write $\Prob_{H_0}$ for the probability law under $H_0$, and $\Prob_{H_1}$ for the law under $H_1$ with a given $Q$ and $t_0$.
Let $\sF_t := \sigma(\sD_0, X_1, \ldots, X_t)$ denote the natural filtration, with $\sF_0 := \sigma(\sD_0)$.
A sequence $(\eta_t)_{t \geq 1}$ is \emph{predictable} if each $\eta_t$ is $\sF_{t-1}$-measurable.
A sequential test is a stopping time $\tau$ with respect to $(\sF_t)_{t \geq 0}$: it raises a false alarm under $H_0$ if $\tau < \infty$, and $\tau = \infty$ means that no alarm is raised.
Our goals are (a) \emph{anytime false-alarm control}, i.e., $\Prob_{H_0}\{\tau < \infty \mid \sD_0\} \leq \alpha$ for a given level $\alpha \in (0,1)$, and (b) under~\eqref{eq:change-point}, a short detection delay $(\tau - t_0)_+$.

\paragraph{Testing By Betting.}
Testing by betting~\citep{shafer2021testing,ramdas2023game} measures evidence against $H_0$ by the wealth of a gambler who bets against it.
The gambler starts with the wealth $M_0 = 1$ and, before seeing $X_t$, chooses an $\sF_{t-1}$-measurable betting function $b_t : \R \to [0, \infty)$, so that
\[
    M_t = M_{t-1} b_t\left(X_t\right) \quad \text{and} \quad \E_{H_0}\left[b_t\left(X_t\right) \mid \sF_{t-1}\right] \leq 1~.
\]
The constraint says that no bet is favorable under the null, and it makes the wealth $(M_t)_{t \geq 0}$ a nonnegative supermartingale under $H_0$:
\[
    \E_{H_0}\left[M_t \mid \sF_{t-1}\right] = M_{t-1} \E_{H_0}\left[b_t\left(X_t\right) \mid \sF_{t-1}\right] \leq M_{t-1}~.
\]
Ville's inequality~\citep{ville1939etude} then bounds the probability that the wealth ever reaches the level $1/\alpha$:
\[
    \Prob_{H_0}\{\exists t \geq 1 : M_t \geq 1/\alpha\} \leq \alpha~.
\]
Hence the test $\tau := \inf\ \{t \geq 1 : M_t \geq 1/\alpha\}$ meets goal (a), and goal (b) asks the gambler to grow the wealth quickly once the shift has happened.

If the true CDF $F$ were known, the values $F(X_t)$ would be i.i.d.\ uniform on $[0,1]$ under $H_0$, and the factor $1 + \eta_t (p_t - 1/2)$ with $p_t := F(X_t)$ and a predictable $\eta_t \in [-2,2]$ would be a fair bet on whether, and by how much, $F(X_t)$ deviates from its mean $1/2$ under the null.
In practice, we replace $F$ by the accessible empirical CDF $\hat F_0(x) := n^{-1} \sum_{i=1}^n \ind\{X_i^0 \leq x\}$ of the reference sample.
Given $\sD_0$, the resulting $p$-values $\hat p_t := \hat F_0(X_t)$ may no longer be i.i.d.\ uniformly distributed under $H_0$; their mean can differ from $1/2$~\citep{bates2023testing}.
Betting on the sign of this bias grows the wealth even under the null and inflates false alarms~\citep{shaer2026testing}.

\paragraph{The Conditional CTM.}
\citet{shaer2026testing} compensate for this error by bounding the finite-sample approximation error of $\hat F_0$ to $F$ with a confidence band, and penalizing the bet by a term proportional to the bet size as well as the band width.
Specifically, fix a confidence level $\delta \in (0,1)$ and let $\epsilon : [0,1] \to [0,1]$ be a band width, computed from $\sD_0$, such that the event
\begin{equation}
    \label{eq:band}
    \abs{\hat{F}_0\left(x\right) - F\left(x\right)} \leq \epsilon\left(\hat{F}_0\left(x\right)\right) \text{ for all } x \in \R
\end{equation}
has probability at least $1 - \delta$ over the draw of $\sD_0$.
We write $\sG$ for the set of reference datasets on which~\eqref{eq:band} holds.
For example, the Dvoretzky--Kiefer--Wolfowitz (DKW) inequality~\citep{dvoretzky1956asymptotic} gives the constant band $\epsilon \equiv \min\ \{1, \sqrt{\ln(2/\delta)/(2n)}\}$, and likelihood-based bands give widths that vary with $\hat{F}_0(x)$~\citep{owen1995nonparametric}.
With the scale $\kappa := 1 / (1/2 + \max_{p \in [0,1]}\ \epsilon(p)) \in [2/3, 2]$, the penalized betting function of~\citet{shaer2026testing} is
\begin{equation}
    \label{eq:bet}
    b\left(\eta, p\right)
    := 1 + \kappa (\eta (p - 1/2) - \abs{\eta} \epsilon\left(p\right)),
\end{equation}
for a bet $\eta \in [-1, 1]$ and $p \in [0,1]$.
The first part is the ideal bet, with $\hat{p}_t$ in place of $p_t$.
The second part works like a \emph{tax}: a bet of size $|\eta|$ pays $\kappa |\eta| \epsilon(\hat{p}_t)$, regardless of the outcome.
A bet therefore increases the gambler's wealth only when $\hat{p}_t$ deviates from $1/2$ by more than the band width, in the direction of the bet.
Since $|\eta(p - 1/2) - |\eta| \epsilon(p)| \leq 1/2 + \max_{p \in [0,1]}\ \epsilon(p)$, the normalization by $\kappa$ ensures $b(\eta, p) \in [0, 2]$.
For a predictable sequence of bets, the resulting wealth is a valid test supermartingale given any $\sD_0 \in \sG$; the level $\alpha$ then controls false alarms given such a $\sD_0$, while $\delta$ bounds the probability that $\sD_0 \notin \sG$.

\begin{proposition}[\citealp{shaer2026testing}]
    \label{prop:validity}
    Let $(\eta_t)_{t \geq 1}$ be predictable with values in $[-1, 1]$, and let $M_t := \prod_{i=1}^{t} b(\eta_i, \hat{p}_i)$ with $M_0 := 1$.
    If $\sD_0 \in \sG$, then under $H_0$, $(M_t)_{t \geq 0}$ is a nonnegative $(\sF_t)_{t \geq 0}$-supermartingale, and
    \[
        \Prob_{H_0}\left\{\exists t \geq 1 : M_t \geq 1/\alpha \ \middle|\ \sD_0\right\} \leq \alpha~.
    \]
    Consequently, $\Prob_{H_0}\{\exists t \geq 1 : M_t \geq 1/\alpha\} \leq \alpha + (1 - \alpha)\delta$.
\end{proposition}

\paragraph{Learning The Bets Online.}
Proposition~\ref{prop:validity} holds for every predictable sequence of bets $(\eta_t)_{t \geq 1}$, and \emph{statistical power} demands choosing one that grows the wealth quickly after a shift.
Maximizing $\ln M_T = \sum_{t=1}^{T} \ln b(\eta_t, \hat{p}_t)$ over $T$ rounds is an online learning problem with losses $-\ln b(\eta, \hat{p}_t)$.
These losses are convex in $\eta$, but not differentiable at $\eta = 0$ when $\epsilon(\hat{p}_t) > 0$.

\paragraph{Issues With ONS.}
Following \citet{cutkosky2018black}, \citet{shaer2026testing} use ONS to learn the bets, using a surrogate in which $|\eta|$ is replaced by the smooth approximation $\sqrt{\eta^2 + k^2}$ for a small $k > 0$, and restrict the bets to $[-1/2, 1/2]$.
Before the shift, the bets fluctuate around zero; every nonzero bet pays tax, so the wealth decays and must be won back after the shift, a problem already observed for CTMs~\citep{volkhonskiy2017inductive}.
To prevent this, \citet{shaer2026testing} set $\eta$ to zero whenever its magnitude falls below a threshold $C$, and no bets are placed during an initial warm-up period (after smoothing, even a zero bet pays a tax proportional to $k \epsilon(\hat{p}_t)$).
Notably, the threshold trades off wealth loss before the shift against a slower reaction after it.
The smoothing constant $k$, the range of bets, the threshold $C$, and the warm-up length must all be set by the user.

\subsection{A Three-Asset Market}
\label{sec:three-asset}

Here, we show how to remove the non-differentiability by an equivalent reparameterization of~\citet{shaer2026testing}'s original construction. The reframing turns the choice of bets into portfolio selection, a problem for which parameter-free algorithms are available.

\paragraph{Intuition.}
Split a bet into its positive and negative parts, $\eta = \eta^+ - \eta^-$ with $\eta^+ := \max\ \{\eta, 0\}$ and $\eta^- := \max\ \{-\eta, 0\}$, so that $|\eta| = \eta^+ + \eta^-$.
Substituting both identities into~\eqref{eq:bet} gives
\[
    b\left(\eta, p\right) = \left(1 - \abs{\eta}\right) + \eta^+ b\left(1, p\right) + \eta^- b\left(-1, p\right)~.
\]
The penalized bet is thus the return of a portfolio that keeps a fraction $1 - |\eta|$ of the wealth in cash and invests the fractions $\eta^+$ and $\eta^-$ in two assets whose returns are the two most aggressive bets, $b(1, p)$ and $b(-1, p)$.
The three fractions are all nonnegative and sum to one.
This suggests treating the three fixed bets as assets.

\begin{definition}[The three-asset market]
    \label{def:market}
    In round $t$, the \emph{price relatives} of the three assets are
    \[
        \bx_t := \left(x_t^0, x_t^+, x_t^-\right) := \left(1, b\left(1, \hat{p}_t\right), b\left(-1, \hat{p}_t\right)\right) \in [0,2]^3~.
    \]
    A portfolio is a vector $\bw = (w^0, w^+, w^-)$ in the simplex $\simplex := \{\bw \in [0,1]^3 : w^0 + w^+ + w^- = 1\}$. Investing with $\bw$ in round $t$ multiplies the wealth by $\inner{\bw}{\bx_t}$.
\end{definition}

We call the three assets \emph{cash}, the \emph{up asset}, and the \emph{down asset}.
The up asset pays $x_t^+ = 1 + \kappa(\hat{p}_t - 1/2 - \epsilon(\hat{p}_t))$, so the wealth invested in it grows when $\hat{p}_t$ exceeds $1/2$ by more than the band width, which happens systematically when the scores shift upwards. The down asset behaves similarly when scores shift downwards.
Applying Proposition~\ref{prop:validity} to the constant bets $\pm 1$ shows that neither risky asset can grow in expectation under $H_0$ when $\sD_0 \in \sG$, so cash is the only safe asset.
Without estimation error, that is, with $\epsilon \equiv 0$, the market reduces to the standard two-asset form of betting on a coin~\citep{orabona2016coin,orabona2019modern}.
Expanding the inner product shows how a portfolio bets:
\begin{equation}
    \label{eq:portfolio-return}
    \inner{\bw}{\bx_t} = 1 + \kappa \left(w^+ - w^-\right) \left(\hat{p}_t - 1/2\right) - \kappa \left(w^+ + w^-\right) \epsilon\left(\hat{p}_t\right),
\end{equation}
so a portfolio bets the difference $w^+ - w^-$ but pays the tax on the sum (or hedging) $w^+ + w^- \geq |w^+ - w^-|$.
The two coincide exactly when the portfolio holds at most one risky asset, which gives the following lemma.
\begin{lemma}
    \label{lem:equivalence}
    For every $\bw \in \simplex$ and every round $t$,
    \begin{equation}
        \label{eq:gap}
        \inner{\bw}{\bx_t} = b\left(w^+ - w^-, \hat{p}_t\right) - 2\kappa \min\left\{w^+, w^-\right\} \epsilon\left(\hat{p}_t\right)~.
    \end{equation}
    Conversely, for every $\eta \in [-1, 1]$, the portfolio $\bw(\eta) := (1 - |\eta|, \eta^+, \eta^-) \in \simplex$ satisfies $\inner{\bw(\eta)}{\bx_t} = b(\eta, \hat{p}_t)$ for every round $t$.
    Consequently, for all rounds $1 \leq s \leq t$ and every realization of $\hat{p}_s, \ldots, \hat{p}_t \in [0, 1]$,
    \[
        \max_{\bw \in \simplex}\ \prod_{i=s}^{t} \inner{\bw}{\bx_i} = \max_{\eta \in [-1, 1]}\ \prod_{i=s}^{t} b\left(\eta, \hat{p}_i\right)~.
    \]
\end{lemma}

In portfolio selection, the standard comparator is the best \emph{constant-rebalanced portfolio} (CRP) in hindsight, which keeps the same fraction $\bw$ of wealth in each asset at every round~\citep{cover1991universal}.
Lemma~\ref{lem:equivalence} shows that, in the three-asset market, the best CRP earns exactly as much as the best constant bet, the comparator of ONS in~\citet{shaer2026testing}.
More importantly, the market construction turns the choice of betting strategies into \emph{online portfolio selection} while keeping the comparator: any portfolio algorithm with a regret guarantee against CRPs thus competes with the best constant bet \emph{automatically}.
Since the return $\inner{\bw}{\bx_t}$ is linear in the allocation $\bw$, the portfolio view also allows mixes over strategies with different start times, which will be crucial for adapting to an unknown shift time in Section~\ref{sec:restarting}.
Finally, in contrast to the ONS solution, the full range of bets $[-1, 1]$ is available.

\paragraph{From Portfolio Wealth To Test Martingale.}
A portfolio algorithm chooses a predictable sequence of portfolios $(\bw_t)_{t \geq 1}$ in $\simplex$.
We will use it only to select the bet $\eta$ and keep the penalized betting function for the test~\eqref{eq:bet}.
With $\eta_t := w_t^+ - w_t^-$, this involves two wealth processes,
\begin{equation}
    \label{eq:martingale}
    M_t := \prod_{i=1}^{t} b\left(\eta_i, \hat{p}_i\right) \text{ and } \Wealth_t := \prod_{i=1}^{t} \inner{\bw_i}{\bx_i},
\end{equation}
with $\Wealth_0 = M_0 = 1$.
The two processes play different roles.
The portfolio algorithm sees only the market and controls $\Wealth_t$, while the alarm is raised by $M_t$.
Since $M_t \geq \Wealth_t$ (Proposition~\ref{prop:validity-2} below), any lower bound on $\ln \Wealth_t$, such as a regret bound against some comparator, also holds for $\ln M_t$.
All our guarantees on the detection delay are proved for $\Wealth_t$ and therefore apply to the test based on $M_t$.
In short, we choose the bets in the market and measure the evidence with the original betting function.

\begin{proposition}[Validity]
    \label{prop:validity-2}
    Let $(\bw_t)_{t \geq 1} \in \simplex$ be predictable with respect to $(\sF_t)$, define $(\eta_t)_{t \geq 1}$, $(M_t)_{t \geq 0}$, and $(\Wealth_t)_{t \geq 0}$ by~\eqref{eq:martingale}, and let $\tau := \inf\ \{t \geq 1 : M_t \geq 1/\alpha\}$.
    Then $M_t \geq \Wealth_t$ for all $t \geq 0$.
    If $\sD_0 \in \sG$, then $(M_t)_{t \geq 0}$ is a nonnegative supermartingale under $\Prob_{H_0}$ and $\Prob_{H_0}\{\tau < \infty \mid \sD_0\} \leq \alpha$.
    Without conditioning, $\Prob_{H_0}\{\tau < \infty\} \leq \alpha + (1 - \alpha)\delta$.
\end{proposition}

\section{Restarting Universal Portfolios}
\label{sec:restarting}

Lemma~\ref{lem:equivalence} holds for any sub-interval of all the rounds $\{1, 2, \ldots, T\}$, and any sequence of scalar bets $(\eta_t)_{t \geq 1}$ is realized exactly by the portfolios $(\bw(\eta_t))_{t \geq 1}$.
The correspondence is therefore not limited to constant strategies, and a comparator may change over time.
For monitoring purposes, after a long null phase, however, the best constant bet could be a weak comparator.
Intuitively, the right comparator should place no bets until the shift and then bet in the right direction, i.e., it holds $\bw_t = (1, 0, 0)$ for $t \leq t_0$ and $\bw_t = \bw(\eta)$ for $t > t_0$ with a fixed $\eta \in [-1, 1]$.
Its log-wealth at a round $t > t_0$ is $\sum_{i=t_0+1}^{t} \ln b(\eta, \hat{p}_i)$, which, unlike the log-wealth of a constant bet, suffers no decay during the null period.
In this section, we construct a parameter-free portfolio algorithm that competes with all such comparators at once, without knowing $t_0$.
We first explain why a single constant bet is a weak comparator for late shifts, then recall Cover's universal portfolio, and finally design a variant that restarts at every round with a heavy-tailed allocation of the initial wealth.
Detailed proofs are given in Appendix~\ref{app:restarting}.

\paragraph{Why Constant Bets Adapt Slowly.}
Under the null, no bet can grow the wealth in expectation by the supermartingale property, yet every nonzero constant bet keeps paying the tax, so its wealth decays throughout the null period and the best constant bet becomes a weak comparator for late shifts.
We formalize this argument in Appendix~\ref{app:constant}: tests with constant bets must first recover the wealth lost before the shift, so their expected detection delay grows \emph{linearly} in $t_0$.

% \subsection{Universal Portfolios}
% \label{sec:up}
\paragraph{Universal Portfolios.}
Let $\mu$ be the Jeffreys prior $\Dirichlet(1/2, 1/2, 1/2)$ on $\simplex$.
The universal portfolio~\citep{cover1991universal,cover1996universal} invests at round $t$ in the average of all constant portfolios, weighted by the wealth they would have earned so far,
\[
    \bw_t := \frac{\int_{\simplex} \bw \prod_{i=1}^{t-1} \inner{\bw}{\bx_i} \mu(d\bw)}{\int_{\simplex} \prod_{i=1}^{t-1} \inner{\bw}{\bx_i} \mu(d\bw)}~.
\]
By telescoping the wealth updates, the universal portfolio started at round $s$ has wealth
\[
    \int \prod_{i=s}^t \inner{\bw}{\bx_i} \mu(d\bw)
\]
by round $t$. This is the mixture of all CRPs' wealths with respect to $\mu$.
\citet[Theorem~2]{cover1996universal} showed that this mixture loses at most a logarithmic factor against the best CRP on \emph{every} sequence of price relatives; for three assets and all $\bx_s, \ldots, \bx_t \in [0, \infty)^3$ with $1 \leq s \leq t$,
\begin{equation}
    \label{eq:up-regret}
    \ln \int \prod_{i=s}^t \inner{\bw}{\bx_i} \mu(d\bw) \geq \max_{\bw \in \simplex}\ \ln \prod_{i=s}^t \inner{\bw}{\bx_i} - \ln(2 (t - s) + 3).
\end{equation}
By Lemma~\ref{lem:equivalence}, the universal portfolio started at round $t_0 + 1$ would thus compete with the ideal comparator described at the beginning of this section, at a regret of $\ln(2(t - t_0) + 1)$ in log-wealth. Unlike ONS, however, the universal portfolio has no step size, smoothing, restricted range of bets, or any other hyperparameter.
Since $t_0$ is unknown, we start one at every round.

\subsection{Restarting at Every Round}
\label{sec:restart}

Fix positive weights $(\pi_s)_{s \geq 1}$ with $\sum_{i=1}^{\infty} \pi_i = 1$, to be chosen later.
The restarting portfolio assigns the initial wealth $\pi_s$ to a portfolio sequence that starts at round $s$.
The sequence holds cash during the rounds $1, \ldots, s - 1$, and from round $s$ on it is invested in a fresh universal portfolio that has not seen the earlier data. Its wealth component is
\[
    \Wealth_t^{s} := \begin{cases}
        \pi_s \int_{\simplex} \prod_{i=s}^t \inner{\bw}{\bx_i} \mu(d\bw), & t \geq s, \\
        \pi_s , & t < s~.
    \end{cases}
\]
After time $t$, the restarting portfolio holds the mixture
\begin{equation}
    \label{eq:restarting-wealth}
    \Wealth_t := \sum_{i=1}^{\infty} \Wealth_t^{i} = \sum_{i=1}^{t} \Wealth_t^{i} + r_t,
\end{equation}
where the residual $r_t := \sum_{i=t+1}^{\infty} \pi_i$ is the part of the unit initial wealth of the portfolios that have not started yet.
Starting a new component does not add new wealth since all portfolios share the initial unit of wealth, and a new start only moves wealth out of the residual.
This is what keeps the detector valid as a supermartingale, in contrast to procedures that add a new unit of wealth at every round and therefore control only the ARL to false alarm~\citep{shin2022detectors}.
The next proposition writes~\eqref{eq:restarting-wealth} as the wealth of a predictable portfolio with the recursion for Algorithm~\ref{alg:test}.

\begin{proposition}
    \label{prop:restarting}
    Let $B_1 := \pi_1$ and $B_{t+1}(\bw) := \inner{\bw}{\bx_t} B_t(\bw) + \pi_{t+1}$ for $t \geq 1$, and let
    \begin{equation}
        \label{eq:restarting-allocation}
        \bw_t := \frac{\int_{\simplex} \bw B_t\left(\bw\right) \mu\left(d\bw\right) + r_t \left(1, 0, 0\right)}{\int_{\simplex} B_t\left(\bw\right) \mu\left(d\bw\right) + r_t}~.
    \end{equation}
    Then $\bw_t \in \simplex$ is $\sF_{t-1}$-measurable and $\prod_{i=1}^{t} \inner{\bw_i}{\bx_i}$ equals~\eqref{eq:restarting-wealth} for all $t \geq 0$.
\end{proposition}

Note that $B_t(\bw) = \sum_{i=1}^t \pi_i \prod_{j=i}^{t-1} \inner{\bw}{\bx_j}$ by unrolling the recursion, 
so $\bw_t$ is the weighted (by wealth) average of the allocations of all active universal portfolios, together with the cash residual (proof in Appendix~\ref{app:restarting}).
Algorithm~\ref{alg:test} summarizes the resulting test.
We emphasize that Algorithm~\ref{alg:test} has no tuning parameters: its only inputs are the levels $\alpha$ and $\delta$. Combining~\eqref{eq:up-regret} with Lemma~\ref{lem:equivalence} yields an adaptive guarantee.

\begin{lemma}[Restarting regret]
\label{lem:restarting-regret}
For every sequence $\hat p_1, \hat p_2, \dots \in [0,1]$ and all $1 \leq s \leq t$,
\[
    \ln M_t \geq \ln \Wealth_t \geq \max_{\eta \in [-1,1]}\ \sum_{i=s}^t \ln b(\eta, \hat p_i) - \ln\frac{1}{\pi_s} - \ln\left(2(t - s) + 3\right)~.
\]
\end{lemma}

\begin{algorithm}[t]
    \caption{Restarting universal portfolio test}
    \label{alg:test}
    \small
    \begin{algorithmic}[1]
        \Require reference dataset $\sD_0$, test level $\alpha$, confidence level $\delta$
        \State Compute $\hat{F}_0$, the band width $\epsilon$, and $\kappa$ from $\sD_0$ and $\delta$
        \State Set $M_0 \gets 1$ and $B_1 \gets \pi_1$
        \For{$t = 1, 2, \ldots$}
            \State Compute $\bw_t$ by~\eqref{eq:restarting-allocation}
            \State Observe $X_t$ and set $\hat{p}_t \gets \hat{F}_0(X_t)$
            \State Set $\eta_t \gets w_t^+ - w_t^-$ and $M_t \gets M_{t-1} b(\eta_t, \hat{p}_t)$
            \If{$M_t \geq 1/\alpha$}
                \State Reject $H_0$ and stop
            \EndIf
            \State Set $B_{t+1}(\bw) \gets \inner{\bw}{\bx_t} B_t(\bw) + \pi_{t+1}$
        \EndFor
    \end{algorithmic}
\end{algorithm}

\paragraph{A Heavy-Tailed Allocation.}
Lemma~\ref{lem:restarting-regret} shows that the portfolio started at round $s$ pays a restart cost of $\ln(1/\pi_s)$ in log-wealth, on top of the regret in~\eqref{eq:up-regret}.
The delay after a shift at $t_0$ therefore grows with $\ln(1/\pi_{t_0+1})$, and the allocation should decay as slowly as possible, while being summable to one.
We choose the following heavy-tailed allocation
\begin{equation}
    \label{eq:heavy-tailed}
    \pi_s := \ln 2 \left(\frac{1}{\ln\left(s + 1\right)} - \frac{1}{\ln\left(s + 2\right)}\right) \quad \text{for } s \geq 1,
\end{equation}
which behaves like $\sO(1/(s \ln^2 s))$.
Obviously, the sum telescopes, so $\sum_{i=1}^{\infty} \pi_i = 1$, and the residual also has the closed form $r_t = \ln 2 / \ln(t + 2)$.

\begin{corollary}
    \label{cor:heavy-tailed-regret}
    With the allocation~\eqref{eq:heavy-tailed}, for every sequence $\hat{p}_1, \hat{p}_2, \ldots \in [0,1]$ and all $1 \leq s \leq t$,
    \[
        \begin{aligned}
            \ln M_t \geq \ln \Wealth_t \geq {} & \max_{\eta \in [-1,1]}\ \sum_{i=s}^{t} \ln b\left(\eta, \hat{p}_i\right) - \ln\left(2\left(t - s\right) + 3\right) + \ln \ln 2 \\
            & - \ln\left(s + 2\right) - 2 \ln \ln\left(s + 2\right)~.
        \end{aligned}
    \]
\end{corollary}

For a shift after round $t_0$, taking $s = t_0 + 1$ shows that the restarting portfolio competes with the ideal comparator that places no bets until $t_0$ and then makes the best constant bet on the shifted data (described at the beginning of this section).
Unlike for constant bets, the null period enters only through the prior cost of $\ln t_0 + \sO(\ln\ln t_0)$.
Wealth decay under $H_0$, which \citet{shaer2026testing} manually handled by clipping, is thus automatically controlled by our construction.

\subsection{Efficient Implementation}
\label{sec:implementation}

The allocation~\eqref{eq:restarting-allocation} needs the integrals of $B_t$ and of $w^i B_t$, $i \in \{0, +, -\}$, under the Dirichlet prior in every round.
Since $B_t$ is a polynomial of degree $t - 1$ with $t(t+1)/2$ coefficients, and the moments of the prior have closed forms, a direct computation costs $\sO(t^2)$ operations in round $t$ and $\sO(T^3)$ over $T$ rounds.
Our key observation is that all portfolios started in a block of rounds grow by the \emph{same} factor after the block, so their contributions to many later rounds can be handled at once.
We bisect the horizon recursively, store the polynomials by their values on Chebyshev grids, after a change of variables that maps the simplex to the unit square, and evaluate the contributions of each block with fast cosine transforms.
This computes the same allocations, exactly in exact arithmetic without knowing the horizon, in $\sO(T^2 \ln T)$ time and $\sO(T^2)$ memory (Appendix~\ref{app:implementation}).
On one CPU core, it is faster than the direct computation from about $T = 2000$ rounds onwards, and $5.7$ times faster at $T = 12{,}000$.

\section{Statistical Guarantees}
\label{sec:guarantees}

We now show that the detector of Section~\ref{sec:restarting} controls false alarms at any time.
Let $\tau := \inf\ \{t \geq 1: M_t \geq 1/\alpha\}$ be the detection time (or alarm time) of Algorithm~\ref{alg:test}.
We provide a bound on $\tau$ under a change after an arbitrary round $t_0$.
A lower bound then shows that the logarithmic dependence cannot be improved by any detector.
Detailed proofs are given in Appendix~\ref{app:guarantees}.

\paragraph{False-Alarm Control.}
Type-I error control is guaranteed by the supermartingale property of the detector.
By Proposition~\ref{prop:restarting}, the restarting portfolio is predictable, so Proposition~\ref{prop:validity-2} applies directly.

\paragraph{Detection Delay.}
To turn Corollary~\ref{cor:heavy-tailed-regret} into a bound on the delay, the shift must be detectable by \emph{some} constant bet.
We measure the signal strength by
\begin{equation}
    \label{eq:signal}
     \Delta := \abs{\E_Q\left[\hat{p}_t - 1/2 \ \middle | \ \sD_0\right]} - \E_Q\left[\epsilon(\hat{p}_t) \ \middle | \ \sD_0\right],
\end{equation}
the expected gain of the best unit bet after the change.
It is the effective signal strength of~\citet{shaer2026testing} without smoothing, and $\Delta \leq 1/2$.
If $\Delta \leq 0$, no constant bet has positive expected log-return after the change, so we call a shift \emph{detectable} if $\Delta > 0$.
The following theorem holds for any valid allocation $(\pi_s)_{s \geq 1}$.

\begin{theorem}
\label{thm:delay}
Under~\eqref{eq:change-point} with a shift after round $t_0 \geq 0$, let $\sD_0$ satisfy $\Delta > 0$, let $\tau$ be the rejection time of Algorithm~\ref{alg:test}, and let
$
    n(t_0) := \left\lceil \frac{20}{\Delta^2} \ln \frac{2}{\alpha \pi_{t_0 + 1}} + \frac{40}{\Delta^2} \ln \frac{40}{\Delta^2} \right\rceil$.
Then,
\[
    \E_{H_1}[(\tau - t_0)_+ \mid \sD_0] \leq n(t_0) + \frac{170}{\Delta^2}~.
\]
\end{theorem}

Theorem~\ref{thm:delay-full} in Appendix~\ref{app:delay} adds an exponential tail bound on the delay and a bound given no false alarm before the shift.
With the heavy-tailed allocation~\eqref{eq:heavy-tailed},  Theorem~\ref{thm:delay} gives the following corollary.

\begin{corollary}
\label{cor:heavy}
Under the conditions of Theorem~\ref{thm:delay},
\begin{align*}
    & \E_{H_1}\left[(\tau - t_0)_+ \ \middle | \ \sD_0\right] \leq \frac{20}{\Delta^2} \ln \frac{1}{\alpha} + \frac{40}{\Delta^2} \ln \frac{40}{\Delta^2} + \frac{192}{\Delta^2} \\
    & \quad + \frac{20}{\Delta^2} \left(\ln(t_0 + 3) + 2 \ln\ln(t_0 + 3)\right)~.
\end{align*}
\end{corollary}

The bound separates three sources of delay: the evidence $\ln(1/\alpha)$ required by the test, a statistical cost of order $\Delta^{-2} \ln(1/\Delta)$, and the restarting cost of not knowing the shift time $t_0$.
At $t_0 = 0$ it recovers the order $\Delta^{-2} \ln(1/(\alpha\Delta))$ that~\citet{shaer2026testing} prove for a shift at the first observation.
For $t_0 > 0$, additional null observations enter only through the restart cost $\ln t_0 + 2\ln\ln t_0 + \sO(1)$, in contrast with the linear growth that constant bets incur (Appendix~\ref{app:constant}).

\paragraph{Unavoidable Logarithmic Cost.}
The logarithmic growth cannot be avoided.
Classical change detectors control the ARL to a false alarm, and their worst-case delay does not depend on $t_0$~\citep{lorden1971procedures}.
When a false alarm may \emph{ever} occur with probability at most $\alpha$, however, every detector must become more conservative as time passes, even one that knows $P$ and $Q$ \emph{a priori}: for every large $t$, some shift time $t_0$ between $t$ and $2t$ costs it an expected delay of order at least $\ln(t/\alpha)/\KL(Q \| P)$ (Appendix~\ref{app:lower-bound}).
The dependence of Corollary~\ref{cor:heavy} on $t_0$ is therefore optimal in order, and since Pinsker's inequality gives $\Delta^2 \leq \KL(Q \| P)/2$ for detectable shifts on $\sG$, the two bounds differ only in how they measure the size of the shift.

\section{Experiments}
\label{sec:experiments}

We test empirically how the detection delay depends on the unknown shift time.
We extend the synthetic and ImageNet-C benchmarks of~\citet{shaer2026testing} with a null phase of varying length $t_0$ before the shift.
Appendix~\ref{app:experiments} gives the full setup and all results.

\begin{figure}[t]
    \centering
    \includegraphics[width=\textwidth]{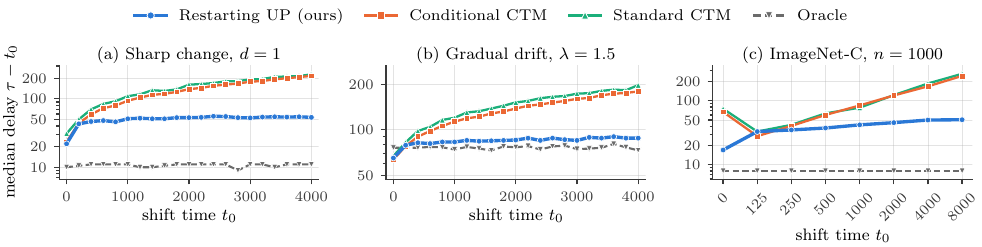}
    \caption{
        \textbf{Median detection delay vs.\ shift time} on logarithmic $y$ axes: (a) sharp change to $\sN(1, 1)$ and (b) gradual drift, with $100$ runs per $t_0$, and (c) ImageNet-C with $n = 1000$, with $150$ runs per $t_0$.
    }
    \label{fig:exp-delay}
\end{figure}

\begin{figure}[t]
    \centering
    \includegraphics[width=\textwidth]{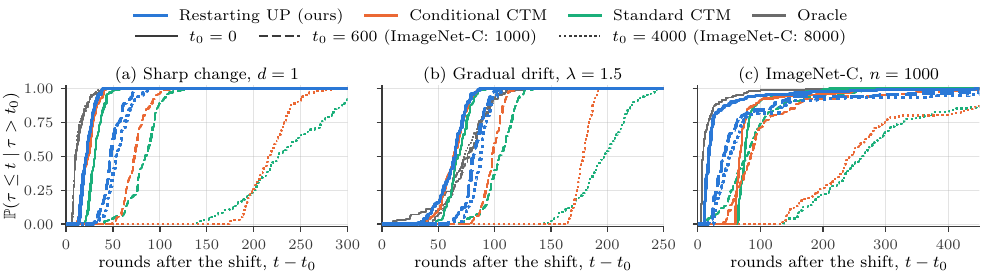}
    \caption{
        \textbf{Power given no false alarm}, 
        versus the number of rounds $t - t_0$ after the shift.
    }
    \label{fig:exp-power}
\end{figure}

\begin{figure}[t]
    \centering
    \includegraphics[width=\textwidth]{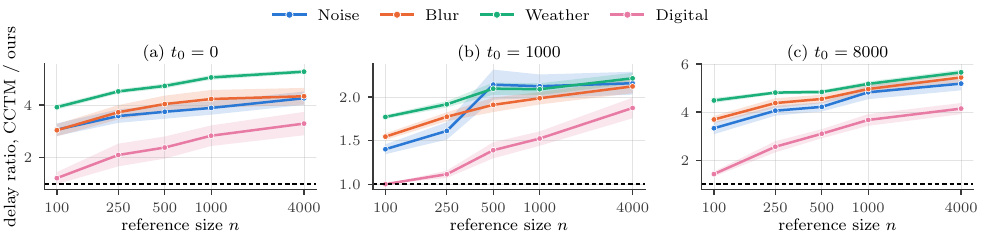}
    \caption{
        \textbf{ImageNet-C by corruption group.} 
        For each seed, we take the median ratio over the corruptions of the group; lines show the mean over $10$ seeds and bands one standard error.
        Values above 1 indicate faster detection by our detector.
    }
    \label{fig:exp-groups}
\end{figure}

\paragraph{Setup.}
In the synthetic benchmark, $P = \sN(0, 1)$, and after round $t_0 \in \{0, 200, \ldots, 4000\}$, the scores follow either a sharp change to $\sN(d, 1)$ with $d \in \{1, 1.5, 2\}$ and $n = 1000$, or a gradual drift with mean $\lambda (t - t_0)/100$ with $\lambda \in \{1.5, 3, 5\}$ and $n = 2000$, with $100$ runs per configuration.
In the ImageNet-C benchmark, the scores are the predictive entropies of a pretrained ViT-B/16 classifier~\citep{dosovitskiy2021image}, and $t_0 \in \{0, 125, \ldots, 8000\}$ clean ImageNet images~\citep{deng2009imagenet} precede their severity-$5$ corruptions~\citep{hendrycks2019benchmarking}, for $15$ corruptions, $10$ seeds, and $n \in \{100, 250, 500, 1000, 4000\}$ reference images.
All tests use $\alpha = 0.05$ and the DKW band.
Our detector is Algorithm~\ref{alg:test} with the heavy-tailed allocation~\eqref{eq:heavy-tailed}; it bets from the first round and has no hyperparameters.
We compare it with the conditional and the standard CTM, which use ONS with the hyperparameters and the warm-up of~\citet{shaer2026testing}, and place no bets during the warm-up.
As a reference, the \emph{oracle} holds cash until $t_0$ and then makes the best constant bet in hindsight, the comparator of Corollary~\ref{cor:heavy-tailed-regret}.
We report the median delay $\tau - t_0$ and the power $\Prob(\tau \leq t \mid \tau > t_0)$ over the runs without a false alarm, counting misses as infinite delays.
On both benchmarks, every method keeps its false-alarm rate below $\alpha = 0.05$, except that of the standard CTM on ImageNet-C with $n = 4000$, which is $0.10$ for $t_0 \geq 500$.

\paragraph{Detection Delay.}
Figure~\ref{fig:exp-delay} shows the median delay against $t_0$ for one configuration of each benchmark; Appendix~\ref{app:experiments} gives all configurations.
For an immediate shift ($t_0 = 0$), all methods are close on synthetic data, the case for which the CCTM was designed and tuned, while on ImageNet-C, most of the larger delay of the CTMs is their warm-up on corrupted images.
As $t_0$ grows, the delays of both CTMs grow roughly linearly, e.g., for the CCTM with $d = 1$, from $24$ rounds at $t_0 = 0$ to $216$ at $t_0 = 4000$.
In contrast, the delay of our detector grows logarithmically, as Theorem~\ref{thm:delay} predicts, by only $2.2$ to $3.8$ rounds each time $t_0$ grows by a factor of $e$ on synthetic data.
At the largest shift time, our detector is $2.0$ to $4.2$ times faster than the CCTM on synthetic data, and $3.4$ to $5.3$ times faster on ImageNet-C, e.g., $51$ against $245$ rounds for $n = 1000$.
On ImageNet-C with $n = 1000$, the CTMs are ahead only at $t_0 = 125$, where their warm-up manually avoids wealth loss early in the null phase.

\paragraph{Power.}
Figure~\ref{fig:exp-power} shows that the power curve of our detector hardly changes with $t_0$, while those of the CTMs move to the right as $t_0$ grows.
For $d = 1$ at $t_0 = 4000$, our detector detects $90\%$ of the shifts within $72$ rounds, against $245$ for the CCTM, and after $8000$ clean images, it detects $82\%$ of the shifts within $100$ rounds, while both CTMs detect none that quickly.

\paragraph{Corruption Groups.}
Figure~\ref{fig:exp-groups} compares the CCTM with our detector by corruption group and reference size.
The advantage of our detector grows with $t_0$ and $n$: the delay ratio is $1.0$ to $2.2$ at $t_0 = 1000$, where the warm-up of the CCTM falls into the clean images, and $1.4$ to $5.7$ at $t_0 = 8000$; at $t_0 = 0$, it mostly reflects this warm-up.
The Digital corruptions are the hardest for all methods and show the smallest gains.
For $n = 1000$, our detector is the fastest method in every group and at every shift time (Table~\ref{tab:exp-imagenet} in Appendix~\ref{app:exp-imagenet}).

\section{Discussion}

We introduced a parameter-free conditional conformal martingale for detecting distribution shifts at unknown times. The main idea is to recast the penalized betting problem as portfolio selection in a three-asset market and to restart a universal portfolio at every round. This eliminates the need for hyperparameter tuning in existing CCTMs while retaining anytime false-alarm control. It adapts to an unknown shift time at a logarithmic cost that our lower bound shows is unavoidable. We also developed an exact FFT-based implementation that runs in $\sO(T^2 \ln T)$ time. Experiments on synthetic data and ImageNet-C agree with the theory: our detection delay depends only weakly on the length of the null period, unlike that of existing CTMs. 

\paragraph{Limitations And Future Work.}
The detection-delay theory applies to shifts that are detectable by a constant post-shift bet, quantified by $\Delta>0$, and assumes a fixed alternative distribution $Q$. 
It thus does not directly cover nonstationary alternatives, such as recurring or gradual changes, although our experiments include the latter. 
One future direction is to replace the constant post-shift comparator by richer portfolio classes that can track time-varying alternatives while preserving anytime validity.
Our method also uses a fixed score and a fixed reference dataset, so its power depends on whether the score reveals the shift. 
Designing the score and the betting strategy jointly may improve sensitivity to weak shifts.

\bibliography{references}

\clearpage
\appendix

\section{Additional Related Work}
\label{app:extended_related_work}

% Here we provide additional discussion of related work on sequential testing, especially for testing exchangeability/i.i.d. and conformal test martingales.
% For a relatively recent review of sequential testing-by-betting literature, we refer to~\citep{ramdas2023game}.

\paragraph{Classical sequential hypothesis testing and change-point detection:} 
Classical approaches to sequential hypothesis testing and change-point detection assumed simple, parametric settings where the null and alternative hypotheses are each specified by a single parametric distribution, and they generally focused on finite monitoring horizons.
This literature dates as far back as Wald's sequential probability ratio test (SPRT)~\citep{wald1945sequential}, designed for testing the null that all data in a sequence are sampled from a given distribution $P$ against the alternative of all the data being sampled from a different distribution $Q$.
Change-point detection, in contrast, refers to sequential tests where the alternative hypothesis allows for a change to a different distribution $Q$ at an unknown time $t_0 \geq 0$. 
\citet{page1954continuous} introduced the CUSUM procedure for a simple, parametric change-point detection setting, where the procedure can be viewed as starting a Wald SPRT at every time step and rejecting the null as soon as the \textit{maximum} over sequentially started SPRTs exceeds a threshold.
The Shiryaev-Roberts procedure~\citep{shiryaev1961problem, shiryaev1963optimum, roberts1966comparison} is another classical approach to the same setting, which can be viewed as the \textit{sum} of SPRTs started at every timestep.
Validity guarantees for these classical procedures, however, have generally concerned the average run length (ARL) to a false alarm under the null hypothesis, rather than the anytime-valid setting of this paper. 
\citet{lorden1971procedures} and \citet{moustakides1986optimal} proved asymptotic and exact optimality results respectively for CUSUM with respect to minimizing the worst-case detection delay, for i.i.d. data from known distributions $P$ and $Q$.
Various optimality results exist for the Shiryaev-Roberts procedure as well, including those due to~\citet{shiryaev1963optimum, pollak1985optimal, pollak2009optimality}.

\paragraph{Sequential anytime-valid testing of composite and nonparametric hypotheses:}
Recent work has substantially advanced sequential testing of \textit{composite} and \textit{nonparametric} hypotheses with \textit{anytime-valid} guarantees over unbounded monitoring horizons.
A null or alternative hypothesis is composite if it specifies a non-singleton set of distributions, and it is nonparametric if such a set is not parametrically specified.
We refer to~\citet{ramdas2023game} for a more comprehensive review of this literature sometimes referred to as sequential anytime-valid inference (SAVI), to~\citet{shafer2021testing} for an overview of the closely related framework of testing-by-betting, and to~\citet{ramdas2025hypothesis} for a recent textbook on the closely related topic of hypothesis testing with e-values.
Anytime-valid guarantees fundamentally build on properties of test supermartingales~\citep{ramdas2020admissible}, especially Ville's inequality~\citep{ville1939etude}, which yields anytime-valid false-alarm control for nonnegative supermartingales.
Conformal test martingales (CTMs)~\citep{vovk2021testing} are a type of test supermartingale with anytime-valid guarantees for testing the null of exchangeability/i.i.d., and thus for testing for distribution shifts.
E-processes are an important generalization of test supermartingales (see \citealp[Section~2.4]{ramdas2023game}); for the purpose of change-point detection (with ARL control rather than anytime guarantees), e-detectors \citep{shin2022detectors} can be constructed by running CUSUM or Shiryaev-Roberts procedures on top of an e-process.

\paragraph{Standard and conditional conformal test martingales:}
Monitoring the null hypothesis of exchangeable or i.i.d. data from an \textit{unknown} distribution is a highly composite and nonparametric setting, and it is especially difficult when the time of a potential shift is also unknown.
Standard conformal test martingales (CTMs), originally developed by~\citet{vovk2003testing} and reviewed by~\citet{vovk2021testing}, provided the first framework for nontrivial anytime-valid testing of this null hypothesis.
CTMs construct a betting process adapted to the filtration generated by conformal p-values. This filtration is coarser than the canonical data filtration, which may help explain their power for change-point detection~\citep{ramdas2022testing,vovk2023power,volkhonskiy2017inductive,vovk2021retrain}.
Nonetheless, standard CTMs' detection speed can suffer due to test-time contamination~\citep{shaer2026testing} or when the shift occurs after a long null period due to difficulty recovering wealth lost or difficulty adapting to the shift~\citep{eliades2022betting}. 

\citet{shaer2026testing} inspired the present work with their introduction of conditional conformal test martingales (CCTMs), which accommodated a fixed reference dataset to avoid contamination from test-time samples by penalizing the betting function to account for finite-sample bias.
The fixed reference dataset additionally allowed \citet{shaer2026testing} to prove guarantees on the detection power and detection delay of CCTMs, which had previously been elusive for standard CTMs, but only for an immediate change point occurring at the first round of testing.
However, \citet{shaer2026testing} used the online Newton step of \citet{cutkosky2018black} to fit the penalized betting function, which involves several hyperparameters that require tuning and can adapt slowly when a shift occurs after a long null period.
The present paper addresses these limitations of CCTMs by introducing a parameter-free approach that is provably fast to adapt to an unknown shift time and empirically beats standard CTMs and the original CCTM when shifts occur after many null samples.
Our approach combines a three-asset reformulation of CCTM bet selection, universal portfolios restarted every round and mixed using a heavy-tailed wealth allocation, and a fast FFT-based implementation.
These mixture weights are distinct from the weights in weighted CTMs \citep{prinster2025watch}, which weight nonconformity scores to extend standard CTMs to the null of covariate shift and an invariant conditional label distribution.

\section{Proofs for Section~\ref{sec:market}}
\label{app:market}

\subsection{Proof of Proposition~\ref{prop:validity}}

The argument follows~\citet{shaer2026testing}; we include it for completeness.
Fix $\sD_0 \in \sG$ and a round $t \geq 1$.
Under the null, $X_t$ has law $P$ and is independent of $\sF_{t-1}$.
Since $F$ is continuous, $F(X_t)$ is uniform on $[0,1]$ and independent of $\sF_{t-1}$, so $\E_{H_0}[F(X_t) \mid \sF_{t-1}] = 1/2$.
Because $\sD_0 \in \sG$, the band~\eqref{eq:band} at $x = X_t$ gives $|\hat{p}_t - F(X_t)| \leq \epsilon(\hat{p}_t)$.
Hence
\[
    \abs{\E_{H_0}\left[\hat{p}_t \mid \sF_{t-1}\right] - \frac{1}{2}} = \abs{\E_{H_0}\left[\hat{p}_t - F\left(X_t\right) \mid \sF_{t-1}\right]} \leq \E_{H_0}\left[\epsilon\left(\hat{p}_t\right) \mid \sF_{t-1}\right]~.
\]
The bet $\eta_t$ is $\sF_{t-1}$-measurable, and $b(\eta_t, \cdot)$ is the sum of an affine function of $\hat{p}_t$ and the term $-\kappa |\eta_t| \epsilon(\hat{p}_t)$.
Therefore
\begin{align*}
    \E_{H_0}\left[b\left(\eta_t, \hat{p}_t\right) \mid \sF_{t-1}\right]
    &= 1 + \kappa \left(\eta_t \left(\E_{H_0}\left[\hat{p}_t \mid \sF_{t-1}\right] - \frac{1}{2}\right) - \abs{\eta_t} \E_{H_0}\left[\epsilon\left(\hat{p}_t\right) \mid \sF_{t-1}\right]\right) \\
    &\leq 1 + \kappa \abs{\eta_t} \left(\abs{\E_{H_0}\left[\hat{p}_t \mid \sF_{t-1}\right] - \frac{1}{2}} - \E_{H_0}\left[\epsilon\left(\hat{p}_t\right) \mid \sF_{t-1}\right]\right) \leq 1~.
\end{align*}
Since $M_{t-1}$ is $\sF_{t-1}$-measurable and nonnegative, $\E_{H_0}[M_t \mid \sF_{t-1}] = M_{t-1} \E_{H_0}[b(\eta_t, \hat{p}_t) \mid \sF_{t-1}] \leq M_{t-1}$.
Together with $b(\eta, p) \geq 0$ for all $\eta \in [-1,1]$ and $p \in [0,1]$ (Section~\ref{sec:setup}), this shows that $(M_t)_{t \geq 0}$ is a nonnegative supermartingale with $M_0 = 1$.
Ville's inequality~\citep{ville1939etude} gives the conditional bound.
For the unconditional bound, average over $\sD_0$, and use the conditional bound on $\sG$ and the trivial bound $1$ outside it:
\[
    \Prob_{H_0}\left\{\exists t \geq 1 : M_t \geq \frac{1}{\alpha}\right\} \leq \alpha \Prob\left\{\sD_0 \in \sG\right\} + \Prob\left\{\sD_0 \notin \sG\right\} = \alpha + \left(1 - \alpha\right) \Prob\left\{\sD_0 \notin \sG\right\}~.
\]
The right-hand side is at most $\alpha + (1 - \alpha)\delta$, since $\Prob\{\sD_0 \notin \sG\} \leq \delta$ by the choice of the band.
\hfill$\qed$

\subsection{Proof of Lemma~\ref{lem:equivalence}}

Fix $\bw \in \simplex$ and let $\eta := w^+ - w^- \in [-1, 1]$.
Since $w^+ + w^- = |w^+ - w^-| + 2 \min\ \{w^+, w^-\}$, substituting into~\eqref{eq:portfolio-return} gives
\[
    \inner{\bw}{\bx_t} = 1 + \kappa \left(\eta \left(\hat{p}_t - \frac{1}{2}\right) - \abs{\eta} \epsilon\left(\hat{p}_t\right)\right) - 2\kappa \min\ \left\{w^+, w^-\right\} \epsilon\left(\hat{p}_t\right),
\]
which is~\eqref{eq:gap} by the definition~\eqref{eq:bet}.

For $\eta \in [-1, 1]$, the coordinates of $\bw(\eta) = (1 - |\eta|, \eta^+, \eta^-)$ are nonnegative and sum to $1 - |\eta| + \eta^+ + \eta^- = 1$, so $\bw(\eta) \in \simplex$.
Its risky weights satisfy $\eta^+ - \eta^- = \eta$ and $\min\ \{\eta^+, \eta^-\} = 0$, so~\eqref{eq:gap} gives $\inner{\bw(\eta)}{\bx_t} = b(\eta, \hat{p}_t)$.

Fix rounds $1 \leq s \leq t$.
Both maxima in the last claim exist, since they maximize continuous functions over compact sets.
For every $\bw \in \simplex$, \eqref{eq:gap} and $\bx_i \in [0, 2]^3$ give $0 \leq \inner{\bw}{\bx_i} \leq b(w^+ - w^-, \hat{p}_i)$ in every round $i$.
Multiplying these nonnegative factors over $i = s, \ldots, t$ gives
\[
    \prod_{i=s}^{t} \inner{\bw}{\bx_i} \leq \prod_{i=s}^{t} b\left(w^+ - w^-, \hat{p}_i\right) \leq \max_{\eta \in [-1, 1]}\ \prod_{i=s}^{t} b\left(\eta, \hat{p}_i\right),
\]
and maximizing over $\bw$ proves the inequality ``$\leq$''.
Conversely, for every $\eta \in [-1, 1]$, the second claim gives $\prod_{i=s}^{t} b(\eta, \hat{p}_i) = \prod_{i=s}^{t} \inner{\bw(\eta)}{\bx_i}$, which is at most the maximum over $\simplex$.
Maximizing over $\eta$ proves the inequality ``$\geq$''.
\hfill$\qed$

\subsection{Proof of Proposition~\ref{prop:validity-2}}

Since $\bw_t \in \simplex$ is $\sF_{t-1}$-measurable, so is $\eta_t = w_t^+ - w_t^- \in [-1, 1]$.
By~\eqref{eq:gap}, $b(\eta_t, \hat{p}_t) = \inner{\bw_t}{\bx_t} + 2\kappa \min\ \{w_t^+, w_t^-\} \epsilon(\hat{p}_t) \geq \inner{\bw_t}{\bx_t} \geq 0$ in every round.
Multiplying these nonnegative factors gives $M_t \geq \Wealth_t$ for every $t \geq 0$.
Since $\{\tau < \infty\} = \{\exists t \geq 1 : M_t \geq 1/\alpha\}$, the remaining claims are Proposition~\ref{prop:validity} applied to $(\eta_t)_{t \geq 1}$.
\hfill$\qed$

\section{Proofs for Section~\ref{sec:restarting}}
\label{app:restarting}

\subsection{Constant Bets After a Long Null Phase}
\label{app:constant}

This appendix makes precise the claim of Section~\ref{sec:restarting} that constant bets adapt slowly to late shifts.
The loss of a constant bet before the shift and its gain after the shift are governed by
\[
    \Delta_P := \E_P\left[\epsilon\left(\hat{p}_t\right) \ \middle|\ \sD_0\right] - \abs{\E_P\left[\hat{p}_t - \frac{1}{2} \ \middle|\ \sD_0\right]} \quad \text{and} \quad \Delta_Q := \abs{\E_Q\left[\hat{p}_t - \frac{1}{2} \ \middle|\ \sD_0\right]} - \E_Q\left[\epsilon\left(\hat{p}_t\right) \ \middle|\ \sD_0\right],
\]
where the subscript gives the law of $X_t$, so that $\Delta_Q$ is the signal strength~\eqref{eq:signal}.
For a bet $\eta \in [-1, 1]$, let $\tau_{\eta} := \inf\ \{t \geq 1 : \prod_{i=1}^{t} b(\eta, \hat{p}_i) \geq 1/\alpha\}$ be the rejection time of the test that bets $\eta$ in every round.

\begin{proposition}
    \label{prop:constant}
    Under~\eqref{eq:change-point} with a shift after round $t_0 \geq 0$, let $\sD_0 \in \sG$.
    Then $\Delta_P \geq 0$, and the following hold.
    \begin{enumerate}
        \item[(a)] For every $\eta \in [-1, 1]$ and every $t \geq t_0$,
        \[
            \E_{H_1}\left[\sum_{i=1}^{t} \ln b\left(\eta, \hat{p}_i\right) \ \middle|\ \sD_0\right] \leq \kappa \abs{\eta} \left(\left(t - t_0\right) \Delta_Q - t_0 \Delta_P\right)~.
        \]
        \item[(b)] For every $\eta$ with $0 < |\eta| < 1$, if $\Delta_Q > 0$, then
        \[
            \E_{H_1}\left[\left(\tau_{\eta} - t_0\right)_+ \ \middle|\ \sD_0\right] \geq \left(1 - \alpha\right) \left(\frac{\Delta_P}{\Delta_Q} t_0 + \frac{\ln\left(1/\alpha\right)}{\kappa \abs{\eta} \Delta_Q}\right),
        \]
        and if $\Delta_Q \leq 0$, the left-hand side is infinite.
    \end{enumerate}
\end{proposition}

By (a), if $\Delta_Q > 0$, the expected log-wealth of every constant bet is nonpositive until $t \geq t_0 + t_0 \Delta_P / \Delta_Q$, a number of post-shift rounds linear in $t_0$.
By (b), the same holds for the expected detection delay whenever $\Delta_P > 0$: the wealth lost before the shift must be recovered before the test can reject.
In contrast, Corollary~\ref{cor:heavy} bounds the expected delay of the restarting portfolio by a function of $t_0$ that grows only logarithmically.
Part (b) does not apply to learners that change their bets, such as ONS, but by (a), a regret bound against constant bets guarantees them little over the same number of post-shift rounds.
We exclude $|\eta| = 1$ from (b) only to keep the log-returns bounded.

\paragraph{Proof of Proposition~\ref{prop:constant}.}
Throughout, we condition on $\sD_0 \in \sG$ and omit it from the notation.
Given $\sD_0$, the $p$-values $\hat{p}_1, \hat{p}_2, \ldots$ are independent, and $\hat{p}_i$ has the law of $\hat{F}_0(X)$ with $X \sim P$ for $i \leq t_0$ and with $X \sim Q$ for $i > t_0$.
We first show $\Delta_P \geq 0$.
Under $H_0$, the conditional law of $\hat{p}_t$ given $\sF_{t-1}$ is the law of $\hat{F}_0(X)$ with $X \sim P$, so the first display in the proof of Proposition~\ref{prop:validity} reads $|\E_P[\hat{p}_t - 1/2]| \leq \E_P[\epsilon(\hat{p}_t)]$.

(a) Fix $\eta \in [-1, 1]$.
For $i \leq t_0$, Jensen's inequality, the inequality $\ln(1 + z) \leq z$, and $\eta y \leq |\eta| |y|$ give
\[
    \E_{H_1}\left[\ln b\left(\eta, \hat{p}_i\right)\right] \leq \ln \E_{H_1}\left[b\left(\eta, \hat{p}_i\right)\right] \leq \kappa \abs{\eta} \left(\abs{\E_P\left[\hat{p}_i - \frac{1}{2}\right]} - \E_P\left[\epsilon\left(\hat{p}_i\right)\right]\right) = -\kappa \abs{\eta} \Delta_P~.
\]
For $i > t_0$, the same steps under $Q$ give the upper bound $\kappa |\eta| \Delta_Q$.
Summing over $i = 1, \ldots, t$ proves (a).

(b) Fix $\eta$ with $0 < |\eta| < 1$, and write $\tau := \tau_{\eta}$, $N := (\tau - t_0)_+$, $Y_i := \ln b(\eta, \hat{p}_i)$, and $S_t := \sum_{i=1}^{t} Y_i$, so that $\tau = \inf\ \{t \geq 1 : S_t \geq \ln(1/\alpha)\}$.
Since $1 - |\eta| \leq b(\eta, p) \leq 2$ for all $p \in [0, 1]$ by the definition of $\kappa$, the variables $Y_i$ are bounded.
They are independent, $Y_i$ is independent of $\sF_{i-1}$, and their means $\mu_P := \E_{H_1}[Y_i]$ for $i \leq t_0$ and $\mu_Q := \E_{H_1}[Y_i]$ for $i > t_0$ satisfy $\mu_P \leq -\kappa |\eta| \Delta_P \leq 0$ and $\mu_Q \leq \kappa |\eta| \Delta_Q$ by the proof of (a).

We first bound the log-wealth at the shift on the event of no false alarm.
The event $\{\tau \leq t_0\}$ depends only on $\sD_0, X_1, \ldots, X_{t_0}$, whose law is the same under $H_0$ and $H_1$, so Proposition~\ref{prop:validity} gives $\Prob_{H_1}\{\tau \leq t_0\} \leq \alpha$.
Since $\sum_{i=1}^{t_0} Y_i \ind\{\tau < i\}$ equals $S_{t_0} - S_{\tau}$ on $\{\tau \leq t_0\}$ and vanishes on $\{\tau > t_0\}$, we have $S_{t_0} \ind\{\tau \leq t_0\} = S_{\tau} \ind\{\tau \leq t_0\} + \sum_{i=1}^{t_0} Y_i \ind\{\tau < i\}$, where $S_{\tau} \geq \ln(1/\alpha) > 0$ on $\{\tau \leq t_0\}$.
Since $\{\tau < i\} \in \sF_{i-1}$, each term satisfies $\E_{H_1}[Y_i \ind\{\tau < i\}] = \mu_P \Prob_{H_1}\{\tau < i\} \geq \alpha \mu_P$, and therefore
\[
    \E_{H_1}\left[-S_{t_0} \ind\left\{\tau > t_0\right\}\right] = -t_0 \mu_P + \E_{H_1}\left[S_{t_0} \ind\left\{\tau \leq t_0\right\}\right] \geq -t_0 \mu_P + \alpha t_0 \mu_P = \left(1 - \alpha\right) t_0 \abs{\mu_P}~.
\]

We now turn to the rounds after the shift.
If $\E_{H_1}[N] = \infty$, there is nothing to prove, so assume $\E_{H_1}[N] < \infty$.
The delay $N$ is a stopping time with respect to $(\sF_{t_0 + k})_{k \geq 0}$, and $Y_{t_0 + k}$ is independent of $\sF_{t_0 + k - 1}$ with mean $\mu_Q$ for every $k \geq 1$, so Wald's identity gives $\E_{H_1}[\sum_{k=1}^{N} Y_{t_0 + k}] = \mu_Q \E_{H_1}[N]$.
On $\{\tau > t_0\}$, the delay $N = \tau - t_0$ is finite almost surely, and $S_{t_0} + \sum_{k=1}^{N} Y_{t_0 + k} = S_{\tau} \geq \ln(1/\alpha)$; on $\{\tau \leq t_0\}$, the sum is empty.
Together with $\Prob_{H_1}\{\tau > t_0\} \geq 1 - \alpha$ and the previous display,
\[
    \mu_Q \E_{H_1}\left[N\right] \geq \E_{H_1}\left[\left(\ln \frac{1}{\alpha} - S_{t_0}\right) \ind\left\{\tau > t_0\right\}\right] \geq \left(1 - \alpha\right) \left(\ln \frac{1}{\alpha} + t_0 \abs{\mu_P}\right) > 0~.
\]
Hence $\mu_Q > 0$, and thus $\Delta_Q > 0$, since $\mu_Q \leq \kappa |\eta| \Delta_Q$.
This shows that $\E_{H_1}[N] = \infty$ if $\Delta_Q \leq 0$.
If $\Delta_Q > 0$, dividing by $\mu_Q \leq \kappa |\eta| \Delta_Q$ and using $|\mu_P| \geq \kappa |\eta| \Delta_P$ proves (b).
\hfill$\qed$

\subsection{Proof of Proposition~\ref{prop:restarting}}

Throughout, $\Wealth_t$ denotes the mixture~\eqref{eq:restarting-wealth}, so that $\Wealth_0 = r_0 = 1$, and $Z_t := \int_{\simplex} B_t(\bw) \mu(d\bw) + r_t$ denotes the denominator in~\eqref{eq:restarting-allocation}.
Unrolling the recursion gives $B_t(\bw) = \sum_{i=1}^{t} \pi_i \prod_{j=i}^{t-1} \inner{\bw}{\bx_j}$ for every $t \geq 1$, where the empty product is one.
Since $\bx_u \in [0, 2]^3$, the polynomial $B_t$ is nonnegative on $\simplex$, and the definition of $\Wealth_t^s$ gives, for every $t \geq 1$,
\[
    \int_{\simplex} \inner{\bw}{\bx_t} B_t\left(\bw\right) \mu\left(d\bw\right) = \sum_{i=1}^{t} \pi_i \int_{\simplex} \prod_{j=i}^{t} \inner{\bw}{\bx_j} \mu\left(d\bw\right) = \sum_{i=1}^{t} \Wealth_t^i~.
\]
We first show that $Z_t = \Wealth_{t-1}$.
Splitting off the term $i = t$, whose product is empty, gives $\int_{\simplex} B_t(\bw) \mu(d\bw) = \sum_{i=1}^{t-1} \Wealth_{t-1}^i + \pi_t$, and $\pi_t + r_t = r_{t-1}$, so $Z_t = \sum_{i=1}^{t-1} \Wealth_{t-1}^i + r_{t-1} = \Wealth_{t-1}$.
In particular, $Z_t \geq r_t > 0$, because all weights $\pi_s$ are positive, so $\bw_t$ is well defined.
The numerator of~\eqref{eq:restarting-allocation} has nonnegative coordinates, which sum to $Z_t$ because $w^0 + w^+ + w^- = 1$ on $\simplex$, so $\bw_t \in \simplex$.
Moreover, $\bw_t$ depends only on $\bx_1, \ldots, \bx_{t-1}$, which are functions of $\sD_0, X_1, \ldots, X_{t-1}$, so it is $\sF_{t-1}$-measurable.
Finally, since $\inner{(1, 0, 0)}{\bx_t} = x_t^0 = 1$,
\[
    \Wealth_{t-1} \inner{\bw_t}{\bx_t} = \int_{\simplex} \inner{\bw}{\bx_t} B_t\left(\bw\right) \mu\left(d\bw\right) + r_t = \sum_{i=1}^{t} \Wealth_t^i + r_t = \Wealth_t~.
\]
By induction from $\Wealth_0 = 1$, the wealth $\prod_{i=1}^{t} \inner{\bw_i}{\bx_i}$ of $(\bw_t)_{t \geq 1}$ equals~\eqref{eq:restarting-wealth} for every $t \geq 0$.
\hfill$\qed$

\subsection{The Regret Bound\texorpdfstring{~\eqref{eq:up-regret}}{}}

\citet[Theorem~2]{cover1996universal} show that, in a market with three assets and arbitrary nonnegative price relatives, the universal portfolio with the prior $\mu = \Dirichlet(1/2, 1/2, 1/2)$ started at round $s$ earns by round $t$ at least the fraction
\[
    \frac{\Gamma(3/2) \Gamma(t - s + 3/2)}{\Gamma(1/2) \Gamma(t - s + 5/2)}
\]
of the wealth of the best CRP in hindsight over the rounds $s, \ldots, t$.
Since $\Gamma(3/2) = \Gamma(1/2)/2$ and $\Gamma(t - s + 5/2) = (t - s + 3/2) \Gamma(t - s + 3/2)$, this fraction equals $1/(2(t - s) + 3)$.
The wealth of this universal portfolio is the mixture on the left-hand side of~\eqref{eq:up-regret}, so taking logarithms gives~\eqref{eq:up-regret}.

\subsection{Proof of Lemma~\ref{lem:restarting-regret}}

By Proposition~\ref{prop:restarting}, the wealth $\Wealth_t$ of the restarting portfolio equals~\eqref{eq:restarting-wealth}, whose terms are all nonnegative, so $\Wealth_t \geq \Wealth_t^s$ for every $s \leq t$.
By the regret bound~\eqref{eq:up-regret} and Lemma~\ref{lem:equivalence},
\[
    \ln \Wealth_t^s - \ln \pi_s \geq \max_{\bw \in \simplex}\ \sum_{i=s}^{t} \ln \inner{\bw}{\bx_i} - \ln\left(2(t - s) + 3\right) = \max_{\eta \in [-1, 1]}\ \sum_{i=s}^{t} \ln b\left(\eta, \hat{p}_i\right) - \ln\left(2(t - s) + 3\right)~.
\]
This proves the second inequality, and Proposition~\ref{prop:validity-2} gives $M_t \geq \Wealth_t$.
\hfill$\qed$

\subsection{The Restart Cost of the Heavy-Tailed Allocation}
\label{app:heavy-tail}

We show that the allocation~\eqref{eq:heavy-tailed} satisfies, for every $s \geq 1$,
\begin{equation}
    \label{eq:restarting-regret}
    \ln \frac{1}{\pi_s} \leq \ln\left(s + 2\right) + 2 \ln \ln\left(s + 2\right) + \ln \frac{1}{\ln 2}~,
\end{equation}
which, plugged into Lemma~\ref{lem:restarting-regret}, gives Corollary~\ref{cor:heavy-tailed-regret}.
Since $x \mapsto 1/(x \ln^2 x)$ monotonically decreases for $x > 1$,
\[
    \pi_s = \ln 2 \int_{s+1}^{s+2} \frac{dx}{x \ln^2 x} \geq \frac{\ln 2}{\left(s + 2\right) \ln^2\left(s + 2\right)},
\]
and taking logarithms gives~\eqref{eq:restarting-regret}.
\hfill$\qed$

\section{Proofs for Section~\ref{sec:guarantees}}
\label{app:guarantees}

Throughout this appendix, $X$ denotes a random variable with law $Q$ that is independent of $\sD_0$, and $\hat{p} := \hat{F}_0(X)$.
Under~\eqref{eq:change-point} and given $\sD_0$, the $p$-values $\hat{p}_t$ with $t > t_0$ are i.i.d.\ copies of $\hat{p}$, so that $\Delta = |\E_Q[\hat{p} - 1/2 \mid \sD_0]| - \E_Q[\epsilon(\hat{p}) \mid \sD_0]$.

\subsection{False-Alarm Control}

\begin{corollary}
    \label{cor:validity}
    Let $(\bw_t)_{t \geq 1}$ be the restarting portfolio~\eqref{eq:restarting-allocation}, define $(M_t)_{t \geq 0}$ by~\eqref{eq:martingale}, and let $\tau := \inf\ \{t \geq 1 : M_t \geq 1/\alpha\}$.
    If $\sD_0 \in \sG$, then $\Prob_{H_0}\{\tau < \infty \mid \sD_0\} \leq \alpha$.
    Without conditioning, $\Prob_{H_0}\{\tau < \infty\} \leq \alpha + (1 - \alpha)\delta$.
\end{corollary}

\begin{proof}
By Proposition~\ref{prop:restarting}, the restarting portfolio is predictable, so the claims follow from Proposition~\ref{prop:validity-2}.
\end{proof}

\subsection{A Constant Bet With Positive Growth}

The proof of Theorem~\ref{thm:delay-full} compares the detector with a single constant bet, which is chosen with the knowledge of $Q$ and $\sD_0$ but not of the test observations.

\begin{lemma}
    \label{lem:good-bet}
    Fix $\sD_0$ with $\Delta > 0$, let $h := \sign(\E_Q[\hat{p} - 1/2 \mid \sD_0])$, and let $\hat{\eta} := h \Delta / (2\kappa(1 + \Delta^2))$.
    Then $|\hat{\eta}| \leq 3/10$, and $Y := \ln b(\hat{\eta}, \hat{p})$ satisfies
    \[
        Y \in \left[\ln \frac{7}{10}, \ln \frac{13}{10}\right], \quad \bar{Y} := \E_Q\left[Y \mid \sD_0\right] \geq \frac{\Delta^2}{4\left(1 + \Delta^2\right)} \geq \frac{\Delta^2}{5}, \quad \text{and} \quad \Var_Q\left(Y \mid \sD_0\right) \leq 4 \bar{Y}~.
    \]
\end{lemma}

\begin{proof}
All expectations and variances below are conditional on $\sD_0$, and we omit the conditioning.
The sign $h$ is nonzero because $\Delta > 0$.
Let $Z := h(\hat{p} - 1/2) - \epsilon(\hat{p})$, so that $\E_Q[Z] = \Delta$ by the choice of $h$.
Since $\epsilon$ takes values in $[0, 1]$, the variable $Z$ lies in $[-1/2 - \max_{p \in [0,1]}\ \epsilon(p), 1/2] \subseteq [-3/2, 1/2]$, an interval of length at most $2$, so $\Var_Q(Z) \leq 1$ and $\E_Q[Z^2] \leq 1 + \Delta^2$.
Let $c := \kappa |\hat{\eta}| = \Delta / (2(1 + \Delta^2))$.
Since $\Delta \leq 1/2$ and $z \mapsto z/(1 + z^2)$ increases on $[0, 1]$, we have $c \leq 1/5$, and $\kappa \geq 2/3$ gives $|\hat{\eta}| \leq 3/10$.
Because $\hat{\eta} = h |\hat{\eta}|$, the definition~\eqref{eq:bet} gives $b(\hat{\eta}, \hat{p}) = 1 + c Z$ with $|c Z| \leq 3/10$, which proves the range of $Y$.
The inequality $\ln(1 + z) \geq z - z^2$ for $z \geq -1/2$ gives
\[
    \bar{Y} \geq c \Delta - c^2 \left(1 + \Delta^2\right) = \frac{\Delta^2}{4\left(1 + \Delta^2\right)} \geq \frac{\Delta^2}{5},
\]
where the last step uses $\Delta \leq 1/2$.
Finally, $z \mapsto \ln(1 + z)$ is $2$-Lipschitz on $[-1/2, \infty)$, and a $2$-Lipschitz function of a random variable has at most four times its variance, so
\[
    \Var_Q\left(Y\right) \leq 4 c^2 \Var_Q\left(Z\right) \leq \frac{\Delta^2}{\left(1 + \Delta^2\right)^2} \leq \frac{\Delta^2}{1 + \Delta^2} \leq 4 \bar{Y}~. \qedhere
\]
\end{proof}

\subsection{Full Statement and Proof of Theorem~\ref{thm:delay}}
\label{app:delay}

Theorem~\ref{thm:delay} states the bound on the expected delay.
We prove the following full version, which also bounds the tail of the delay and the delay given no false alarm before the shift.

\begin{theorem}[Detection delay, full statement]
    \label{thm:delay-full}
    Under~\eqref{eq:change-point} with a shift after round $t_0 \geq 0$, let $\sD_0$ satisfy $\Delta > 0$, let $\tau$ be the rejection time of Algorithm~\ref{alg:test}, and let $n(t_0)$ be as in Theorem~\ref{thm:delay}.
    Then for every integer $t_1 \geq n(t_0)$,
    \[
        \Prob_{H_1}\left\{\tau > t_0 + t_1 \ \middle|\ \sD_0\right\} \leq \exp\left(-\frac{\Delta^2 t_1}{170}\right)~.
    \]
    Consequently,
    \[
        \E_{H_1}\left[\left(\tau - t_0\right)_+ \ \middle|\ \sD_0\right] \leq n(t_0) + \frac{170}{\Delta^2}~.
    \]
    If moreover $\sD_0 \in \sG$, then $\Prob_{H_1}\{\tau > t_0 + t_1 \mid \tau > t_0, \sD_0\} \leq \exp(-\Delta^2 t_1 / 170) / (1 - \alpha)$ for all $t_1 \geq n(t_0)$.
\end{theorem}

The first two bounds hold for every reference dataset with $\Delta > 0$, and the second is Theorem~\ref{thm:delay}.
The conditional bound uses $\sD_0 \in \sG$ only to ensure that a false alarm before the shift is unlikely, so that conditioning on $\tau > t_0$ costs at most a factor $1/(1 - \alpha)$.

\paragraph{Proof of Theorem~\ref{thm:delay-full}.}
Fix $\sD_0$ with $\Delta > 0$ and an integer $t_1 \geq n(t_0)$, and write $s := t_0 + 1$ and $t := t_0 + t_1$.
Let $\hat{\eta}$, $Y$, and $\bar{Y}$ be as in Lemma~\ref{lem:good-bet}, and let $\Lambda := \sum_{i=s}^{t} \ln b(\hat{\eta}, \hat{p}_i)$ be the log-wealth that the constant bet $\hat{\eta}$ earns after the shift.
Since $\tau$ is the first round with $M_t \geq 1/\alpha$, Lemma~\ref{lem:restarting-regret} and $2 t_1 + 1 \leq 2(t_1 + 1)$ give
\[
    \left\{\tau > t\right\} \subseteq \left\{\ln M_t < \ln \frac{1}{\alpha}\right\} \subseteq \left\{\Lambda < \ln \frac{2}{\alpha \pi_s} + \ln\left(t_1 + 1\right)\right\}~.
\]
We bound both terms of the threshold by $t_1 \bar{Y}/4$.
Both terms in the definition of $n(t_0)$ are nonnegative, so $t_1 \geq n(t_0)$ implies $\ln(2/(\alpha \pi_s)) \leq t_1 \Delta^2/20 \leq t_1 \bar{Y}/4$ by Lemma~\ref{lem:good-bet}.
For the second term, let $c := 40/\Delta^2 \geq 160$, so that $t_1 \geq c \ln c$ and $\Delta^2/20 = 2/c$.
Since $z \mapsto \ln(z + 1)/z$ decreases on $(0, \infty)$, and since $c \ln c + 1 \leq c^2$ by $\ln c \leq c - 1$,
\[
    \frac{\ln\left(t_1 + 1\right)}{t_1} \leq \frac{\ln\left(c \ln c + 1\right)}{c \ln c} \leq \frac{2 \ln c}{c \ln c} = \frac{\Delta^2}{20} \leq \frac{\bar{Y}}{4}~.
\]
The threshold is therefore at most $t_1 \bar{Y}/2$, and $\{\tau > t\} \subseteq \{t_1 \bar{Y} - \Lambda > t_1 \bar{Y}/2\}$.
Given $\sD_0$, the terms of $\Lambda$ are i.i.d.\ copies of $Y$, so $t_1 \bar{Y} - \Lambda$ is a sum of $t_1$ independent centered variables, each with variance at most $4 \bar{Y}$ by Lemma~\ref{lem:good-bet}, and each bounded above by $\ln(13/10) - \ln(7/10) \leq \ln 3$.
Bernstein's inequality~\citep[Corollary~2.11]{boucheron2013concentration} gives
\[
    \Prob_{H_1}\left\{\tau > t_0 + t_1 \ \middle|\ \sD_0\right\} \leq \exp\left(-\frac{\left(t_1 \bar{Y}/2\right)^2}{2\left(4 t_1 \bar{Y} + (\ln 3) t_1 \bar{Y}/6\right)}\right) = \exp\left(-\frac{t_1 \bar{Y}}{32 + (4/3) \ln 3}\right) \leq \exp\left(-\frac{\Delta^2 t_1}{170}\right),
\]
where the last step uses $\bar{Y} \geq \Delta^2/5$ and $5(32 + (4/3) \ln 3) < 170$.
This proves the tail bound.

For the expectation, let $\rho := \Delta^2/170$.
Since $n(t_0) \geq 1$, the tail bound gives
\[
    \E_{H_1}\left[\left(\tau - t_0\right)_+ \ \middle|\ \sD_0\right] = \sum_{i=0}^{\infty} \Prob_{H_1}\left\{\tau > t_0 + i \ \middle|\ \sD_0\right\} \leq n(t_0) + \sum_{i=n(t_0)}^{\infty} e^{-\rho i} \leq n(t_0) + \frac{1}{e^{\rho} - 1} \leq n(t_0) + \frac{1}{\rho}~.
\]

Finally, let $\sD_0 \in \sG$.
The event $\{\tau > t_0\}$ depends only on $\sD_0, X_1, \ldots, X_{t_0}$, whose law is the same under $H_0$ and $H_1$, so Corollary~\ref{cor:validity} gives $\Prob_{H_1}\{\tau > t_0 \mid \sD_0\} = \Prob_{H_0}\{\tau > t_0 \mid \sD_0\} \geq 1 - \alpha$.
Dividing the tail bound by this probability proves the last claim.
\hfill$\qed$

\subsection{Proof of Corollary~\ref{cor:heavy}}

By~\eqref{eq:restarting-regret} with $s = t_0 + 1$,
\[
    \ln \frac{2}{\alpha \pi_{t_0+1}} \leq \ln \frac{1}{\alpha} + \ln\left(t_0 + 3\right) + 2 \ln \ln\left(t_0 + 3\right) + \ln 2 + \ln \frac{1}{\ln 2}~.
\]
We substitute this bound into $n(t_0) \leq (20/\Delta^2) \ln(2/(\alpha \pi_{t_0+1})) + (40/\Delta^2) \ln(40/\Delta^2) + 1$ and use $20 (\ln 2 + \ln(1/\ln 2)) < 21.2$ and $1 \leq 1/(4\Delta^2)$, which holds since $\Delta \leq 1/2$.
The expectation bound of Theorem~\ref{thm:delay-full} then gives the claim with the constant $21.2 + 0.25 + 170 < 192$.
\hfill$\qed$

\subsection{A Lower Bound on the Expected Delay}
\label{app:lower-bound}

Classical change detectors control the average run length to a false alarm, and their worst-case delay does not depend on $t_0$~\citep{lorden1971procedures}.
Under the stronger requirement that a false alarm \emph{ever} occurs with probability at most $\alpha$, the detector must become more conservative as time passes.
The next theorem shows that every such detector, even one that knows $P$ and $Q$ \emph{a priori}, has an expected delay that grows at least logarithmically with the shift time.

\begin{theorem}[Lower bound on the expected delay]
    \label{thm:lower-bound}
    Suppose that $Q$ is absolutely continuous with respect to $P$ and that the Kullback--Leibler divergence $\KL(Q \| P)$ is finite and positive.
    Let $\tau$ be any stopping time with $\Prob_{H_0}\{\tau < \infty\} \leq \alpha$.
    Then, as $t \to \infty$,
    \begin{equation*}
        \begin{aligned}
            &\max_{t_0 \in \{t, \ldots, 2t\}}\ \E_{H_1}\left[\left(\tau - t_0\right)_+\right] \\
            & \quad \geq \frac{\left(1 - \alpha\right)^2}{4 \KL(Q \| P)} \ln \frac{t}{\alpha} - \sO\left(\ln \ln t\right),
        \end{aligned}
    \end{equation*}
    where the constant in $\sO(\cdot)$ depends only on $\alpha$ and $\KL(Q \| P)$.
\end{theorem}

In words, for any choice of a detector, there are shift times $t_0$ between $t$ and $2t$ at which its expected delay is at least of order $\ln(t_0/\alpha)/\KL(Q \| P)$.
The theorem applies to our setting conditionally on $\sD_0$, and a detector that knows neither $P$ nor $Q$ can only do worse.

\paragraph{Proof of Theorem~\ref{thm:lower-bound}.}
Write $I := \KL(Q \| P)$.
The proof first shows that a detector cannot raise alarms in short windows after many different shift times without paying in false alarms, and then converts a bound on the expected delay into such windows with Markov's inequality.
Throughout, $\Prob_{H_1}$ and $\E_{H_1}$ refer to the shift time $t_0$ indicated in the event, and $\tau$ is a stopping time with respect to $(\sF_t)_{t \geq 0}$.
Since $\sD_0$ is independent of the test observations and has the same law under $H_0$ and $H_1$, the same argument applies conditionally on $\sD_0$.

\begin{lemma}
    \label{lem:windows}
    Let $\tau$ be a stopping time with $\Prob_{H_0}\{\tau < \infty\} \leq \alpha$, and fix integers $t \geq d \geq 1$ and a number $q \in (0, 1]$.
    If $\Prob_{H_1}\{t_0 < \tau \leq t_0 + d\} \geq q$ for every shift time $t_0 \in \{t, t + 1, \ldots, 2t - d\}$, then
    \[
        d I \geq q \ln \frac{\lfloor t / d \rfloor}{\alpha} - \ln 2~.
    \]
\end{lemma}

\begin{proof}
Let $m := \lfloor t / d \rfloor \geq 1$ and consider the shift times $t^{(j)} := t + (j - 1) d$ for $j = 1, \ldots, m$, which lie in $\{t, \ldots, 2t - d\}$.
The events $A_j := \{t^{(j)} < \tau \leq t^{(j)} + d\}$ are disjoint, so their probabilities under $H_0$ sum to at most $\Prob_{H_0}\{\tau < \infty\} \leq \alpha$, and some $j$ satisfies $p := \Prob_{H_0} A_j \leq \alpha / m$.
Write $\Prob_j$ for the law under $H_1$ with the shift time $t^{(j)}$, so that $q_j := \Prob_j A_j \geq q$.
Since $\tau$ is a stopping time, the event $A_j$ is determined by $\sD_0, X_1, \ldots, X_{t^{(j)} + d}$.
Under $\Prob_j$ and under $H_0$, these variables have the same law except for the last $d$ observations, which are i.i.d.\ from $Q$ and from $P$, respectively, and independent of the others.
By the chain rule, the Kullback--Leibler divergence between the two laws is $d I$.
Moreover, since $Q$ is absolutely continuous with respect to $P$, so is the first law with respect to the second, and $q_j > 0$ forces $p > 0$.
The data-processing inequality applied to the indicator of $A_j$, and the bound $-q_j \ln q_j - (1 - q_j) \ln (1 - q_j) \leq \ln 2$ on the binary entropy, give
\[
    d I \geq q_j \ln \frac{q_j}{p} + \left(1 - q_j\right) \ln \frac{1 - q_j}{1 - p} \geq q_j \ln \frac{1}{p} - \ln 2 \geq q \ln \frac{m}{\alpha} - \ln 2,
\]
where the second inequality drops the nonnegative term $(1 - q_j)\ln(1/(1 - p))$, and the third uses $q_j \geq q$ and $1/p \geq m/\alpha \geq 1$.
\end{proof}

We now prove the theorem.
Let $c := (1 - \alpha)^2/(4I)$ and $D := \max_{t_0 \in \{t, \ldots, 2t\}}\ \E_{H_1}[(\tau - t_0)_+]$.
If $D \geq c \ln(t/\alpha)$, there is nothing to prove, so assume $D < c \ln(t / \alpha)$.
For every shift time $t_0$, the event $\{\tau > t_0\}$ depends only on $\sD_0, X_1, \ldots, X_{t_0}$, whose law is the same under $H_0$ and $H_1$, so $\Prob_{H_1}\{\tau > t_0\} \geq 1 - \alpha$.
Since $(\tau - t_0)_+ \geq 1$ on this event, $D \geq 1 - \alpha > 0$.
Let $d := \lceil 2D/(1 - \alpha) \rceil \geq 2$, so that $D \geq (1 - \alpha)(d - 1)/2$ and
\[
    d < \bar{d}(t) := \frac{1 - \alpha}{2 I} \ln \frac{t}{\alpha} + 1~.
\]
Since $\bar{d}(t)$ grows logarithmically, $\bar{d}(t) \leq t$ for all $t$ beyond a threshold that depends only on $\alpha$ and $I$, and we consider such $t$.
For every $t_0 \in \{t, \ldots, 2t - d\}$, Markov's inequality gives $\Prob_{H_1}\{\tau > t_0 + d\} \leq D/d \leq (1 - \alpha)/2$, hence
\[
    \Prob_{H_1}\left\{t_0 < \tau \leq t_0 + d\right\} = \Prob_{H_1}\left\{\tau > t_0\right\} - \Prob_{H_1}\left\{\tau > t_0 + d\right\} \geq \frac{1 - \alpha}{2}~.
\]
Lemma~\ref{lem:windows} with $q = (1 - \alpha)/2$, together with $\lfloor t/d \rfloor \geq t/(2d) \geq t/(2\bar{d}(t))$, gives $d I \geq ((1 - \alpha)/2) \ln(t/(2 \alpha \bar{d}(t))) - \ln 2$.
Substituting into $D \geq (1 - \alpha)(d - 1)/2$ yields
\[
    D \geq \frac{\left(1 - \alpha\right)^2}{4 I} \ln \frac{t}{\alpha} - \frac{\left(1 - \alpha\right)^2}{4 I} \ln\left(2 \bar{d}(t)\right) - \frac{\left(1 - \alpha\right) \ln 2}{2 I} - \frac{1 - \alpha}{2}~.
\]
Since $\bar{d}(t) = \sO(\ln t)$ with constants that depend only on $\alpha$ and $I$, the last three terms are $\sO(\ln \ln t)$, which proves the theorem.
\hfill$\qed$

\paragraph{Comparison With the Signal Strength.}
We verify the bound $\Delta^2 \leq \KL(Q \| P)/2$ used in Section~\ref{sec:guarantees} for detectable shifts on $\sG$.
Let $\sD_0 \in \sG$ with $\Delta > 0$.
By the band~\eqref{eq:band}, $|\E_Q[\hat{p} - F(X) \mid \sD_0]| \leq \E_Q[\epsilon(\hat{p}) \mid \sD_0]$, and $\E_P[F(X)] = 1/2$ since $F$ is continuous.
Hence
\[
    \Delta \leq \abs{\E_Q\left[F\left(X\right)\right] - \frac{1}{2}} = \abs{\E_Q\left[F\left(X\right)\right] - \E_P\left[F\left(X\right)\right]} \leq \sup_{A}\ \abs{Q(A) - P(A)} \leq \sqrt{\frac{\KL(Q \| P)}{2}},
\]
where the supremum is over measurable sets, the second inequality uses that $F$ takes values in $[0, 1]$, and the last one is Pinsker's inequality.
Since $\Delta > 0$, squaring gives the bound.

\section{Efficient Implementation}
\label{app:implementation}

This appendix details the computation summarized in Section~\ref{sec:implementation}.
Appendix~\ref{app:overview} explains the main idea and states its guarantee, Appendices~\ref{app:direct}--\ref{app:dc} give the direct and the divide-and-conquer methods in full, and the remaining subsections prove the guarantee, discuss floating-point accuracy, and compare the two methods empirically.

\subsection{Overview}
\label{app:overview}

We show that the allocations of Algorithm~\ref{alg:test} can be computed \emph{exactly} in $\sO(T^2 \ln T)$ time over $T$ rounds, instead of the $\sO(T^3)$ time of a direct computation.
For a function $f$ on $\simplex$, write
\[
    \Phi(f) := \left(\int_{\simplex} w^i f(\bw) \mu(d\bw)\right)_{i \in \{0, +, -\}} \in \R^3~.
\]
Since $\bw \in \simplex$, the components of $\Phi(B_t)$ sum to $\int_{\simplex} B_t(\bw) \mu(d\bw)$, so the allocation~\eqref{eq:restarting-allocation} depends on the data only through $\Phi(B_t)$.

\paragraph{Direct Computation.}
The polynomial $B_t$ has degree $t - 1$ in $\bw$ and thus $t(t+1)/2$ monomial coefficients.
Since the moments of the Dirichlet prior have closed forms, both the update $B_{t+1} = \inner{\bw}{\bx_t} B_t + \pi_{t+1}$ and the vector $\Phi(B_t)$ take $\sO(t^2)$ operations on these coefficients (Appendix~\ref{app:direct}), which adds up to $\sO(T^3)$ time and $\sO(T^2)$ memory.
The bottleneck is that the full polynomial $B_t$ is updated and integrated at \emph{every} round $t$.
We instead group the total horizon $T$ into blocks and handle the contributions of many start times $s$ to many rounds $t$ at once, which requires a different way of storing polynomials than by their monomial coefficients.
By reusing shared computations within blocks, we avoid integrating $B_t$ separately at every round.

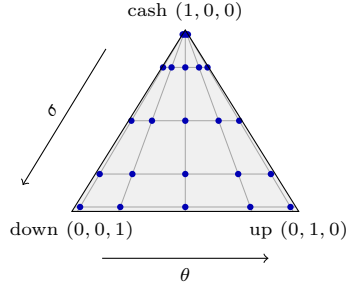
\begin{figure}[t]
    \centering
    \begin{tikzpicture}[every node/.style={font=\scriptsize}]
        \coordinate (C) at (0,2.4);
        \coordinate (D) at (-1.5,0);
        \coordinate (U) at (1.5,0);
        \fill[gray!12] (C) -- (D) -- (U) -- cycle;
        \foreach \j in {0,...,4} {
            \pgfmathsetmacro{\sg}{(1 + cos(180*(\j + 0.5)/5))/2}
            \draw[gray!70, very thin] ($(C)!\sg!(D)$) -- ($(C)!\sg!(U)$);
            \pgfmathsetmacro{\th}{(1 + cos(180*(\j + 0.5)/5))/2}
            \draw[gray!70, very thin] (C) -- ($(D)!\th!(U)$);
        }
        \foreach \j in {0,...,4} {
            \foreach \k in {0,...,4} {
                \pgfmathsetmacro{\sg}{(1 + cos(180*(\j + 0.5)/5))/2}
                \pgfmathsetmacro{\th}{(1 + cos(180*(\k + 0.5)/5))/2}
                \fill[blue!70!black] ($(C)!\sg!($(D)!\th!(U)$)$) circle (1.2pt);
            }
        }
        \draw (C) -- (D) -- (U) -- cycle;
        \node[above] at (C) {cash $(1, 0, 0)$};
        \node[below] at (D) {down $(0, 0, 1)$};
        \node[below] at (U) {up $(0, 1, 0)$};
        \draw[->] (-1.05,2.15) -- node[sloped, above] {$\sigma$} (-2.15,0.35);
        \draw[->] (-1.1,-0.62) -- node[below] {$\theta$} (1.1,-0.62);
    \end{tikzpicture}
    \caption{
        \textbf{The grid for $N = 5$.}
    }
    \label{fig:grid}
\end{figure}

\paragraph{Grid Representation.}
We first change variables. Write $\sigma := w^+ + w^-$ and $\theta := w^+ / \sigma$, so that a CRP can be written as (Figure~\ref{fig:grid})
\[
    \bw = \left(1 - \sigma, \sigma \theta, \sigma \left(1 - \theta\right)\right) \text{ with } \left(\sigma, \theta\right) \in [0, 1]^2~.
\]
Every pair $(\sigma, \theta)$ gives a valid CRP, and the simplex $\simplex$ is now mapped to the unit square $[0, 1]^2$, facilitating tensor representation.
Write $F_{s,t} := \prod_{i=s}^{t-1} \inner{\bw}{\bx_i}$, so that $B_t = \sum_{i=1}^{t} \pi_i F_{i,t}$.
Note that $F_{s,t}$ is a bivariate polynomial of degree at most $t - s$ in each of $\sigma$ and $\theta$.
We store such a polynomial by its values on the tensor grid of $N \times N$ Chebyshev points, with $N \geq t - s + 1$ larger than its degree~\citep{trefethen2013approximation}.
In the transformed coordinates, the integral $\int_{\simplex} w^i B_t(\bw) \mu(d\bw)$ becomes a weighted sum of $B_t$ values on the grid, using Gauss--Chebyshev quadrature in $\theta$ and an interpolatory rule in $\sigma$.
On its own, the grid representation does not reduce the cost of a single integration, since $B_t$ still has about $t^2$ grid values to update at every round, but as we will show, it enables faster evaluation of $\int_{\simplex} w^i B_t(\bw) \mu(d\bw)$ once many rounds $t$ are processed together. 

\paragraph{Divide And Conquer.}

\begin{figure}[t]
    \centering
    \begin{tikzpicture}[x=0.36cm, y=0.36cm, every node/.style={font=\scriptsize}]
        % (a) pairs (s, t) with s <= t, as L lower-triangular array
        \begin{scope}
            \fill[gray!20] (0,0) -- (0,-8) -- (8,-8) -- cycle;
            \fill[blue!50] (0,-8) rectangle (4,-4);
            \fill[blue!30] (0,-4) rectangle (2,-2);
            \fill[blue!30] (4,-8) rectangle (6,-6);
            \foreach \i in {0,2,4,6} {
                \fill[blue!15] (\i,-\i-2) rectangle (\i+1,-\i-1);
            }
            \draw (0,0) -- (0,-8) -- (8,-8) -- cycle;
            \draw[densely dotted] (4,0.4) -- (4,-4);
            \draw[densely dotted] (-0.4,-4) -- (4,-4);
            \node[white] at (2,-6) {merge};
            \node[above] at (4,0.3) {$K$};
            \node[left] at (-0.3,-4) {$K$};
            \draw[->] (0,1.8) -- node[above] {start round $s$} (8,1.8);
            \draw[->] (-1.8,0) -- node[above, rotate=90] {round $t$} (-1.8,-8);
            \node at (4,-9.4) {(a)};
        \end{scope}
        % (b) merging at the midpoint and moving forward
        \begin{scope}[xshift=4.0cm]
            \foreach \s/\h in {0.5/2.2, 1.5/1.65, 2.5/1.1, 3.5/0.55} {
                \draw (\s,-3.5+\h) -- (3.6,-3.5+\h) -- (4,-3.5);
                \fill (\s,-3.5+\h) circle (2.5pt);
            }
            \draw[line width=1.6pt, ->] (4,-3.5) -- (8.6,-3.5);
            \foreach \t in {4.6,5.6,6.6,7.6} {
                \draw[fill=white] (\t,-3.5) circle (3pt);
            }
            \node[below] at (4,-3.7) {merge};
            \draw (0,-5.6) -- (8,-5.6);
            \foreach \x/\l in {0/$L$, 4/$K$, 8/$R$} {
                \draw (\x,-5.4) -- (\x,-5.8);
                \node[below] at (\x,-5.8) {\l};
            }
            % \draw[decorate, decoration={brace, amplitude=3pt, mirror}] (4,-7.0) -- node[below=2pt] {$\times F_{M,M'}$ once} (6,-7.0);
            \node at (4,-9.4) {(b)};
        \end{scope}
        % (c) epochs of doubling length, no horizon needed
        \begin{scope}[yshift=-4.6cm, x=0.42cm]
            \foreach \l/\r/\j/\f in {1/2/0/15, 2/4/1/30, 4/8/2/15, 8/16/3/30} {
                \fill[blue!\f] (\l-1,0) rectangle (\r-1,0.8);
                \draw (\l-1,0) rectangle (\r-1,0.8);
                \node at ({(\l+\r)/2-1},0.4) {$\j$};
            }
            \draw[dashed] (15,0.4) -- (16.2,0.4);
            \foreach \x in {1,2,4,8,16} {
                \node[below] at (\x-1,0) {$\x$};
            }
            \foreach \x in {2,4,8} {
                \draw[->] (\x-1,1.7) -- (\x-1,0.9);
            }
            \node[right] at (7.2,1.45) {merge all earlier portfolios};
            \node at (7.5,-1.4) {(c)};
        \end{scope}
    \end{tikzpicture}
    \caption{
        (a) The pairs of start rounds $s$ and rounds $t \geq s$. The split at $K$ handles the dark block at once, later splits handle the lighter blocks, and the gray cells are computed directly.
        (b) The portfolios started in $[L, K)$ are merged at round $K$ into one polynomial, whose integrals are needed at every round in $[K, R)$ (white dots).
        (c) Without a known horizon, doubling trick is used, and all earlier portfolios are merged at the start of each epoch.
    }
    \label{fig:recursion-demo}
\end{figure}
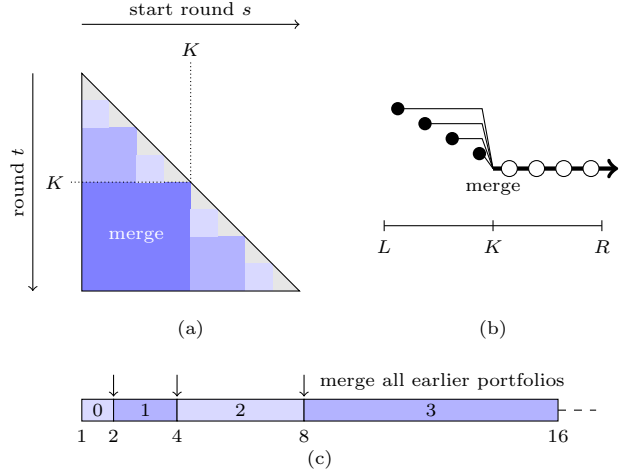

Since $B_t = \sum_{i=1}^{t} \pi_i F_{i,t}$, we need $\Phi(B_t)$ for all $t \leq T$, with terms $F_{s,t}$ for all $1 \leq s \leq t \leq T$. 
These can be illustrated by a triangle of pairs $s \leq t$ (Figure~\ref{fig:recursion-demo}(a)).
The diagonal is explicit, $\Phi(\pi_t F_{t,t}) = \pi_t (1/3, 1/3, 1/3)$, so it remains to sum $\pi_s \Phi(F_{s,t})$ over the pairs $s < t$.
We recursively bisect the time horizon, forming a binary partition of the time indices.  
For simplicity, we assume that $T$ is known in advance and is a power of two:
\[
    [1,T+1) \
    \longrightarrow
    [1,T/2+1) \cup [T/2+1,T+1)
    \longrightarrow
    [1,T/4+1) \cup 
    \cdots~.
\]
Consider a block of rounds $[L, R) := \{L, \ldots, R - 1\}$ and its midpoint $K := \lfloor (L + R)/2 \rfloor$.
The pairs within each half are two smaller copies of the same problem, which we solve recursively, starting from the block $[1, T + 1)$ of all rounds.
For a pair $(s, t)$ that crosses the midpoint, with $s \in [L, K)$ and $t \in [K, R)$, all portfolios started in the first half grow by the same factor $F_{K,t}$ after round $K$, so
\[
        \sum_{i=L}^{K-1} \pi_i F_{i,t} = A_{L,K} F_{K,t}
        \text{ with } A_{L,K} := \sum_{i=L}^{K-1} \pi_i F_{i,K}~.
\]
Given $A_{L,K}$, computing the crossing contributions requires applying the \emph{single fixed} linear functional $f \mapsto \Phi(A_{L,K} f)$ to the products $F_{K,t}$ for $t \in [K, R)$.
With our earlier grid representation, this step can be performed in $\sO((R-L)^2 \ln (R-L))$ with discrete cosine transforms.
This step handles the darkest block in Figure~\ref{fig:recursion-demo}(a) at once.
Recursing on both halves handles the remaining pairs, i.e., the two triangles, and the call on $[L, K)$ also returns $A_{L,K}$.

Summing the computational cost over the blocks yields the following guarantee.

\begin{proposition}
    \label{prop:fast}
    The divide-and-conquer method computes each allocation $\bw_t$ in~\eqref{eq:restarting-allocation} from $\bx_1, \ldots, \bx_{t-1}$ only, exactly in exact arithmetic and without knowing the horizon.
    Its first $T$ rounds cost $\sO(T^2 \ln T)$ operations and $\sO(T^2)$ memory.
\end{proposition}

\paragraph{Running Online.}
For simplicity, we have described the recursion for a known horizon $T$.
Every step uses only the price relatives of earlier rounds, so the recursion can run online in a straightforward way.
The horizon can be removed by intervals $[2^J, 2^{J+1})$ of doubling length, at the start of which all earlier portfolios are merged into one polynomial (Figure~\ref{fig:recursion-demo}(c)) and will be reused afterwards.
The rest of this appendix gives this online schedule, the proof of Proposition~\ref{prop:fast}, and a discussion of floating-point accuracy.

\paragraph{Notation For The Proofs.}
The rest of this appendix uses the notation above: the moment vector $\Phi = (\Phi^0, \Phi^+, \Phi^-)$, the growth $F_{s,t}$ for $1 \leq s \leq t$, the blocks $[L, R) := \{L, \ldots, R - 1\}$, and the merged growth
\[
    A_{L,R} := \sum_{i=L}^{R-1} \pi_i F_{i,R} \quad \text{for } 1 \leq L \leq R,
\]
so that $A_{L,L} = 0$.
Unrolling the recursion of Proposition~\ref{prop:restarting} gives $B_t = \sum_{i=1}^{t} \pi_i F_{i,t}$, and since $w^0 + w^+ + w^- = 1$ on $\simplex$, the allocation~\eqref{eq:restarting-allocation} is
\[
    \bw_t = \frac{\left(\Phi^0(B_t) + r_t, \Phi^+(B_t), \Phi^-(B_t)\right)}{\Phi^0(B_t) + \Phi^+(B_t) + \Phi^-(B_t) + r_t}~.
\]
Both methods therefore compute $\Phi(B_1), \Phi(B_2), \ldots$.
For $L \leq K \leq R$ and $t \geq K$, the definitions give the identities
\begin{equation}
    \label{eq:merge}
    F_{L,R} = F_{L,K} F_{K,R}, \quad A_{L,R} = A_{L,K} F_{K,R} + A_{K,R}, \quad \text{and} \quad \sum_{i=L}^{K-1} \pi_i F_{i,t} = A_{L,K} F_{K,t},
\end{equation}
as well as $A_{L,t+1} = \inner{\bw}{\bx_t} (A_{L,t} + \pi_t)$ for $t \geq L$.

\subsection{The Direct Method}
\label{app:direct}

For a multi-index $\bk = (k_0, k_+, k_-)$ of nonnegative integers, let $|\bk| := k_0 + k_+ + k_-$ and $\bw^{\bk} := (w^0)^{k_0} (w^+)^{k_+} (w^-)^{k_-}$, and let $\be_i$ be the multi-index with a one in position $i \in \{0, +, -\}$ and zeros elsewhere.
Since $w^0 + w^+ + w^- = 1$ on $\simplex$, the constant $\pi_{t+1}$ in the recursion $B_{t+1}(\bw) = \inner{\bw}{\bx_t} B_t(\bw) + \pi_{t+1}$ may be replaced by $\pi_{t+1} (w^0 + w^+ + w^-)^t$.
The recursion then produces homogeneous polynomials $B_t(\bw) = \sum_{|\bk| = t - 1} \lambda_{t, \bk} \bw^{\bk}$ of degree $t - 1$ that agree with $B_t$ on $\simplex$.
We store their coefficients weighted by the moments of the prior, $D_{t, \bk} := \lambda_{t, \bk} \E_{\mu}[\bw^{\bk}]$.

\begin{proposition}
    \label{prop:exact}
    For every $t \geq 1$ and $i \in \{0, +, -\}$,
    \[
        \Phi^i\left(B_t\right) = \sum_{|\bk| = t - 1} \frac{k_i + 1/2}{t + 1/2} D_{t, \bk}~.
    \]
    Moreover, $D_{1, \boldsymbol{0}} = \pi_1$, and for every $\bk$ with $|\bk| = t$,
    \[
        D_{t+1, \bk} = \frac{1}{t + 1/2} \sum_{i \in \{0, +, -\} : k_i \geq 1} x_t^i \left(k_i - \frac{1}{2}\right) D_{t, \bk - \be_i} + \pi_{t+1} \frac{\Gamma(3/2) t!}{\Gamma(t + 3/2)} \prod_{i \in \{0, +, -\}} \binom{2 k_i}{k_i} 4^{-k_i}~.
    \]
\end{proposition}

\begin{proof}
For $\mu = \Dirichlet(1/2, 1/2, 1/2)$, the moments are
\[
    \E_{\mu}\left[\bw^{\bk}\right] = \frac{\Gamma(3/2)}{\Gamma(|\bk| + 3/2)} \prod_{i \in \{0, +, -\}} \frac{\Gamma(k_i + 1/2)}{\Gamma(1/2)},
\]
so that $\E_{\mu}[\bw^{\bk + \be_i}] = \E_{\mu}[\bw^{\bk}] (k_i + 1/2)/(|\bk| + 3/2)$ for every $i$.
With $|\bk| = t - 1$, this gives
\[
    \Phi^i\left(B_t\right) = \sum_{|\bk| = t - 1} \lambda_{t, \bk} \E_{\mu}\left[\bw^{\bk + \be_i}\right] = \sum_{|\bk| = t - 1} \frac{k_i + 1/2}{t + 1/2} D_{t, \bk}~.
\]
For the recursion, the coefficient of $\bw^{\bk}$ with $|\bk| = t$ in
\[
    \inner{\bw}{\bx_t} B_t(\bw) = \sum_{|\bk'| = t - 1} \lambda_{t, \bk'} \sum_{i \in \{0, +, -\}} x_t^i \bw^{\bk' + \be_i}
\]
is $\sum_{i : k_i \geq 1} x_t^i \lambda_{t, \bk - \be_i}$.
Multiplying it by $\E_{\mu}[\bw^{\bk}] = \E_{\mu}[\bw^{\bk - \be_i}] (k_i - 1/2)/(t + 1/2)$ gives the first term.
The multinomial theorem gives $\pi_{t+1} (w^0 + w^+ + w^-)^t = \pi_{t+1} \sum_{|\bk| = t} (t! / (k_0! k_+! k_-!)) \bw^{\bk}$, and multiplying each coefficient by $\E_{\mu}[\bw^{\bk}]$ gives the second term, since $\Gamma(k + 1/2) / (\Gamma(1/2) k!) = \binom{2k}{k} 4^{-k}$ for every integer $k \geq 0$.
\end{proof}

A homogeneous polynomial of degree $t - 1$ in three variables has $t(t+1)/2$ coefficients, so round $t$ costs $\sO(t^2)$ operations, and $T$ rounds cost $\sO(T^3)$ time and $\sO(T^2)$ memory.
In floating point, we store $D_{t, \cdot}$ divided by its sum, together with the logarithm of this sum, which avoids overflow.
Since all terms are nonnegative, the computation involves no cancellation, and each round adds a relative error of the order of the machine precision to every coefficient.

\subsection{The Grid Representation}
\label{app:grid}

\paragraph{Coordinates.}
For $\bw \in \simplex$ with $w^+ + w^- > 0$, let $\sigma := w^+ + w^-$ and $\theta := w^+ / \sigma$, so that $\bw = (1 - \sigma, \sigma \theta, \sigma (1 - \theta))$ with $(\sigma, \theta) \in [0, 1]^2$; the excluded cash corner has measure zero under $\mu$.
The density of $\mu$ with respect to the Lebesgue measure in $(w^+, w^-)$ is $(w^0 w^+ w^-)^{-1/2}/(2\pi)$, and the map $(\sigma, \theta) \mapsto (\sigma \theta, \sigma (1 - \theta))$ has Jacobian $\sigma$, so for every $\mu$-integrable function $f$,
\[
    \int_{\simplex} f\left(\bw\right) \mu\left(d\bw\right) = \int_0^1 \int_0^1 f\left(1 - \sigma, \sigma \theta, \sigma \left(1 - \theta\right)\right) \frac{d\theta}{\pi \sqrt{\theta \left(1 - \theta\right)}} \frac{d\sigma}{2 \sqrt{1 - \sigma}}~.
\]
The two factors are the densities of $\Beta(1/2, 1/2)$ and $\Beta(1, 1/2)$, so $\theta$ and $\sigma$ are independent under $\mu$ with these laws.
Since $x_u^0 = 1$, the return $\inner{\bw}{\bx_u} = 1 + \sigma (\theta (x_u^+ - 1) + (1 - \theta) (x_u^- - 1))$ has degree at most one in each of $\sigma$ and $\theta$, and so do $w^0 = 1 - \sigma$, $w^+ = \sigma \theta$, and $w^- = \sigma (1 - \theta)$.
For $N \geq 1$, let $\sP_N$ denote the polynomials in $(\sigma, \theta)$ of degree at most $N - 1$ in each variable, so that $f g \in \sP_{N + N' - 1}$ for all $f \in \sP_N$ and $g \in \sP_{N'}$.
In particular, $F_{s,t} \in \sP_{t - s + 1}$ and $A_{L,R} \in \sP_{R - L + 1}$.

\paragraph{Grids and Resampling.}
For $N \geq 1$, let $\xi_j^N := (1 + \cos(\pi (j + 1/2)/N))/2$ for $j = 0, \ldots, N - 1$ be the Chebyshev points of the first kind on $[0, 1]$, and let $V_N(f) := (f(\xi_j^N, \xi_k^N))_{j, k = 0}^{N - 1} \in \R^{N \times N}$ collect the values of a function $f$ on the tensor grid.
Since the points are distinct, $V_N$ is a bijection from $\sP_N$ to $\R^{N \times N}$.
Let $T_m$ denote the Chebyshev polynomials, defined by $T_m(\cos \phi) = \cos(m \phi)$.
In one variable, a polynomial $g(z) = \sum_{m=0}^{N-1} a_m T_m(2z - 1)$ satisfies $g(\xi_j^N) = \sum_{m=0}^{N-1} a_m \cos(\pi m (j + 1/2)/N)$, so its values and its coefficients are related by a discrete cosine transform, which the fast Fourier transform computes in $\sO(N \ln N)$ operations~\citep{trefethen2013approximation}.
For $N, N' \geq 1$, the \emph{resampling} $S_{N \to N'} : \R^{N \times N} \to \R^{N' \times N'}$ applies, along each axis, the transform that maps $N$ values to $N$ coefficients, keeps the first $\min\ \{N, N'\}$ coefficients and pads them with zeros to $N'$ coefficients, and evaluates the result at the $N'$ points.

\paragraph{Quadrature.}
Let $q^{\theta}_k := 1/N$ for all $k$, which are the weights of the Gauss--Chebyshev rule for the arcsine law $\Beta(1/2, 1/2)$ at the points $\xi_k^N$, and let
\[
    q^{\sigma}_j := \frac{1}{N} \left(1 + 2 \sum_{m=1}^{N-1} \frac{(-1)^{m+1}}{4m^2 - 1} \cos\left(\frac{\pi m (j + 1/2)}{N}\right)\right)
\]
be the weights of the interpolatory rule for $\Beta(1, 1/2)$ at the same points.
These weights follow from the discrete orthogonality of the vectors $(\cos(\pi m (j + 1/2)/N))_{j=0}^{N-1}$ for $m = 0, \ldots, N - 1$, together with the moments $\E[T_m(2\sigma - 1)] = (-1)^{m+1}/(4m^2 - 1)$ for $\sigma \sim \Beta(1, 1/2)$, which the substitution $2\sigma - 1 = \cos \phi$ turns into $\frac{1}{2} \int_0^{\pi} \cos(m \phi) \cos(\phi/2) d\phi$.
Since $\sum_{m=1}^{N-1} 1/(4m^2 - 1) = (1 - 1/(2N - 1))/2$, every weight satisfies $q^{\sigma}_j \geq 1/(N(2N - 1)) > 0$.
For $i \in \{0, +, -\}$, define the \emph{moment weights} $\Omega^{(i)}_N := (q^{\sigma}_j q^{\theta}_k w^i(\xi_j^N, \xi_k^N))_{j,k = 0}^{N-1}$, where $w^i$ is written in the coordinates $(\sigma, \theta)$; they are nonnegative.
We also write $\inner{\cdot}{\cdot}$ for the entrywise inner product of arrays, and $\odot$ for their entrywise product.

\begin{lemma}[Grid operations]
    \label{lem:grid}
    Let $N, N' \geq 1$ and $N_{\max} := \max\ \{N, N'\}$.
    \begin{enumerate}
        \item[(a)] The resampling $S_{N \to N'}$ costs $\sO(N_{\max}^2 \ln N_{\max})$ operations, and $S_{N \to N'} V_N(f) = V_{N'}(f)$ for every $f \in \sP_{\min\{N, N'\}}$.
        \item[(b)] If $N \geq 2$, then $\Phi^i(f) = \inner{\Omega^{(i)}_N}{V_N(f)}$ for every $f \in \sP_{N - 1}$ and every $i \in \{0, +, -\}$.
        \item[(c)] If $N' \leq N$, then $\inner{\omega}{V_N(f)} = \inner{(N/N')^2 S_{N \to N'} \omega}{V_{N'}(f)}$ for every $\omega \in \R^{N \times N}$ and every $f \in \sP_{N'}$.
    \end{enumerate}
\end{lemma}

\begin{proof}
(a) Each one-dimensional transform costs $\sO(N_{\max} \ln N_{\max})$ operations, and each axis requires at most $N_{\max}$ of them.
For $f \in \sP_{\min\{N, N'\}}$, the Chebyshev coefficients of $f$ along each axis vanish from the index $\min\ \{N, N'\}$ on, so truncating or padding them does not change $f$, and the last step evaluates $f$ itself.

(b) The product $w^i f$ lies in $\sP_N$.
The interpolatory rule with $N$ points is exact for polynomials of degree at most $N - 1$, and the Gauss--Chebyshev rule with $N$ points is exact for polynomials of degree at most $2N - 1$.
Since $\mu$ is the product of the two laws in the coordinates $(\sigma, \theta)$, the tensor rule is exact on $\sP_N$, which is spanned by products of univariate polynomials, so $\Phi^i(f) = \sum_{j=0}^{N-1} \sum_{k=0}^{N-1} q^{\sigma}_j q^{\theta}_k (w^i f)(\xi_j^N, \xi_k^N)$.

(c) By (a), $V_N(f) = S_{N' \to N} V_{N'}(f)$, so $\inner{\omega}{V_N(f)} = \inner{S_{N' \to N}^{\top} \omega}{V_{N'}(f)}$, where $\top$ denotes the adjoint with respect to the entrywise inner product.
Let $C_N$ be the orthonormal type-II discrete cosine transform of size $N$, and let $E_{N' \to N} \in \R^{N \times N'}$ pad a vector with zeros, so that $E_{N' \to N}^{\top} = E_{N \to N'}$ truncates.
Along one axis, $S_{N' \to N}$ acts as $(N/N')^{1/2} C_N^{\top} E_{N' \to N} C_{N'}$, and $S_{N \to N'}$ acts as $(N'/N)^{1/2} C_{N'}^{\top} E_{N \to N'} C_N$, where the factors account for the normalization of the orthonormal transforms.
Hence the adjoint of the first map is $(N/N')^{1/2} C_{N'}^{\top} E_{N \to N'} C_N$, which is $N/N'$ times the second map.
Two axes give the factor $(N/N')^2$, so $S_{N' \to N}^{\top} = (N/N')^2 S_{N \to N'}$.
\end{proof}

\paragraph{Functionals.}
We store a linear functional $\psi$ on polynomials by weights $\omega \in \R^{N \times N}$, and we say that $\omega$ \emph{represents} $\psi$ on $\sP_{N'}$, for some $N' \leq N$, if $\psi(f) = \inner{\omega}{V_N(f)}$ for every $f \in \sP_{N'}$.
Weights can be formed and transformed exactly in three ways.
First, by Lemma~\ref{lem:grid}(c), the weights $(N/N')^2 S_{N \to N'} \omega$ on the grid with $N'$ points also represent $\psi$ on $\sP_{N'}$, and we call them the \emph{restriction} of $\omega$ to $N'$ points.
Second, if $\omega$ represents $\psi$ on $\sP_N$ and $g \in \sP_{N''}$ for some $N'' \leq N$, then $\omega \odot V_N(g)$ represents the functional $f \mapsto \psi(g f)$ on $\sP_{N - N'' + 1}$, because $g f \in \sP_N$ for such $f$.
Third, by Lemma~\ref{lem:grid}(b), if $a \in \sP_{N_a}$ for some $N_a < N$, then $\Omega^{(i)}_N \odot V_N(a)$ represents the functional $f \mapsto \Phi^i(a f)$ on $\sP_{N - N_a}$, because $a f \in \sP_{N - 1}$ for such $f$.
We write $\Phi(a \cdot)$ for the triple of these functionals, $i \in \{0, +, -\}$, and $\psi(g \cdot)$ for the functional $f \mapsto \psi(g f)$.

\subsection{The Divide-and-Conquer Method}
\label{app:dc}

Algorithm~\ref{alg:dc} gives the method in a form that runs online, without knowledge of the horizon.
It keeps an accumulator $\Psi_t \in \R^3$ for every round $t$, initialized to the contribution $\pi_t \Phi(F_{t,t}) = \pi_t (1/3, 1/3, 1/3)$ of the portfolio started at round $t$, since $\E_{\mu}[w^i] = 1/3$.
The procedure $\textsc{Solve}(L, R)$ adds the contributions $\pi_s \Phi(F_{s,t})$ of all pairs $L \leq s < t < R$, as in Appendix~\ref{app:overview}.
The procedure $\textsc{Prefix}(\psi, L, R)$ applies a triple $\psi = (\psi^0, \psi^+, \psi^-)$ of linear functionals to the products $F_{L,t}$ for $t \in [L, R)$ and adds the results to the accumulators; it evaluates the functional $\psi = \Phi(A_{L,K} \cdot)$ of the crossing pairs in Appendix~\ref{app:overview}.
The procedure $\textsc{Epochs}$ removes the horizon: epoch $J = 0, 1, 2, \ldots$ consists of the rounds $[2^J, 2^{J+1})$, and at its start, all portfolios started before round $2^J$ are merged into $A_{1, 2^J}$.
Blocks of at most $m_0$ rounds, for a fixed integer $m_0 \geq 1$, are \emph{leaves}, which are handled round by round.

\paragraph{Representation.}
In a call on a block $[L, R)$ of $m := R - L$ rounds, the polynomials $P$ and $A$ are stored by their values on the grid with $m + 1$ points per axis, and products and sums are formed on this grid after resampling the operands.
In $\textsc{Solve}$, the weights of $\Phi(A_1 \cdot)$ are formed on the grid with $m + 1$ points as $\Omega^{(i)}_{m+1} \odot V_{m+1}(A_1)$, and they are restricted to $R - K$ points before the call of $\textsc{Prefix}$.
In $\textsc{Prefix}$, the weights $\omega$ of $\psi$ lie on the grid with $m$ points; the first sub-call receives their restriction to $K - L$ points, and the second sub-call receives the restriction of $\omega \odot V_m(P_1)$ to $R - K$ points.
In a leaf of $\textsc{Prefix}$, $\psi(P)$ is evaluated on the grid of $\psi$, after resampling $P$ to it.
In $\textsc{Epochs}$, the weights of $\Phi(A \cdot)$ at the start of epoch $J$ are formed on the grid with $2^{J+1}$ points and restricted to $2^J$ points, and the update of $A$ at the end of epoch $J$ is formed on the grid with $2^{J+1}$ points.
In practice, grid sizes may be rounded up to lengths for which the fast transforms are efficient, which preserves exactness.

\paragraph{Scheduling.}
The procedures run as coroutines.
The statement ``wait for $\bx_t$'' suspends the current call until $\bx_t$ has been observed, and returns at once if it has.
The statement ``run concurrently'' starts two calls, which are then resumed independently, and continues once both have returned.
After observing $\bx_t$, the main loop resumes every suspended call, in any order, until each call waits for $\bx_{t+1}$ or has returned.
The calls share no state except the accumulators, to which they only add, so the order does not matter.
When all price relatives are known in advance, every wait returns at once, and the same procedures compute $\Phi(B_1), \ldots, \Phi(B_T)$ offline, as in Appendix~\ref{app:overview}.

\begin{algorithm}[t]
    \caption{Online computation of $\Phi(B_t)$ and $\bw_t$ without a known horizon}
    \label{alg:dc}
    \begin{algorithmic}[1]
        \Procedure{Solve}{$L, R$} \Comment{adds $\pi_s \Phi(F_{s,t})$ to $\Psi_t$ for all $L \leq s < t < R$; returns $(F_{L,R}, A_{L,R})$}
            \If{$R - L \leq m_0$}
                \State $P \gets 1$ and $A \gets 0$
                \For{$t = L, \ldots, R - 1$}
                    \State $\Psi_t \gets \Psi_t + \Phi(A)$ and wait for $\bx_t$
                    \State $A \gets \inner{\bw}{\bx_t} (A + \pi_t)$ and $P \gets \inner{\bw}{\bx_t} P$
                \EndFor
                \State \Return $(P, A)$
            \EndIf
            \State $K \gets \lfloor (L + R)/2 \rfloor$
            \State $(P_1, A_1) \gets \Call{Solve}{L, K}$
            \State Run $(P_2, A_2) \gets \Call{Solve}{K, R}$ and \Call{Prefix}{$\Phi(A_1 \cdot), K, R$} concurrently \Comment{crossing pairs}
            \State \Return $(P_1 P_2, A_1 P_2 + A_2)$
        \EndProcedure
        \Procedure{Prefix}{$\psi, L, R$} \Comment{adds $\psi(F_{L,t})$ to $\Psi_t$ for all $L \leq t < R$; returns $F_{L,R}$}
            \If{$R - L \leq m_0$}
                \State $P \gets 1$
                \For{$t = L, \ldots, R - 1$}
                    \State $\Psi_t \gets \Psi_t + \psi(P)$ and wait for $\bx_t$
                    \State $P \gets \inner{\bw}{\bx_t} P$
                \EndFor
                \State \Return $P$
            \EndIf
            \State $K \gets \lfloor (L + R)/2 \rfloor$
            \State $P_1 \gets \Call{Prefix}{\psi, L, K}$
            \State $P_2 \gets \Call{Prefix}{\psi(P_1 \cdot), K, R}$
            \State \Return $P_1 P_2$
        \EndProcedure
        \Procedure{Epochs}{} 
            \State $A \gets 0$ \Comment{$A = A_{1, 2^J}$ at the start of epoch $J$}
            \For{$J = 0, 1, 2, \ldots$}
                \State Run $(P, A') \gets \Call{Solve}{2^J, 2^{J+1}}$ and \Call{Prefix}{$\Phi(A \cdot), 2^J, 2^{J+1}$} concurrently
                \State $A \gets A P + A'$
            \EndFor
        \EndProcedure
        \Statex
        \State Set $\Psi_t \gets \pi_t (1/3, 1/3, 1/3)$ for all $t \geq 1$, and run \Call{Epochs}{} until it waits for $\bx_1$
        \For{$t = 1, 2, \ldots$}
            \State Compute $\bw_t$ from $\Psi_t$ and $r_t$ as in~\eqref{eq:restarting-allocation}, and invest with $\bw_t$
            \State Observe $\bx_t$, and resume every suspended call until it waits for $\bx_{t+1}$
        \EndFor
    \end{algorithmic}
\end{algorithm}

The first lemma shows that the procedures compute the right quantities, regardless of the schedule.

\begin{lemma}
    \label{lem:dc}
    Let $1 \leq L < R$.
    \begin{enumerate}
        \item[(a)] If the weights of $\psi$ represent $\psi$ on $\sP_{R - L}$, then \textup{\textsc{Prefix}}$(\psi, L, R)$ adds $\psi(F_{L,t})$ to $\Psi_t$ for every $t \in [L, R)$ and returns $F_{L,R}$.
        \item[(b)] \textup{\textsc{Solve}}$(L, R)$ adds $\pi_s \Phi(F_{s,t})$ to $\Psi_t$ for every pair $L \leq s < t < R$ and returns $(F_{L,R}, A_{L,R})$.
        \item[(c)] For every $t \geq 1$, the calls started by \textup{\textsc{Epochs}} add $\sum_{i=1}^{t-1} \pi_i \Phi(F_{i,t})$ to $\Psi_t$ in total, so that $\Psi_t$ reaches $\Phi(B_t)$.
    \end{enumerate}
\end{lemma}

\begin{proof}
We prove (a) and (b) by induction on $R - L$, and we first treat $\textsc{Prefix}$.
In a leaf, $P = F_{L,t}$ when $\psi(P)$ is added to $\Psi_t$, because $F_{L,L} = 1$ and $F_{L,t+1} = \inner{\bw}{\bx_t} F_{L,t}$, and at the end, $P = F_{L,R}$.
Since $F_{L,t} \in \sP_{R - L}$ for $t < R$, the weights of $\psi$ evaluate $\psi(F_{L,t})$ exactly.
Otherwise, $L < K < R$.
The first sub-call receives weights that represent $\psi$ on $\sP_{K - L}$, so by induction it adds $\psi(F_{L,t})$ for $t \in [L, K)$ and returns $P_1 = F_{L,K} \in \sP_{K - L + 1}$.
For $t \in [K, R)$, we have $\psi(F_{L,t}) = \psi(P_1 F_{K,t})$ by~\eqref{eq:merge}, and the weights of the functional $\psi(P_1 \cdot)$ represent it on $\sP_{R - K}$ by the second operation on weights, with $N = R - L$ and $N'' = K - L + 1$.
By induction, the second sub-call adds these values for $t \in [K, R)$ and returns $F_{K,R}$, and $P_1 P_2 = F_{L,R}$ by~\eqref{eq:merge}.

For $\textsc{Solve}$, a leaf has $A = A_{L,t}$ when $\Phi(A)$ is added to $\Psi_t$, because $A_{L,L} = 0$ and $A_{L,t+1} = \inner{\bw}{\bx_t} (A_{L,t} + \pi_t)$.
Since $A_{L,t} \in \sP_{R - L}$ for $t < R$, Lemma~\ref{lem:grid}(b) with $N = R - L + 1$ evaluates $\Phi(A_{L,t}) = \sum_{i=L}^{t-1} \pi_i \Phi(F_{i,t})$ exactly, and the leaf returns $(F_{L,R}, A_{L,R})$.
Otherwise, by induction, the two calls of $\textsc{Solve}$ add the pairs with $L \leq s < t < K$ and with $K \leq s < t < R$, and return $(F_{L,K}, A_{L,K})$ and $(F_{K,R}, A_{K,R})$.
The remaining pairs have $L \leq s < K \leq t < R$, and for each such $t$, their contributions sum to $\Phi(A_{L,K} F_{K,t})$ by~\eqref{eq:merge}.
The weights of $\Phi(A_{L,K} \cdot)$ represent it on $\sP_{R - K}$ by the third operation on weights, with $N = R - L + 1$ and $N_a = K - L + 1$, and so does their restriction to $R - K$ points; by (a), the call of $\textsc{Prefix}$ adds these contributions.
The returned values follow from~\eqref{eq:merge}.
All products and sums are exact, since every polynomial formed in a call on $m$ rounds lies in $\sP_{m + 1}$.

For (c), induction on $J$ shows that $A = A_{1, 2^J}$ at the start of epoch $J$: initially $A_{1,1} = 0$, and $A_{1, 2^{J+1}} = A_{1, 2^J} F_{2^J, 2^{J+1}} + A_{2^J, 2^{J+1}}$ by~\eqref{eq:merge}.
The weights of $\Phi(A_{1, 2^J} \cdot)$, formed on the grid with $2^{J+1}$ points, represent it on $\sP_{2^J}$ by the third operation on weights, since $A_{1, 2^J} \in \sP_{2^J}$.
Every round $t \geq 1$ lies in exactly one epoch, the one with $2^J \leq t < 2^{J+1}$.
In this epoch, the call of $\textsc{Solve}$ adds the pairs with $2^J \leq s < t$ by (b), and the call of $\textsc{Prefix}$ adds $\Phi(A_{1, 2^J} F_{2^J, t}) = \sum_{i=1}^{2^J - 1} \pi_i \Phi(F_{i,t})$ by (a) and~\eqref{eq:merge}.
Calls of other epochs add nothing to $\Psi_t$.
Together with the initial value $\pi_t \Phi(F_{t,t})$, this gives $\Psi_t = \sum_{i=1}^{t} \pi_i \Phi(F_{i,t}) = \Phi(B_t)$.
\end{proof}

The second lemma shows that the schedule never needs a future price relative.

\begin{lemma}
    \label{lem:causal}
    In Algorithm~\ref{alg:dc}, every call on a block $[L, R)$ is started before $\bw_L$ is computed, and after $\bx_{L-1}$ has been observed if $L \geq 2$, and its inputs depend only on $\bx_1, \ldots, \bx_{L-1}$.
    For every $t \in [L, R)$, the call completes its contribution to $\Psi_t$ before it waits for $\bx_t$, using only $\bx_1, \ldots, \bx_{t-1}$.
    Consequently, $\Psi_t = \Phi(B_t)$ when $\bw_t$ is computed, and $\bw_t$ depends only on $\bx_1, \ldots, \bx_{t-1}$.
\end{lemma}

\begin{proof}
A call on $[L, R)$ waits for $\bx_L, \ldots, \bx_{R-1}$ in this order and returns right after $\bx_{R-1}$ has been observed.
We show the first two claims by induction over the calls, in the order in which they start.
The calls of epoch $0$ start before round $1$ with the input $\Phi(0 \cdot)$.
The calls of epoch $J + 1$ start as soon as those of epoch $J$ return, that is, right after $\bx_{2^{J+1} - 1}$ has been observed and before $\bw_{2^{J+1}}$ is computed, and their input depends on $A_{1, 2^{J+1}}$, which depends only on $\bx_1, \ldots, \bx_{2^{J+1} - 1}$.
Within $\textsc{Solve}(L, R)$, the first sub-call starts with its parent, and the other two start as soon as the first returns, that is, right after $\bx_{K-1}$ has been observed, with the input $\Phi(A_{L,K} \cdot)$, which depends only on $\bx_1, \ldots, \bx_{K-1}$.
The same holds within $\textsc{Prefix}(\psi, L, R)$, whose second sub-call receives $\psi(F_{L,K} \cdot)$.
In a leaf, the contribution to $\Psi_t$ is added right before the wait for $\bx_t$, and it depends on the inputs and on $\bx_L, \ldots, \bx_{t-1}$; a call that is not a leaf contributes only through its sub-calls.
This proves the first two claims.

When $\bw_t$ is computed, $\bx_t$ has not been observed, and every call has been resumed until it waits for $\bx_t$ or has returned.
By the first claim, every call whose block contains $t$ has been started, and by the second claim, it has completed its contribution to $\Psi_t$.
No call adds to $\Psi_t$ later, since every contribution to $\Psi_t$ is added before a wait for $\bx_t$.
Hence $\Psi_t$ has its final value, which is $\Phi(B_t)$ by Lemma~\ref{lem:dc}(c), and it depends only on $\bx_1, \ldots, \bx_{t-1}$.
\end{proof}

\subsection{Proof of Proposition~\ref{prop:fast}}

Lemmas~\ref{lem:dc} and~\ref{lem:causal} show that Algorithm~\ref{alg:dc} computes $\bw_t$ from $\Psi_t = \Phi(B_t)$ and $r_t$, using only $\bx_1, \ldots, \bx_{t-1}$, exactly in exact arithmetic and without the horizon.
Running two calls concurrently changes the order of the operations but not the operations themselves, so the cost is the same as for a sequential run.

\paragraph{Time.}
Consider a call on a block of $m$ rounds that is not a leaf.
Apart from its sub-calls, it performs a constant number of resamplings, restrictions, entrywise products, and sums, all on grids with at most $m + 1$ points per axis, which cost $\sO(m^2 \ln m)$ by Lemma~\ref{lem:grid}.
A leaf has at most $m_0$ rounds, so it costs $\sO(1)$.
Let $C_{\mathrm{P}}(m)$ and $C_{\mathrm{S}}(m)$ be the largest costs of $\textsc{Prefix}$ and $\textsc{Solve}$ on blocks of $m$ rounds.
Then, for $m > m_0$,
\[
    C_{\mathrm{P}}(m) \leq C_{\mathrm{P}}\left(\lfloor m/2 \rfloor\right) + C_{\mathrm{P}}\left(\lceil m/2 \rceil\right) + \sO\left(m^2 \ln m\right)~.
\]
At depth $d$ of the recursion of $\textsc{Prefix}$, there are at most $2^d$ calls, each on at most $\lceil m/2^d \rceil \leq 2m/2^d$ rounds, so the calls that are not leaves cost $\sO(2^d (m/2^d)^2 \ln m) = \sO(m^2 \ln m / 2^d)$ in total, and the at most $m$ leaves cost $\sO(m)$.
Summing over $d$ gives $C_{\mathrm{P}}(m) = \sO(m^2 \ln m)$.
Hence
\[
    C_{\mathrm{S}}(m) \leq C_{\mathrm{S}}\left(\lfloor m/2 \rfloor\right) + C_{\mathrm{S}}\left(\lceil m/2 \rceil\right) + C_{\mathrm{P}}\left(\lceil m/2 \rceil\right) + \sO\left(m^2 \ln m\right) = C_{\mathrm{S}}\left(\lfloor m/2 \rfloor\right) + C_{\mathrm{S}}\left(\lceil m/2 \rceil\right) + \sO\left(m^2 \ln m\right),
\]
and the same argument gives $C_{\mathrm{S}}(m) = \sO(m^2 \ln m)$.
Epoch $J \geq 1$ consists of one call of each procedure on $2^J$ rounds, together with the construction of $\Phi(A \cdot)$ and the update of $A$ on grids with $2^{J+1}$ points, so it costs $\sO(4^J J)$.
The first $T$ rounds involve only the epochs with $2^J \leq T$, and the work done in an unfinished epoch is at most the cost of the whole epoch.
Summing the geometric series over $J \leq \log_2 T$ gives a total time of $\sO(T^2 \ln T)$.

\paragraph{Memory.}
At any time, the active calls are those whose blocks contain the current round.
They form one chain of nested calls of $\textsc{Solve}$ in the current epoch, together with, for each of these calls whose second half contains the current round, one chain of nested calls of $\textsc{Prefix}$ on that second half, and the call of $\textsc{Prefix}$ of the epoch with its own chain.
A call on $m$ rounds stores $\sO(m^2)$ numbers, namely its inputs, the values returned by its first sub-call, and its current polynomials, and the block sizes decrease geometrically along each chain and across the chains.
Together with $A$, which has $\sO(T^2)$ grid values, and the accumulators, the memory is $\sO(T^2)$.
\hfill$\qed$

\paragraph{Uneven Work.}
The work is not spread evenly over the rounds.
A call performs its own work at its start, at its midpoint, and right after its last price relative has been observed, and each epoch starts with the construction of $\Phi(A \cdot)$ on its largest grid.
A single round $t$ can therefore require $\sO(t^2 \ln t)$ operations, while the first $T$ rounds take $\sO(T \ln T)$ operations per round on average, compared with $\sO(t^2)$ operations at every round $t$ for the direct method.

\subsection{Numerical Accuracy}
\label{app:accuracy}

In floating point, every grid array is stored as a matrix of values together with a common logarithmic scale, which avoids overflow.
The values of $F_{s,t}$ and $A_{L,R}$ on the grids are nonnegative, and so are the moment weights $\Omega^{(i)}_N$, since both one-dimensional rules have positive weights.
Entrywise products and sums of nonnegative arrays are accurate to machine precision in every entry.
Resampling and restriction, in contrast, mix all entries of an array through orthogonal cosine transforms, so they are accurate only relative to the size of the whole array, and an entry far below the largest one can lose its relative accuracy.
The grid values of $F_{s,t}$ are wealths of CRPs, and they spread over many orders of magnitude when the rounds from $s$ to $t$ carry strong evidence for one direction of the shift, but also during a long null phase, in which the tax shrinks the wealth of the risky portfolios while cash keeps its wealth.
In both cases, an entry that loses accuracy belongs to a portfolio far behind the best one, and it matters only if later rounds lift it by the order of $30$ nats relative to the best one.
We expect this to require strong evidence for a shift, and thus to happen typically long after $M_t$ has crossed $1/\alpha$, but this is a heuristic rather than a guarantee.
In any case, projecting every computed allocation onto $\simplex$ keeps the test valid by Proposition~\ref{prop:validity-2}, so numerical errors can reduce the power of the detector but not its false-alarm control.
Running the method a second time on slightly larger grids, which is exact in exact arithmetic but rounds differently, gives an empirical estimate of the error.
Appendix~\ref{app:benchmark} tests this account against the direct method.

\subsection{Comparison With the Direct Method}
\label{app:benchmark}

We compare the divide-and-conquer method with the direct method of Appendix~\ref{app:direct}, first on the computed allocations and then on cost.
All streams use $n = 1000$ reference scores from $\sN(0, 1)$, the DKW band at level $\delta = 0.1$, and $\alpha = 0.05$.

\paragraph{Agreement.}
Both methods first match the definition~\eqref{eq:restarting-wealth} evaluated literally, with each component integrated separately by a tensor Gauss rule that is exact for its degree.
For $T \in \{16, 32, 64\}$ and a shift to $\sN(1, 1)$ after round $\lfloor 2T/3 \rfloor$, the allocations of both methods differ from the literal ones by at most $7.8 \cdot 10^{-15}$.
We then compare the two methods on four streams of $T \in \{500, 1000, 2000, 4000, 8000\}$ rounds: a null stream; a shift to $\sN(1, 1)$ after round $\lfloor 2T/3 \rfloor$; a drift with mean $1.5 (t - \lfloor T/3 \rfloor)_+/100$; and a reversal, with $\sN(1, 1)$ on the rounds $\lfloor T/3 \rfloor + 1, \ldots, \lfloor 2T/3 \rfloor$ and $\sN(-1, 1)$ afterwards.
On the first three streams, the allocations agree to within $1.6 \cdot 10^{-15}$ in every round and $\ln M_t$ to within $3.5 \cdot 10^{-13}$ (Table~\ref{tab:implementation}).
The reversal stream is the difficult case of Appendix~\ref{app:accuracy}: it first collects strong evidence for an upward shift, between $25$ and $644$ nats in $\ln M_t$ by the reversal, and then evidence for a downward one.
Both methods raise the alarm in the same round, and their allocations agree to within $10^{-9}$ until $15$ to $28$ rounds after the reversal.
From then on, the divide-and-conquer allocations are inaccurate, and for $T \geq 2000$ they even leave $\simplex$.
The error estimate of Appendix~\ref{app:accuracy} detects this: it first exceeds $10^{-9}$ at most three rounds after the allocations first differ by $10^{-9}$.
Since the test stops at the alarm, which comes $142$ to $2627$ rounds before this point, the detector itself is not affected.
A monitor that keeps betting after an alarm can project the allocations onto $\simplex$ (Appendix~\ref{app:accuracy}), which keeps the test valid, although no longer exact.

\begin{table}[htbp]
\centering
\caption{\textbf{Agreement of the direct and the divide-and-conquer methods.} Left: largest difference $\max_{t \leq T} \max_{i}\ \abs{w_t^i - \tilde{w}_t^i}$ between the allocations $\bw_t$ and $\tilde{\bw}_t$ of the two methods. Right, for the reversal stream: the alarm time $\tau$ (the same for both methods), the round $\lfloor 2T/3 \rfloor$ of the reversal, $\ln M$ at the reversal, and the first round at which the allocations differ by more than $10^{-9}$.}
\label{tab:implementation}
\begin{tabular}{rccc@{\hspace{2.5em}}rrrr}
\toprule
 & \multicolumn{3}{c}{largest allocation difference} & \multicolumn{4}{c}{reversal stream} \\
\cmidrule(lr){2-4} \cmidrule(lr){5-8}
$T$ & null & shift & drift & $\tau$ & reversal & $\ln M$ there & first difference \\
\midrule
$500$ & $3.3 \cdot 10^{-16}$ & $5.6 \cdot 10^{-16}$ & $1.3 \cdot 10^{-15}$ & $218$ & $333$ & $25$ & $360$ \\
$1000$ & $3.3 \cdot 10^{-16}$ & $1.2 \cdot 10^{-15}$ & $9.4 \cdot 10^{-16}$ & $401$ & $666$ & $84$ & $694$ \\
$2000$ & $2.2 \cdot 10^{-16}$ & $7.4 \cdot 10^{-16}$ & $1.0 \cdot 10^{-15}$ & $712$ & $1333$ & $165$ & $1350$ \\
$4000$ & $3.3 \cdot 10^{-16}$ & $1.6 \cdot 10^{-15}$ & $7.8 \cdot 10^{-16}$ & $1371$ & $2666$ & $330$ & $2687$ \\
$8000$ & $3.3 \cdot 10^{-16}$ & $1.1 \cdot 10^{-15}$ & $5.8 \cdot 10^{-16}$ & $2721$ & $5333$ & $644$ & $5348$ \\
\bottomrule
\end{tabular}
\end{table}

\begin{figure}[htbp]
    \centering
    \includegraphics[width=\textwidth]{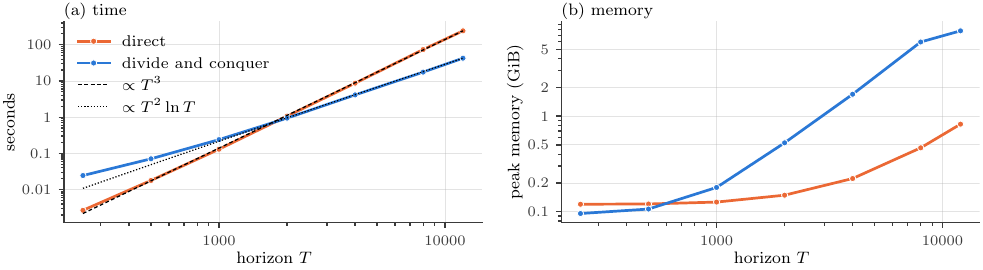}
    \caption{\textbf{Cost of the direct and the divide-and-conquer methods} on one core, against the horizon $T$: (a) wall-clock time, with reference lines through the last point of each method, and (b) peak memory of the process, including about $0.1$~GiB for the Python interpreter.}
    \label{fig:app-runtime}
\end{figure}

\paragraph{Cost.}
Figure~\ref{fig:app-runtime} reports the wall-clock time and the peak memory of both methods on the shift stream, on one core of an Apple M1 Max.
Every run starts a fresh process, and for $T \leq 2000$ we report the fastest of three runs.
The direct method is a compiled loop, and the divide-and-conquer method uses the fast cosine transforms of SciPy, with blocks of at most $m_0 = 32$ rounds as leaves.
The times follow the predicted orders: from $T = 8000$ to $T = 12{,}000$, the direct method slows down by a factor of $3.3$, against $1.5^3 \approx 3.4$, and the divide-and-conquer method by $2.4$, against $1.5^2 \ln(12{,}000) / \ln(8000) \approx 2.35$.
The divide-and-conquer method is faster from $T = 2000$ on, and the gap grows with $T$: at $T = 12{,}000$, it takes $43$ seconds against $242$, a speedup of $5.7$.
The price is memory.
Both methods need $\sO(T^2)$ memory, but the constant of the divide-and-conquer method is much larger: its peak is $7.8$~GiB at $T = 12{,}000$, against $0.8$~GiB for the direct method, which stores only the coefficients of $B_t$.
On a machine with $32$~GiB of memory, this limits a single run of the divide-and-conquer method to roughly $20{,}000$ rounds.

\clearpage
\section{Details of the Experiments}
\label{app:experiments}

This appendix gives the complete setup of the experiments in Section~\ref{sec:experiments}, the results for every configuration, and further checks.
Throughout, the restarting universal portfolio uses the Dirichlet prior $\Dirichlet(1/2, 1/2, 1/2)$ and the heavy-tailed allocation~\eqref{eq:heavy-tailed}, and its allocations are computed exactly with the direct method of Appendix~\ref{app:direct}.
It rejects when its test martingale $M_t$ reaches $1/\alpha$, and we also record its portfolio wealth $\Wealth_t \leq M_t$.

\subsection{Synthetic Experiments}
\label{app:exp-synthetic}

\paragraph{Data.}
The reference dataset has $n$ i.i.d.\ draws from $\sN(0, 1)$, and the test scores satisfy $X_t \sim \sN(0, 1)$ for $t \leq t_0$.
After the shift, they follow one of two alternatives.
\begin{itemize}
    \item \emph{Sharp change point}: $X_t \sim \sN(d, 1)$ with $d \in \{1, 1.5, 2\}$, and $n = 1000$.
    \item \emph{Gradual drift}: $X_t \sim \sN(\lambda (t - t_0)/100, 1)$ with $\lambda \in \{1.5, 3, 5\}$, and $n = 2000$. The per-round rates $0.015$, $0.03$, and $0.05$ are those of~\citet{shaer2026testing}.
\end{itemize}
The shift time runs over $t_0 \in \{0, 200, 400, \ldots, 4000\}$, every stream has length $T = 4500$, and every configuration is repeated $100$ times.
Within a trial, all magnitudes (or rates) share the same draws, so they differ only after $t_0$.

\paragraph{Relation To The Experiments Of \citet{shaer2026testing}.}
The three synthetic experiments of~\citet{shaer2026testing} are slices of this design, with their own random draws.
Their delayed change point is $d = 2$ at $t_0 \in \{200, 600, 4000\}$, and we use their seed for each trial for every $t_0 > 0$.
Their immediate change point (horizon $1000$) and their drift (horizon $100$) are the slices $t_0 = 0$; there, we use their seeds, draw their reference dataset and their test points, and continue the stream to $T = 4500$ with an independent generator.
On all of these slices, both CTM implementations reproduce the authors' code exactly, including the randomized conformal $p$-values and the warm-up.
The median delays read off our results agree with those in Figure~3 of~\citet{shaer2026testing} within three rounds, with one exception: for the CCTM at $t_0 = 4000$, the published curve crosses $1/2$ about eight rounds later than ours.
Since the code paths agree exactly, we attribute this difference to the random draws.

\paragraph{Methods And Hyperparameters.}
All methods use $\alpha = 0.05$ and the Dvoretzky--Kiefer--Wolfowitz band at level $\delta = 0.1$.
\begin{itemize}
    \item \textbf{Restarting universal portfolio} (Algorithm~\ref{alg:test}): no further parameters and no warm-up.
    \item \textbf{CCTM}~\citep{shaer2026testing}: ONS on the smoothed bet, with bets in $[-1/2, 1/2]$, clipping threshold $C = 0.1$, and smoothing constant $k = 10^{-6}$.
    \item \textbf{Standard CTM}: randomized conformal $p$-values against the growing pool, with the same ONS settings.
    \item \textbf{Oracle}: cash up to $t_0$, then the constant bet that maximizes the wealth over the rounds $t_0 + 1, \ldots, T$. It knows $t_0$ and the future.
\end{itemize}
The two CTMs get a warm-up of $\min\ \{100, t_0\}$ rounds, in which ONS runs but no bet is placed.
This is the warm-up of $100$ rounds that~\citet{shaer2026testing} use for their delayed change point and the absence of warm-up in their other two experiments, and it never uses shifted data.

\paragraph{Metrics.}
For each configuration, we report the following quantities.
\begin{itemize}
    \item The \emph{false-alarm rate} $\Prob\{\tau \leq t_0\}$.
    \item The \emph{power} given no false alarm, $\Prob\{\tau \leq T \mid \tau > t_0\}$, whose complement Theorem~\ref{thm:delay-full} bounds.
    \item The \emph{median delay} of $\tau - t_0$ over the trials without a false alarm, counting a missed detection as $\infty$.
    \item The \emph{mean delay} $\E[\min\ \{\tau - t_0, T - t_0 + 1\} \mid \tau > t_0]$, which counts a missed detection as $T - t_0 + 1$. No synthetic trial misses, so this is the plain mean.
\end{itemize}

\paragraph{Computation.}
The restarting portfolio processes each null prefix once and copies its exact state for the three magnitudes (or rates).
At $t_0 \in \{0, 200, 600, 4000\}$, every method runs to $T$ and its whole wealth path is kept; at the other shift times, the restarting portfolio stops once both of its wealths have crossed $1/\alpha$.
The full run takes about $3.5$ hours on nine cores of an Apple M1 Max.

\begin{table*}[t]
    \centering
    \small
    \caption{
        Median detection delay $\tau - t_0$ in the synthetic experiments, over the trials without a false alarm; Ours is the restarting universal portfolio.
        \textbf{Bold}: the fastest method for each configuration and shift time excluding oracle (not implementable).
        Note that the oracle's delay depends very weakly on $t_0$ (here we only show it for $t_0 = 4000$).
        Tables~\ref{tab:sharp-1}--\ref{tab:drift-5} give further statistics.
    }
    \label{tab:exp-synthetic}
    \setlength{\tabcolsep}{5pt}
    \begin{tabular}{@{}lrrrrrrrrrr@{}}
        \toprule
        & \multicolumn{3}{c}{$t_0 = 0$} & \multicolumn{3}{c}{$t_0 = 600$} & \multicolumn{3}{c}{$t_0 = 4000$} & \\
        \cmidrule(lr){2-4} \cmidrule(lr){5-7} \cmidrule(lr){8-10}
        & Ours & CCTM & CTM & Ours & CCTM & CTM & Ours & CCTM & CTM & Oracle \\
        \midrule
        Sharp change, $d = 1$ & \textbf{22} & 24 & 31 & \textbf{48} & 73 & 84 & \textbf{54} & 216 & 227 & 11 \\
        Sharp change, $d = 1.5$ & 14 & \textbf{13} & 21 & \textbf{32} & 52 & 61 & \textbf{36} & 150 & 168 & 7 \\
        Sharp change, $d = 2$ & 12 & \textbf{11} & 17 & \textbf{26} & 43 & 51 & \textbf{30} & 122 & 141 & 6 \\
        \midrule
        Gradual drift, $\lambda = 1.5$ & 65 & \textbf{64} & 67 & \textbf{81} & 98 & 104 & \textbf{88} & 179 & 197 & 73 \\
        Gradual drift, $\lambda = 3$ & \textbf{41} & 43 & 47 & \textbf{56} & 70 & 77 & \textbf{60} & 142 & 161 & 42 \\
        Gradual drift, $\lambda = 5$ & \textbf{31} & 33 & 38 & \textbf{44} & 57 & 64 & \textbf{47} & 126 & 146 & 28 \\
        \bottomrule
    \end{tabular}
\end{table*}

\paragraph{Sharp Change Point.}
Table~\ref{tab:exp-synthetic} summarizes the median delays of all synthetic settings, and Figures~\ref{fig:app-synthetic-delay} and~\ref{fig:app-synthetic-power} and Tables~\ref{tab:sharp-1}--\ref{tab:sharp-2} give the results for the sharp change.
The delays of the CTMs grow steadily with the length of the null phase: least-squares fits through $t_0 \geq 200$ yield slopes of $43$, $30$, and $25$ rounds per $1000$ null rounds for the CCTM at $d = 1$, $1.5$, and $2$, respectively.
The delay of the restarting portfolio grows like $\ln t_0$: regressed on $\ln t_0$ for $t_0 \geq 200$, it gains $3.8$, $2.7$, and $2.3$ additional rounds whenever $t_0$ increases by a factor of $e$.
It has the shortest median delay among the implementable methods at every $t_0 \geq 200$ and every magnitude, and at $t_0 = 4000$ it is $4.0$ to $4.2$ times faster than the CCTM.
At $t_0 = 0$, it is $1.5$ rounds ahead of the CCTM for $d = 1$ and one round behind for $d = 1.5$ and $d = 2$.
The wealth paths in Figure~\ref{fig:app-wealth} show why the CTMs slow down.
At the shift, the median log-wealth of every method is close to zero, between $-1$ and $0$; the difference arises after the shift, when the ONS iterate needs many rounds to move away from zero (Section~\ref{sec:restarting}), and the standard CTM additionally absorbs post-shift scores into its pool.
At the end of the stream, $\ln M_T$ of the restarting portfolio is within $16$ nats of the oracle's, while that of the CCTM is between $87$ and $989$ nats lower than that of the restarting portfolio.
Before the shift, the restarting portfolio raises a false alarm in $0.2\%$ of the runs and the standard CTM, despite its warm-up, in $0.9\%$.

\paragraph{Gradual Drift.}
Figures~\ref{fig:app-synthetic-delay}, \ref{fig:app-synthetic-power}, and~\ref{fig:app-wealth} and Tables~\ref{tab:drift-1p5}--\ref{tab:drift-5} give the results.
The drift is harder for every method, because the evidence arrives slowly at first, but the picture is the same.
The CCTM's median delay grows by $21$ to $25$ rounds per $1000$ null rounds, and the restarting portfolio's by $2.2$ to $3.2$ additional rounds whenever $t_0$ increases by a factor of $e$.
The restarting portfolio is the fastest implementable method at every $t_0 \geq 400$, and at $t_0 = 4000$ it is $2.0$, $2.4$, and $2.7$ times faster than the CCTM for $\lambda = 1.5$, $3$, and $5$; at $t_0 \in \{0, 200\}$, it is within two rounds of the CCTM.
Here, the oracle is not uniformly faster.
The best constant bet over the whole post-shift window is large, because the drift eventually pushes every $\hat{p}_t$ to one, and this bet is too aggressive while the drift is still small.
The universal portfolio instead averages over all portfolios and increases its exposure as the evidence grows.
The final wealth of the standard CTM is capped at $10^{250}$ by the reference implementation, which explains its constant value of $\ln M_T = 575.6$ in the tables.

\paragraph{Null Streams.}
We run $100$ null streams of length $4500$ for each of $n = 1000$ and $n = 2000$, without any warm-up, as in the type-I experiment of~\citet{shaer2026testing}.
Neither the restarting portfolio nor the CCTM ever reaches $1/\alpha$, and the standard CTM does so on $2\%$ of the streams (Table~\ref{tab:type_i} and Figure~\ref{fig:app-null}).
After $4500$ null rounds, the median log-wealth of the restarting portfolio is $-0.9$ at $n = 1000$ and $-1.0$ at $n = 2000$, which reflects the logarithmic cost of betting from the first round; the CCTM, which also bets from the first round here, ends lower, at $-2.6$ and $-2.8$.

\begin{figure}[htbp]
    \centering
    \includegraphics[width=0.5\textwidth]{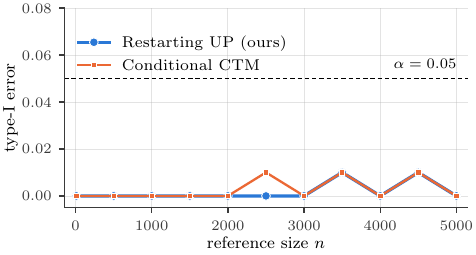}
    \caption{\textbf{Type-I error against the reference size $n$}: fraction of $100$ null streams of $T = 4500$ rounds on which the test martingale ($M_t$ for the restarting portfolio) reaches $1/\alpha$. The two lines coincide except at $n = 2500$.}
    \label{fig:app-type1-n}
\end{figure}

\begin{figure}[p]
    \centering
    \includegraphics[width=\textwidth]{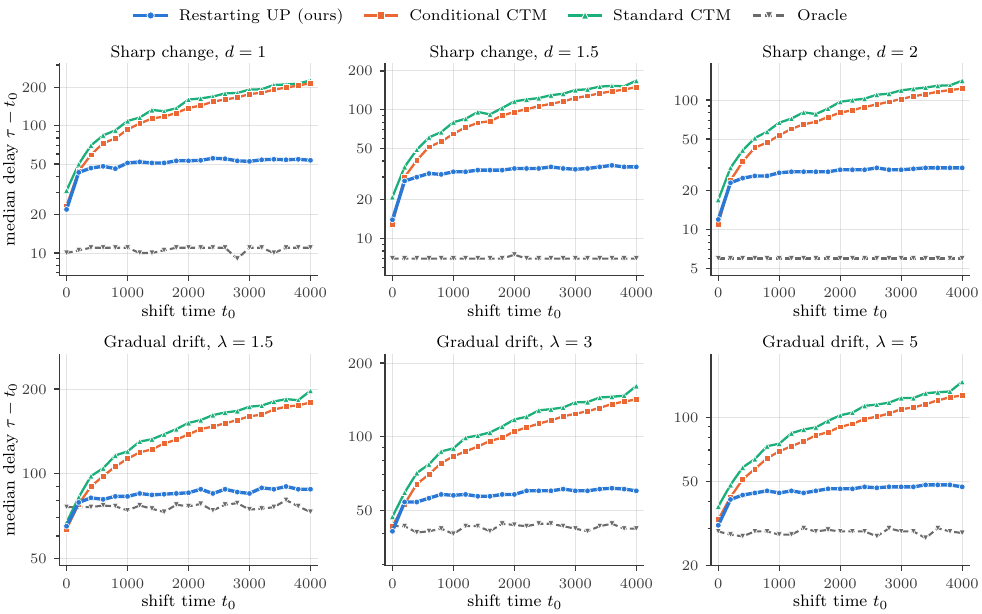}
    \caption{\textbf{Synthetic experiments: median delay $\tau - t_0$ against the shift time}, over the trials without a false alarm, for every sharp change (top) and every drift (bottom), on logarithmic vertical axes.}
    \label{fig:app-synthetic-delay}
\end{figure}

\begin{figure}[p]
    \centering
    \includegraphics[width=\textwidth]{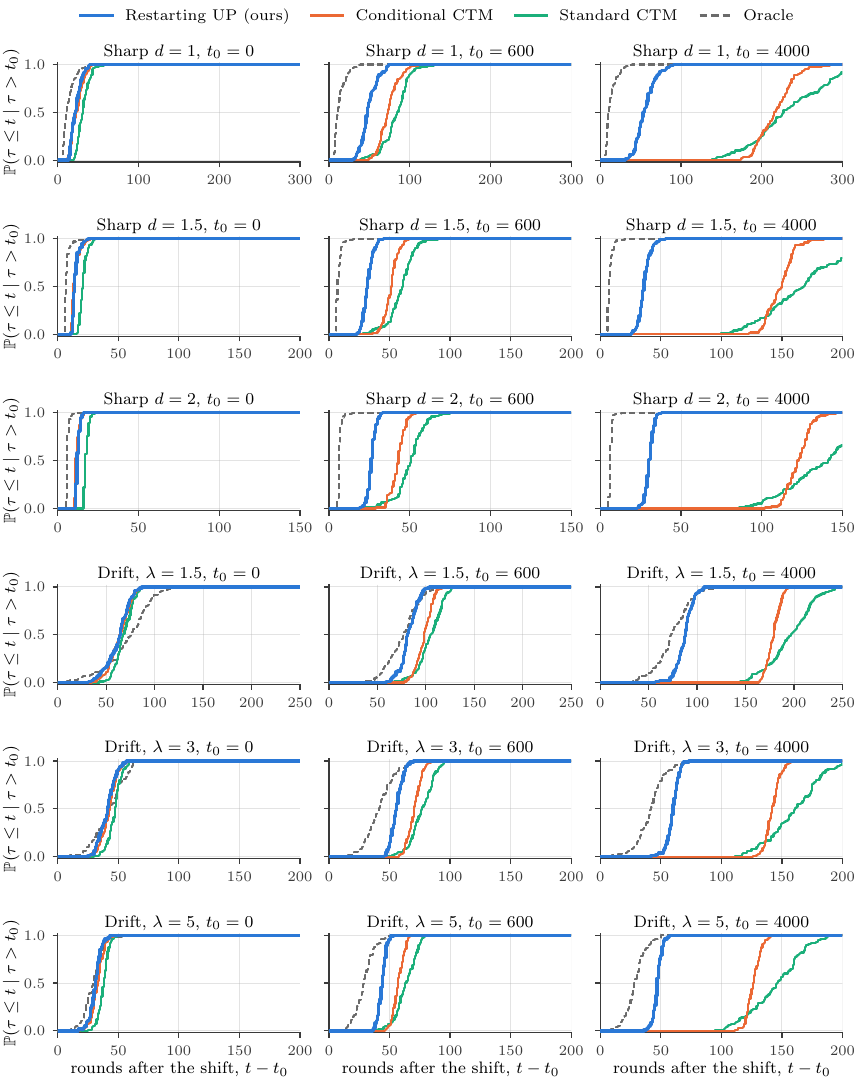}
    \caption{\textbf{Synthetic experiments: power given no false alarm}, $\Prob\{\tau \leq t \mid \tau > t_0\}$, against the number of rounds after the shift, for every configuration (rows) and $t_0 \in \{0, 600, 4000\}$ (columns).}
    \label{fig:app-synthetic-power}
\end{figure}

\begin{figure}[htbp]
    \centering
    \includegraphics[width=\textwidth]{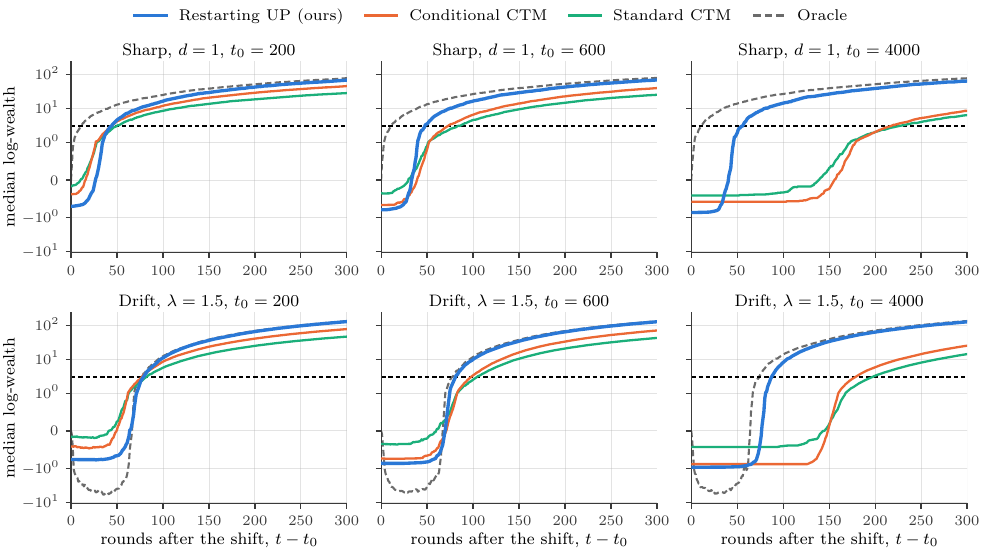}
    \caption{\textbf{Synthetic experiments: median log-wealth after the shift}, for the sharp change with $d = 1$ (top) and the drift with $\lambda = 1.5$ (bottom), and $t_0 \in \{200, 600, 4000\}$ (columns). For the restarting portfolio, the wealth is its test martingale $M_t$. The dashed line is the rejection threshold $\ln(1/\alpha)$, and the vertical axes are linear between $-1$ and $1$ and logarithmic outside.}
    \label{fig:app-wealth}
\end{figure}

\begin{figure}[htbp]
    \centering
    \includegraphics[width=\textwidth]{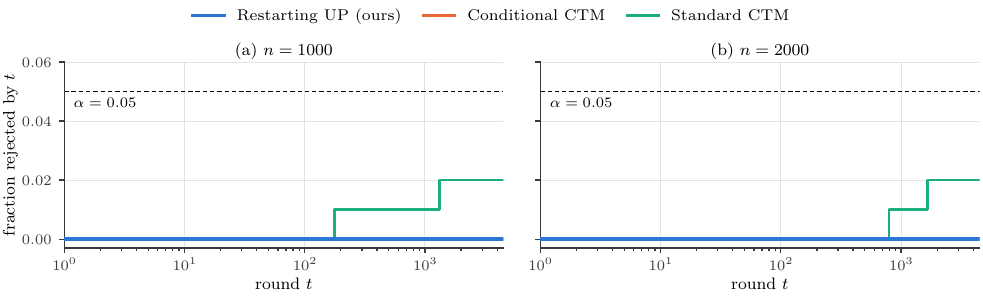}
    \caption{\textbf{Synthetic null streams}: fraction of the $100$ streams whose wealth ($M_t$ for the restarting portfolio) has reached $1/\alpha$ by round $t$, for (a) $n = 1000$ and (b) $n = 2000$. The CCTM never rejects, and its line is hidden under that of the restarting portfolio.}
    \label{fig:app-null}
\end{figure}

\paragraph{Type-I Error Against The Reference Size.}
Following Figure~2 (left) of~\citet{shaer2026testing}, we vary the reference size over $n \in \{10, 500, 1000, \ldots, 5000\}$ and run $100$ null streams of $T = 4500$ rounds for each $n$, drawn as in their code.
They use $T = 20{,}000$ rounds and $n = 10$ in place of $n = 0$; we keep the horizon of the other synthetic experiments, since $1100$ exact runs of $20{,}000$ rounds would take more than a day on our machine (Appendix~\ref{app:benchmark}).
The CTM settings are those above, without warm-up.
Figure~\ref{fig:app-type1-n} shows that both conditional tests reject on at most $1\%$ of the streams for every $n$, far below $\alpha = 0.05$, including $n = 10$, where the band has width $\epsilon \approx 0.39$.
Over all $1100$ streams, the restarting portfolio rejects twice, with both $M_t$ and $\Wealth_t$, and the CCTM three times.

\paragraph{Test Martingale Versus Portfolio Wealth.}
Rejecting on the portfolio wealth $\Wealth_t$ instead of $M_t$ delays detection by a median of $3$ rounds in both scenarios, with a mean of $6$ rounds for the sharp change and $8$ rounds for the drift.
The difference never has the other sign, since $M_t \geq \Wealth_t$ (Proposition~\ref{prop:validity-2}).

\begin{table}[htbp]
\centering
\small
\setlength{\tabcolsep}{3.5pt}
\begin{tabular}{rlrrrrrrrr}
\toprule
& & & & \multicolumn{2}{c}{delay after $t_0$} & \multicolumn{2}{c}{$\ln \Wealth_T$} & \multicolumn{2}{c}{$\ln M_T$} \\
\cmidrule(lr){5-6}\cmidrule(lr){7-8}\cmidrule(lr){9-10}
$t_0$ & strategy & false al. & power & median & mean & median & mean & median & mean \\
\midrule
$0$ & restarting UP & 0.00 & \textbf{1.00} & \textbf{22} & \textbf{22.6} & \textbf{1157.0} & \textbf{1159.2} & \textbf{1158.4} & \textbf{1160.7} \\
 & conditional CTM (ONS) & 0.00 & 1.00 & 24 & 24.0 & 747.8 & 747.5 & 747.8 & 747.5 \\
 & standard CTM & 0.00 & 1.00 & 31 & 31.6 & 171.8 & 172.0 & 171.8 & 172.0 \\
 & best CRP after shift (oracle) & 0.00 & 1.00 & 10 & 13.6 & 1164.0 & 1166.1 & 1164.0 & 1166.1 \\
\midrule
$200$ & restarting UP & 0.00 & \textbf{1.00} & \textbf{43} & 45.6 & \textbf{1088.2} & \textbf{1090.7} & \textbf{1090.9} & \textbf{1093.1} \\
 & conditional CTM (ONS) & 0.00 & 1.00 & 44 & \textbf{44.6} & 709.2 & 707.4 & 709.2 & 707.4 \\
 & standard CTM & 0.00 & 1.00 & 50 & 50.3 & 189.1 & 188.3 & 189.1 & 188.3 \\
 & best CRP after shift (oracle) & 0.00 & 1.00 & 10 & 11.8 & 1101.0 & 1103.7 & 1101.0 & 1103.7 \\
\midrule
$600$ & restarting UP & 0.00 & \textbf{1.00} & \textbf{48} & \textbf{49.3} & \textbf{996.2} & \textbf{997.3} & \textbf{998.5} & \textbf{999.7} \\
 & conditional CTM (ONS) & 0.00 & 1.00 & 73 & 73.9 & 637.2 & 638.5 & 637.2 & 638.5 \\
 & standard CTM & 0.01 & 1.00 & 84 & 83.3 & 208.6 & 209.0 & 208.6 & 209.0 \\
 & best CRP after shift (oracle) & 0.00 & 1.00 & 11 & 13.2 & 1011.2 & 1011.5 & 1011.2 & 1011.5 \\
\midrule
$4000$ & restarting UP & 0.00 & \textbf{1.00} & \textbf{54} & \textbf{55.8} & \textbf{112.3} & \textbf{112.4} & \textbf{114.3} & \textbf{114.4} \\
 & conditional CTM (ONS) & 0.00 & 1.00 & 216 & 217.4 & 27.0 & 27.3 & 27.0 & 27.3 \\
 & standard CTM & 0.01 & 1.00 & 227 & 233.0 & 19.0 & 18.3 & 19.0 & 18.3 \\
 & best CRP after shift (oracle) & 0.00 & 1.00 & 11 & 12.9 & 127.6 & 127.6 & 127.6 & 127.6 \\
\bottomrule
\end{tabular}
\caption{\textbf{Sharp change point, $d = 1$.} $X_t \sim \mathcal{N}(1, 1)$ after $t_0$. Reference set $n = 1000$, stream length $T = 4500$, 100 trials; details in Appendix~\ref{app:exp-synthetic}. \emph{False al.}: fraction rejecting at or before $t_0$; \emph{power}: fraction of the trials without a false alarm that reject by $T$, $\mathbb{P}(\tau\leq T\mid\tau>t_0)$. Delays are over the trials without a false alarm; the median counts misses as $\infty$, the mean counts them as $T-t_0+1$. The restarting portfolio rejects on its test martingale $M_t$; $\Wealth_T$ is its portfolio wealth, and $\Wealth_T = M_T$ for the others. \textbf{Bold}: best implementable strategy in the block.}
\label{tab:sharp-1}
\end{table}

\begin{table}[htbp]
\centering
\small
\setlength{\tabcolsep}{3.5pt}
\begin{tabular}{rlrrrrrrrr}
\toprule
& & & & \multicolumn{2}{c}{delay after $t_0$} & \multicolumn{2}{c}{$\ln \Wealth_T$} & \multicolumn{2}{c}{$\ln M_T$} \\
\cmidrule(lr){5-6}\cmidrule(lr){7-8}\cmidrule(lr){9-10}
$t_0$ & strategy & false al. & power & median & mean & median & mean & median & mean \\
\midrule
$0$ & restarting UP & 0.00 & \textbf{1.00} & 14 & 14.6 & \textbf{1905.2} & \textbf{1903.6} & \textbf{1905.7} & \textbf{1904.2} \\
 & conditional CTM (ONS) & 0.00 & 1.00 & \textbf{13} & \textbf{14.4} & 1113.4 & 1112.8 & 1113.4 & 1112.8 \\
 & standard CTM & 0.00 & 1.00 & 21 & 21.1 & 259.5 & 258.7 & 259.5 & 258.7 \\
 & best CRP after shift (oracle) & 0.00 & 1.00 & 7 & 7.5 & 1914.2 & 1912.5 & 1914.2 & 1912.5 \\
\midrule
$200$ & restarting UP & 0.00 & \textbf{1.00} & \textbf{28} & \textbf{28.4} & \textbf{1801.7} & \textbf{1801.3} & \textbf{1803.2} & \textbf{1802.9} \\
 & conditional CTM (ONS) & 0.00 & 1.00 & 30 & 29.7 & 1055.3 & 1054.4 & 1055.3 & 1054.4 \\
 & standard CTM & 0.00 & 1.00 & 36 & 35.5 & 278.5 & 278.6 & 278.5 & 278.6 \\
 & best CRP after shift (oracle) & 0.00 & 1.00 & 7 & 7.5 & 1817.0 & 1816.4 & 1817.0 & 1816.4 \\
\midrule
$600$ & restarting UP & 0.00 & \textbf{1.00} & \textbf{32} & \textbf{32.0} & \textbf{1636.3} & \textbf{1637.6} & \textbf{1638.0} & \textbf{1639.4} \\
 & conditional CTM (ONS) & 0.00 & 1.00 & 52 & 51.2 & 950.2 & 951.6 & 950.2 & 951.6 \\
 & standard CTM & 0.01 & 1.00 & 61 & 59.5 & 303.5 & 303.8 & 303.5 & 303.8 \\
 & best CRP after shift (oracle) & 0.00 & 1.00 & 7 & 7.8 & 1652.3 & 1654.0 & 1652.3 & 1654.0 \\
\midrule
$4000$ & restarting UP & 0.00 & \textbf{1.00} & \textbf{36} & \textbf{36.4} & \textbf{194.3} & \textbf{194.0} & \textbf{196.0} & \textbf{195.8} \\
 & conditional CTM (ONS) & 0.00 & 1.00 & 150 & 148.6 & 57.5 & 57.6 & 57.5 & 57.6 \\
 & standard CTM & 0.01 & 1.00 & 168 & 166.6 & 35.2 & 34.7 & 35.2 & 34.7 \\
 & best CRP after shift (oracle) & 0.00 & 1.00 & 7 & 7.9 & 211.1 & 210.8 & 211.1 & 210.8 \\
\bottomrule
\end{tabular}
\caption{\textbf{Sharp change point, $d = 1.5$.} $X_t \sim \mathcal{N}(1.5, 1)$ after $t_0$. Reference set $n = 1000$, stream length $T = 4500$, 100 trials; details in Appendix~\ref{app:exp-synthetic}. \emph{False al.}: fraction rejecting at or before $t_0$; \emph{power}: fraction of the trials without a false alarm that reject by $T$, $\mathbb{P}(\tau\leq T\mid\tau>t_0)$. Delays are over the trials without a false alarm; the median counts misses as $\infty$, the mean counts them as $T-t_0+1$. The restarting portfolio rejects on its test martingale $M_t$; $\Wealth_T$ is its portfolio wealth, and $\Wealth_T = M_T$ for the others. \textbf{Bold}: best implementable strategy in the block.}
\label{tab:sharp-1p5}
\end{table}

\begin{table}[htbp]
\centering
\small
\setlength{\tabcolsep}{3.5pt}
\begin{tabular}{rlrrrrrrrr}
\toprule
& & & & \multicolumn{2}{c}{delay after $t_0$} & \multicolumn{2}{c}{$\ln \Wealth_T$} & \multicolumn{2}{c}{$\ln M_T$} \\
\cmidrule(lr){5-6}\cmidrule(lr){7-8}\cmidrule(lr){9-10}
$t_0$ & strategy & false al. & power & median & mean & median & mean & median & mean \\
\midrule
$0$ & restarting UP & 0.00 & \textbf{1.00} & 12 & 12.4 & \textbf{2333.2} & \textbf{2333.5} & \textbf{2333.7} & \textbf{2333.9} \\
 & conditional CTM (ONS) & 0.00 & 1.00 & \textbf{11} & \textbf{11.8} & 1344.8 & 1343.7 & 1344.8 & 1343.7 \\
 & standard CTM & 0.00 & 1.00 & 17 & 17.6 & 315.5 & 315.0 & 315.5 & 315.0 \\
 & best CRP after shift (oracle) & 0.00 & 1.00 & 6 & 6.2 & 2342.5 & 2342.8 & 2342.5 & 2342.8 \\
\midrule
$200$ & restarting UP & 0.00 & \textbf{1.00} & \textbf{23} & \textbf{23.2} & \textbf{2221.1} & \textbf{2221.6} & \textbf{2222.5} & \textbf{2223.1} \\
 & conditional CTM (ONS) & 0.00 & 1.00 & 24 & 24.4 & 1279.6 & 1279.5 & 1279.6 & 1279.5 \\
 & standard CTM & 0.00 & 1.00 & 30 & 29.7 & 340.0 & 340.1 & 340.0 & 340.1 \\
 & best CRP after shift (oracle) & 0.00 & 1.00 & 6 & 6.2 & 2236.7 & 2237.1 & 2236.7 & 2237.1 \\
\midrule
$600$ & restarting UP & 0.00 & \textbf{1.00} & \textbf{26} & \textbf{26.4} & \textbf{2015.9} & \textbf{2015.9} & \textbf{2017.5} & \textbf{2017.5} \\
 & conditional CTM (ONS) & 0.00 & 1.00 & 43 & 42.6 & 1154.6 & 1154.5 & 1154.6 & 1154.5 \\
 & standard CTM & 0.01 & 1.00 & 51 & 50.2 & 368.1 & 368.5 & 368.1 & 368.5 \\
 & best CRP after shift (oracle) & 0.00 & 1.00 & 6 & 6.4 & 2031.1 & 2032.7 & 2031.1 & 2032.7 \\
\midrule
$4000$ & restarting UP & 0.00 & \textbf{1.00} & \textbf{30} & \textbf{30.4} & \textbf{243.3} & \textbf{242.9} & \textbf{245.1} & \textbf{244.6} \\
 & conditional CTM (ONS) & 0.00 & 1.00 & 122 & 122.7 & 82.7 & 82.5 & 82.7 & 82.5 \\
 & standard CTM & 0.01 & 1.00 & 141 & 140.0 & 48.4 & 47.9 & 48.4 & 47.9 \\
 & best CRP after shift (oracle) & 0.00 & 1.00 & 6 & 6.3 & 260.8 & 260.1 & 260.8 & 260.1 \\
\bottomrule
\end{tabular}
\caption{\textbf{Sharp change point, $d = 2$.} $X_t \sim \mathcal{N}(2, 1)$ after $t_0$. Reference set $n = 1000$, stream length $T = 4500$, 100 trials; details in Appendix~\ref{app:exp-synthetic}. \emph{False al.}: fraction rejecting at or before $t_0$; \emph{power}: fraction of the trials without a false alarm that reject by $T$, $\mathbb{P}(\tau\leq T\mid\tau>t_0)$. Delays are over the trials without a false alarm; the median counts misses as $\infty$, the mean counts them as $T-t_0+1$. The restarting portfolio rejects on its test martingale $M_t$; $\Wealth_T$ is its portfolio wealth, and $\Wealth_T = M_T$ for the others. \textbf{Bold}: best implementable strategy in the block.}
\label{tab:sharp-2}
\end{table}

\begin{table}[htbp]
\centering
\small
\setlength{\tabcolsep}{3.5pt}
\begin{tabular}{rlrrrrrrrr}
\toprule
& & & & \multicolumn{2}{c}{delay after $t_0$} & \multicolumn{2}{c}{$\ln \Wealth_T$} & \multicolumn{2}{c}{$\ln M_T$} \\
\cmidrule(lr){5-6}\cmidrule(lr){7-8}\cmidrule(lr){9-10}
$t_0$ & strategy & false al. & power & median & mean & median & mean & median & mean \\
\midrule
$0$ & restarting UP & 0.00 & \textbf{1.00} & 65 & \textbf{62.4} & \textbf{2814.1} & \textbf{2814.6} & \textbf{2815.0} & \textbf{2815.4} \\
 & conditional CTM (ONS) & 0.00 & 1.00 & \textbf{64} & 63.7 & 1634.8 & 1634.5 & 1634.8 & 1634.5 \\
 & standard CTM & 0.00 & 1.00 & 67 & 68.1 & 575.6 & 575.6 & 575.6 & 575.6 \\
 & best CRP after shift (oracle) & 0.00 & 1.00 & 76 & 72.4 & 2819.0 & 2818.6 & 2819.0 & 2818.6 \\
\midrule
$200$ & restarting UP & 0.00 & \textbf{1.00} & 79 & 78.5 & \textbf{2682.9} & \textbf{2682.5} & \textbf{2684.2} & \textbf{2683.8} \\
 & conditional CTM (ONS) & 0.00 & 1.00 & \textbf{78} & \textbf{77.4} & 1558.0 & 1558.0 & 1558.0 & 1558.0 \\
 & standard CTM & 0.00 & 1.00 & 82 & 82.8 & 575.6 & 575.6 & 575.6 & 575.6 \\
 & best CRP after shift (oracle) & 0.00 & 1.00 & 76 & 75.3 & 2690.8 & 2690.4 & 2690.8 & 2690.4 \\
\midrule
$600$ & restarting UP & 0.01 & \textbf{1.00} & \textbf{81} & \textbf{82.3} & \textbf{2425.6} & \textbf{2425.6} & \textbf{2427.1} & \textbf{2427.1} \\
 & conditional CTM (ONS) & 0.01 & 1.00 & 98 & 98.3 & 1402.9 & 1403.0 & 1402.9 & 1403.0 \\
 & standard CTM & 0.00 & 1.00 & 104 & 104.6 & 575.6 & 575.6 & 575.6 & 575.6 \\
 & best CRP after shift (oracle) & 0.00 & 1.00 & 77 & 75.9 & 2434.4 & 2434.2 & 2434.4 & 2434.2 \\
\midrule
$4000$ & restarting UP & 0.00 & \textbf{1.00} & \textbf{88} & \textbf{87.1} & \textbf{251.2} & \textbf{251.0} & \textbf{252.9} & \textbf{252.7} \\
 & conditional CTM (ONS) & 0.00 & 1.00 & 179 & 178.7 & 86.8 & 86.9 & 86.8 & 86.9 \\
 & standard CTM & 0.00 & 1.00 & 197 & 196.2 & 50.1 & 50.0 & 50.1 & 50.0 \\
 & best CRP after shift (oracle) & 0.00 & 1.00 & 73 & 74.2 & 261.0 & 260.1 & 261.0 & 260.1 \\
\bottomrule
\end{tabular}
\caption{\textbf{Gradual drift, $\lambda = 1.5$.} $X_t \sim \mathcal{N}(1.5\,(t - t_0)/100, 1)$ after $t_0$. Reference set $n = 2000$, stream length $T = 4500$, 100 trials; details in Appendix~\ref{app:exp-synthetic}. \emph{False al.}: fraction rejecting at or before $t_0$; \emph{power}: fraction of the trials without a false alarm that reject by $T$, $\mathbb{P}(\tau\leq T\mid\tau>t_0)$. Delays are over the trials without a false alarm; the median counts misses as $\infty$, the mean counts them as $T-t_0+1$. The restarting portfolio rejects on its test martingale $M_t$; $\Wealth_T$ is its portfolio wealth, and $\Wealth_T = M_T$ for the others. \textbf{Bold}: best implementable strategy in the block.}
\label{tab:drift-1p5}
\end{table}

\begin{table}[htbp]
\centering
\small
\setlength{\tabcolsep}{3.5pt}
\begin{tabular}{rlrrrrrrrr}
\toprule
& & & & \multicolumn{2}{c}{delay after $t_0$} & \multicolumn{2}{c}{$\ln \Wealth_T$} & \multicolumn{2}{c}{$\ln M_T$} \\
\cmidrule(lr){5-6}\cmidrule(lr){7-8}\cmidrule(lr){9-10}
$t_0$ & strategy & false al. & power & median & mean & median & mean & median & mean \\
\midrule
$0$ & restarting UP & 0.00 & \textbf{1.00} & \textbf{41} & \textbf{40.8} & \textbf{2842.2} & \textbf{2842.4} & \textbf{2842.9} & \textbf{2843.0} \\
 & conditional CTM (ONS) & 0.00 & 1.00 & 43 & 42.3 & 1649.9 & 1649.9 & 1649.9 & 1649.9 \\
 & standard CTM & 0.00 & 1.00 & 47 & 47.0 & 575.6 & 575.6 & 575.6 & 575.6 \\
 & best CRP after shift (oracle) & 0.00 & 1.00 & 43 & 42.4 & 2848.6 & 2848.5 & 2848.6 & 2848.5 \\
\midrule
$200$ & restarting UP & 0.00 & \textbf{1.00} & 54 & \textbf{53.2} & \textbf{2709.5} & \textbf{2709.6} & \textbf{2710.7} & \textbf{2710.8} \\
 & conditional CTM (ONS) & 0.00 & 1.00 & \textbf{53} & 53.7 & 1573.1 & 1572.9 & 1573.1 & 1572.9 \\
 & standard CTM & 0.00 & 1.00 & 59 & 59.3 & 575.6 & 575.6 & 575.6 & 575.6 \\
 & best CRP after shift (oracle) & 0.00 & 1.00 & 43 & 42.5 & 2721.3 & 2720.7 & 2721.3 & 2720.7 \\
\midrule
$600$ & restarting UP & 0.01 & \textbf{1.00} & \textbf{56} & \textbf{55.8} & \textbf{2452.7} & \textbf{2452.5} & \textbf{2454.2} & \textbf{2453.9} \\
 & conditional CTM (ONS) & 0.01 & 1.00 & 70 & 70.5 & 1418.0 & 1417.9 & 1418.0 & 1417.9 \\
 & standard CTM & 0.00 & 1.00 & 77 & 76.9 & 575.6 & 575.6 & 575.6 & 575.6 \\
 & best CRP after shift (oracle) & 0.00 & 1.00 & 41 & 41.5 & 2464.9 & 2464.6 & 2464.9 & 2464.6 \\
\midrule
$4000$ & restarting UP & 0.00 & \textbf{1.00} & \textbf{60} & \textbf{59.9} & \textbf{277.3} & \textbf{277.2} & \textbf{278.8} & \textbf{278.8} \\
 & conditional CTM (ONS) & 0.00 & 1.00 & 142 & 142.3 & 101.2 & 101.2 & 101.2 & 101.2 \\
 & standard CTM & 0.00 & 1.00 & 161 & 160.4 & 58.6 & 58.5 & 58.6 & 58.5 \\
 & best CRP after shift (oracle) & 0.00 & 1.00 & 42 & 41.8 & 290.2 & 289.5 & 290.2 & 289.5 \\
\bottomrule
\end{tabular}
\caption{\textbf{Gradual drift, $\lambda = 3$.} $X_t \sim \mathcal{N}(3\,(t - t_0)/100, 1)$ after $t_0$. Reference set $n = 2000$, stream length $T = 4500$, 100 trials; details in Appendix~\ref{app:exp-synthetic}. \emph{False al.}: fraction rejecting at or before $t_0$; \emph{power}: fraction of the trials without a false alarm that reject by $T$, $\mathbb{P}(\tau\leq T\mid\tau>t_0)$. Delays are over the trials without a false alarm; the median counts misses as $\infty$, the mean counts them as $T-t_0+1$. The restarting portfolio rejects on its test martingale $M_t$; $\Wealth_T$ is its portfolio wealth, and $\Wealth_T = M_T$ for the others. \textbf{Bold}: best implementable strategy in the block.}
\label{tab:drift-3}
\end{table}

\begin{table}[htbp]
\centering
\small
\setlength{\tabcolsep}{3.5pt}
\begin{tabular}{rlrrrrrrrr}
\toprule
& & & & \multicolumn{2}{c}{delay after $t_0$} & \multicolumn{2}{c}{$\ln \Wealth_T$} & \multicolumn{2}{c}{$\ln M_T$} \\
\cmidrule(lr){5-6}\cmidrule(lr){7-8}\cmidrule(lr){9-10}
$t_0$ & strategy & false al. & power & median & mean & median & mean & median & mean \\
\midrule
$0$ & restarting UP & 0.00 & \textbf{1.00} & \textbf{31} & \textbf{31.1} & \textbf{2852.6} & \textbf{2852.9} & \textbf{2853.2} & \textbf{2853.4} \\
 & conditional CTM (ONS) & 0.00 & 1.00 & 33 & 33.0 & 1655.8 & 1655.7 & 1655.8 & 1655.7 \\
 & standard CTM & 0.00 & 1.00 & 38 & 37.8 & 575.6 & 575.6 & 575.6 & 575.6 \\
 & best CRP after shift (oracle) & 0.00 & 1.00 & 29 & 29.2 & 2860.5 & 2860.5 & 2860.5 & 2860.5 \\
\midrule
$200$ & restarting UP & 0.00 & \textbf{1.00} & \textbf{41} & \textbf{41.4} & \textbf{2720.3} & \textbf{2720.3} & \textbf{2721.6} & \textbf{2721.5} \\
 & conditional CTM (ONS) & 0.00 & 1.00 & 42 & 42.4 & 1578.8 & 1578.8 & 1578.8 & 1578.8 \\
 & standard CTM & 0.00 & 1.00 & 48 & 47.7 & 575.6 & 575.6 & 575.6 & 575.6 \\
 & best CRP after shift (oracle) & 0.00 & 1.00 & 28 & 28.7 & 2733.2 & 2732.8 & 2733.2 & 2732.8 \\
\midrule
$600$ & restarting UP & 0.01 & \textbf{1.00} & \textbf{44} & \textbf{43.8} & \textbf{2463.0} & \textbf{2463.0} & \textbf{2464.4} & \textbf{2464.5} \\
 & conditional CTM (ONS) & 0.01 & 1.00 & 57 & 57.4 & 1423.7 & 1423.7 & 1423.7 & 1423.7 \\
 & standard CTM & 0.00 & 1.00 & 64 & 63.6 & 575.6 & 575.6 & 575.6 & 575.6 \\
 & best CRP after shift (oracle) & 0.00 & 1.00 & 29 & 29.0 & 2476.9 & 2476.7 & 2476.9 & 2476.7 \\
\midrule
$4000$ & restarting UP & 0.00 & \textbf{1.00} & \textbf{47} & \textbf{47.3} & \textbf{287.8} & \textbf{287.6} & \textbf{289.4} & \textbf{289.2} \\
 & conditional CTM (ONS) & 0.00 & 1.00 & 126 & 126.7 & 107.1 & 107.1 & 107.1 & 107.1 \\
 & standard CTM & 0.00 & 1.00 & 146 & 145.3 & 62.1 & 62.1 & 62.1 & 62.1 \\
 & best CRP after shift (oracle) & 0.00 & 1.00 & 28 & 29.0 & 301.8 & 301.4 & 301.8 & 301.4 \\
\bottomrule
\end{tabular}
\caption{\textbf{Gradual drift, $\lambda = 5$.} $X_t \sim \mathcal{N}(5\,(t - t_0)/100, 1)$ after $t_0$. Reference set $n = 2000$, stream length $T = 4500$, 100 trials; details in Appendix~\ref{app:exp-synthetic}. \emph{False al.}: fraction rejecting at or before $t_0$; \emph{power}: fraction of the trials without a false alarm that reject by $T$, $\mathbb{P}(\tau\leq T\mid\tau>t_0)$. Delays are over the trials without a false alarm; the median counts misses as $\infty$, the mean counts them as $T-t_0+1$. The restarting portfolio rejects on its test martingale $M_t$; $\Wealth_T$ is its portfolio wealth, and $\Wealth_T = M_T$ for the others. \textbf{Bold}: best implementable strategy in the block.}
\label{tab:drift-5}
\end{table}

\begin{table}[htbp]
\centering
\footnotesize
\begin{tabular}{lrrrr}
\toprule
 & \multicolumn{2}{c}{$n = 1000$} & \multicolumn{2}{c}{$n = 2000$} \\
\cmidrule(lr){2-3}\cmidrule(lr){4-5}
strategy & rejected & median final log-wealth & rejected & median final log-wealth \\
\midrule
restarting UP & 0.00 & -0.88 & 0.00 & -0.97 \\
conditional CTM (ONS) & 0.00 & -2.56 & 0.00 & -2.75 \\
standard CTM & 0.02 & -0.68 & 0.02 & -0.76 \\
\bottomrule
\end{tabular}
\caption{\textbf{Type-I error.} Fraction of 100 null streams ($X_t \sim \mathcal{N}(0,1)$ throughout, $T = 4500$) whose wealth ever reaches $1/\alpha = 20$ ($M_t$ for the restarting portfolio, the martingale itself for the CTMs), and the median final log-wealth, for the reference sizes of the sharp change point ($n = 1000$) and the drift ($n = 2000$).}
\label{tab:type_i}
\end{table}

\clearpage
\subsection{ImageNet-C}
\label{app:exp-imagenet}

\paragraph{Data.}
We repeat the ImageNet-C experiment of~\citet[Figure~4]{shaer2026testing} with a clean phase of length $t_0$ inserted before the corrupted stream, so that their experiment is the slice $t_0 = 0$.
The scores are the predictive entropies of the ViT-B/16 classifier~\citep{dosovitskiy2021image} of the \texttt{timm} library~\citep{wightman2019timm} (weights \texttt{augreg2}) on the ImageNet validation images~\citep{deng2009imagenet}.
The $50{,}000$ images are split into a clean holdout of $12{,}500$ images and $37{,}500$ images whose severity-$5$ ImageNet-C versions~\citep{hendrycks2019benchmarking} form the shifted stream, for each of the $15$ corruptions.
As in the authors' code, the seed shuffles the holdout and the corrupted images.
The reference dataset is the first $n$ images of the holdout, and the clean phase consists of the next $t_0$ images, which are never in the reference dataset.

\paragraph{Grid And Hyperparameters.}
We keep the grid of~\citet{shaer2026testing}, $n \in \{100, 250,\allowbreak 500, 1000, 4000\}$, the seeds $0, \ldots, 9$, and all $15$ corruptions, and we vary $t_0 \in \{0, 125, 250, 500, 1000,\allowbreak 2000, 4000, 8000\}$, a geometric grid on which a dependence of order $\ln t_0$ appears as a straight line.
The CTMs keep the ImageNet hyperparameters of~\citet{shaer2026testing}: $\alpha = 0.05$, band level $\delta = \alpha$, bets in $[-1/2, 1/2]$, $C = 0.05$, $k = 10^{-6}$, and a warm-up of $50$ rounds; the released code skips the first $51$ observations for the CCTM and $50$ for the standard CTM.
The restarting portfolio uses the same band and no warm-up.
The oracle holds cash until $t_0$ and then makes the best constant bet in hindsight.
Every method runs for $H = 2000$ rounds after the shift, because the direct method costs $\sO((t_0 + H)^3)$; at $t_0 = 0$, almost all runs detect the shift well within $H$.
The restarting portfolio processes the clean phase once for each $(n, \text{seed}, t_0)$ and copies its state for the $15$ corruptions, and the whole grid takes about $40$ minutes on seven cores.

\paragraph{Delay.}
The delay is $\tau - t_0$ for every method, and the warm-up of a CTM is not subtracted.
For $t_0 \geq 125$, the warm-up lies in the clean phase and costs nothing.
At $t_0 = 0$, it falls on shifted data, so the CTMs discard the first $51$ or $50$ corrupted images, and these rounds count towards their delay; \citet{shaer2026testing} instead measure time from the end of the warm-up.
Medians count missed detections, i.e., no rejection within $H$ rounds, as $\infty$, and they are taken over the runs without a false alarm.
Means count a missed detection as $H + 1$ instead of dropping it, and like the medians, they are taken over the runs of each method without a false alarm; where some runs miss, the mean is a lower bound.

\begin{table*}[t]
    \centering
    \footnotesize
    \caption{
        Median detection delay $\tau - t_0$ on ImageNet-C with $n = 1000$, by corruption group, over $10$ seeds and the corruptions of each group.
        The delays of the CTMs include their warm-up of $50$ rounds.
        \textbf{Bold}: the fastest implementable method.
        Table~\ref{tab:imagenet-n1000} gives the false-alarm rates, the power, and the mean delays, and Tables~\ref{tab:imagenet-n100}--\ref{tab:imagenet-n4000} give the other reference sizes.
    }
    \label{tab:exp-imagenet}
    \setlength{\tabcolsep}{2.8pt}
    \begin{tabular}{@{}l rrrrr rrrrr rrrrr@{}}
        \toprule
        & \multicolumn{5}{c}{$t_0 = 0$} & \multicolumn{5}{c}{$t_0 = 1000$} & \multicolumn{5}{c}{$t_0 = 8000$} \\
        \cmidrule(lr){2-6} \cmidrule(lr){7-11} \cmidrule(lr){12-16}
        & All & Noise & Blur & Weather & Digital & All & Noise & Blur & Weather & Digital & All & Noise & Blur & Weather & Digital \\
        \midrule
        Ours & \textbf{17} & \textbf{18} & \textbf{16.5} & \textbf{12} & \textbf{27.5} & \textbf{42} & \textbf{41.5} & \textbf{38} & \textbf{27.5} & \textbf{88.5} & \textbf{50.5} & \textbf{54.5} & \textbf{45} & \textbf{31} & \textbf{130.5} \\
        CCTM & 67 & 69 & 65.5 & 61 & 77.5 & 84.5 & 85 & 74.5 & 59.5 & 143 & 244.5 & 250 & 220.5 & 165 & 425.5 \\
        CTM & 75 & 76 & 72.5 & 66 & 85 & 78 & 82 & 74 & 58 & 106.5 & 264 & 270 & 241 & 192.5 & 390.5 \\
        \midrule
        Oracle & 8 & 7 & 8.5 & 6 & 19 & 8 & 7 & 8.5 & 6 & 19 & 8 & 7 & 8.5 & 6 & 19 \\
        \bottomrule
    \end{tabular}
\end{table*}

\begin{figure}[p]
    \centering
    \includegraphics[width=\textwidth]{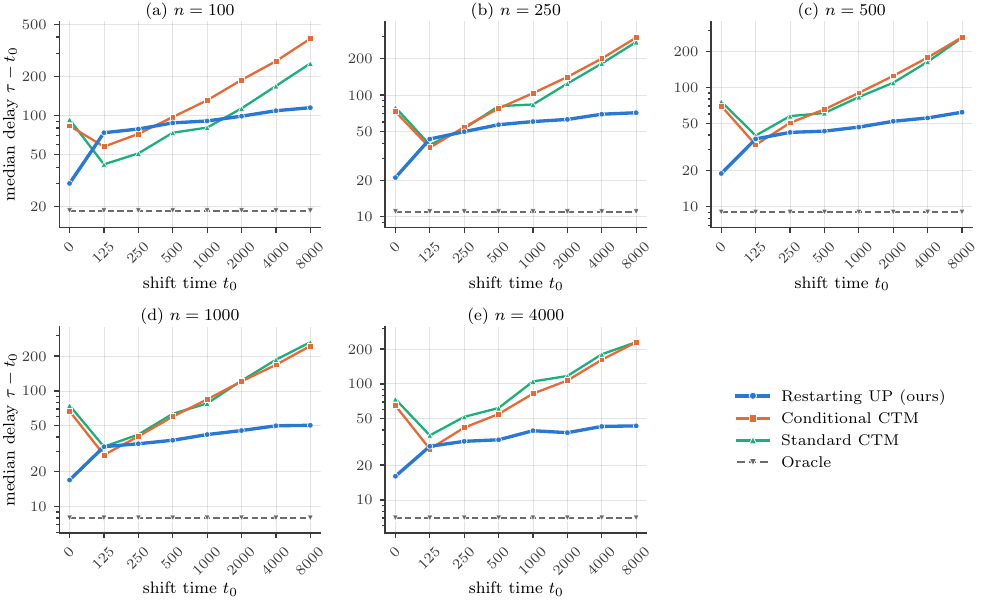}
    \caption{\textbf{ImageNet-C: median delay $\tau - t_0$ against the length $t_0$ of the clean phase}, for every reference size $n$, over $15$ corruptions and $10$ seeds. Runs that miss the shift within $2000$ rounds count as infinite delays. The vertical axes are logarithmic, and the horizontal axes are logarithmic beyond $t_0 = 0$.}
    \label{fig:app-imagenet-delay}
\end{figure}

\begin{figure}[p]
    \centering
    \includegraphics[width=\textwidth]{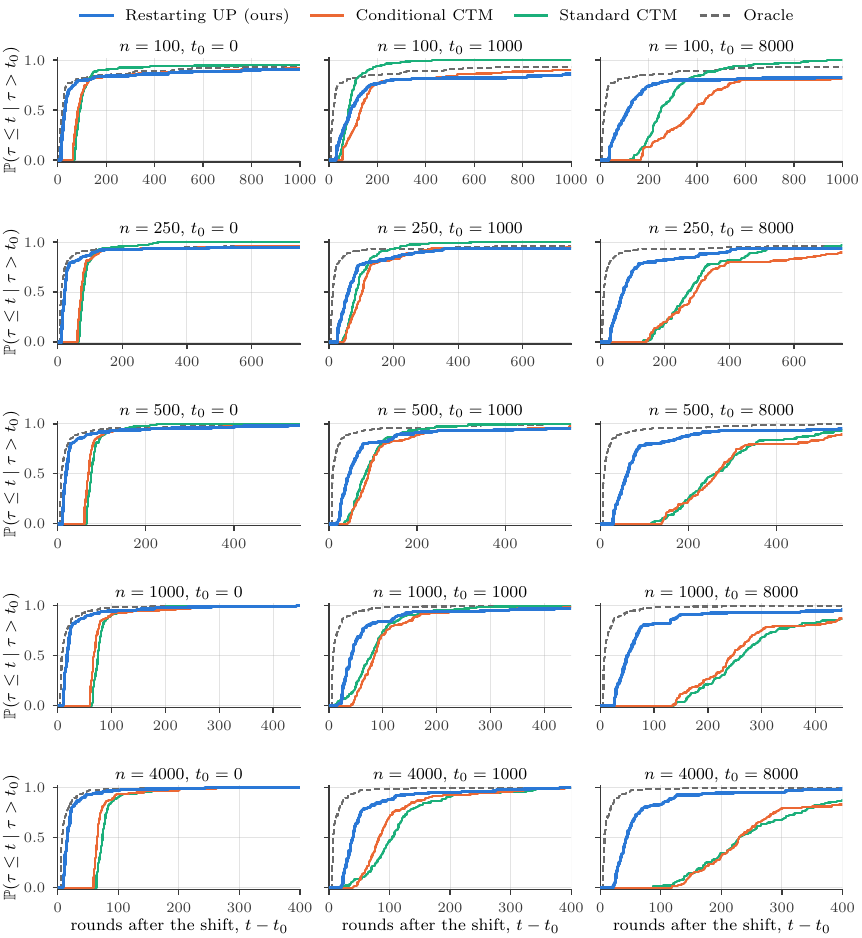}
    \caption{\textbf{ImageNet-C: power given no false alarm}, $\Prob\{\tau \leq t \mid \tau > t_0\}$, against the number of rounds after the shift, for every reference size $n$ (rows) and $t_0 \in \{0, 1000, 8000\}$ (columns), over $15$ corruptions and $10$ seeds.}
    \label{fig:app-imagenet-power}
\end{figure}

\paragraph{Results.}
Figures~\ref{fig:app-imagenet-delay} and~\ref{fig:app-imagenet-power} show the delay against $t_0$ and the power for each $n$, pooled over corruptions and seeds, and Tables~\ref{tab:imagenet-n100}--\ref{tab:imagenet-n4000} break the delay down by corruption group; Table~\ref{tab:exp-imagenet} summarizes them for $n = 1000$.
\begin{itemize}
    \item \emph{At $t_0 = 0$}, the restarting portfolio is the fastest method for every $n$, with $16$ to $30$ rounds against $65$ to $83$ for the CCTM and $74$ to $93$ for the standard CTM. Most of this gap is the warm-up of the CTMs; counted from the end of the warm-up, the restarting portfolio and the CCTM are within two rounds of each other.
    \item \emph{The delay of the restarting portfolio grows like $\ln t_0$.} From $t_0 = 125$ to $t_0 = 8000$, its median delay rises from $33$ to $50.5$ rounds at $n = 1000$, and regressed on $\ln t_0$ it gains $3.6$ (at $n = 4000$) to $10$ (at $n = 100$) additional rounds whenever $t_0$ increases by a factor of $e$. The slope is larger for small $n$, because a wider band makes each nat of restart cost take more rounds to earn.
    \item \emph{The delays of both CTMs grow roughly linearly with $t_0$}, by $24$ to $41$ rounds per $1000$ clean rounds. At $t_0 = 8000$, the CCTM needs $3.4$ ($n = 100$) to $5.3$ ($n = 4000$) times as many rounds as the restarting portfolio.
\end{itemize}
The CTMs are ahead only for short clean phases.
At $t_0 = 125$, one of them is faster for every $n$, e.g., the CCTM with $28$ against $33$ rounds at $n = 1000$.
For $n \geq 250$, the restarting portfolio has the shortest median delay from $t_0 = 250$ on.
For $n = 100$, the standard CTM stays ahead up to $t_0 = 1000$, because its growing pool of clean scores sharpens its $p$-values, and the restarting portfolio leads from $t_0 = 2000$ on.
Among the corruption groups, the Digital corruptions are the hardest.
For brightness at $n \leq 250$, both conditional tests often miss the shift within $H$, because the band width $\epsilon \approx 0.14$ at $n = 100$ exceeds the signal; the standard CTM pays no band and detects it once its pool has grown.
Figure~\ref{fig:exp-groups} in Section~\ref{sec:experiments} summarizes the comparison of the two conditional tests as the ratio of their delays.
At $t_0 = 0$, the ratio is $3.1$ to $5.3$ for Noise, Blur, and Weather and $1.2$ to $3.3$ for Digital, dominated by the warm-up of the CCTM; at $t_0 = 1000$, where the warm-up is free, it is $1.4$ to $2.2$ ($1.0$ to $1.9$ for Digital); and at $t_0 = 8000$, it grows to $3.3$ to $5.7$ ($1.4$ to $4.2$ for Digital).
For $n = 4000$, the standard CTM raises a false alarm at round $392$ of the clean phase for one of the $10$ seeds, and since the $15$ corruptions share the clean phase of a seed, this counts as a false-alarm rate of $0.10$ at every $t_0 \geq 500$, including $t_0 \in \{1000, 8000\}$ in Table~\ref{tab:imagenet-n4000}.

\paragraph{Null Check.}
As in the reproduction, the null stream is the rest of the clean holdout, i.e., $12{,}500 - n$ images, with $100$ seeds for each $n \in \{100, 500, 4000\}$ (Table~\ref{tab:imagenet-null}).
The test martingale of the restarting portfolio reaches $1/\alpha$ on one of the $300$ streams (at $n = 4000$), and its portfolio wealth on none.
The CCTM never rejects, and the standard CTM rejects on $1\%$ to $3\%$ of the streams.
All rates are below $\alpha$.
After $8500$ to $12{,}400$ clean rounds, the median log-wealth of the restarting portfolio is between $-1.1$ and $-0.6$.

\begin{table}[htbp]
\centering
\small
\setlength{\tabcolsep}{3pt}
\begin{tabular}{rlrrrrrrrr}
\toprule
& & & & \multicolumn{5}{c}{median delay} & mean \\
\cmidrule(lr){5-9}
$t_0$ & strategy & false al. & power & all & noise & blur & weather & digital & delay \\
\midrule
$0$ & restarting UP & 0.00 & 0.93 & \textbf{30} & \textbf{30} & \textbf{26.5} & \textbf{16.5} & 177.5 & 235.5$^{\dagger}$ \\
 & conditional CTM (ONS) & 0.00 & 0.93 & 83 & 89 & 78 & 65.5 & 203 & 269.6$^{\dagger}$ \\
 & standard CTM & 0.00 & 0.96 & 93 & 97 & 88 & 75.5 & \textbf{126.5} & \textbf{194.2$^{\dagger}$} \\
 & best CRP after shift (oracle) & 0.00 & 0.93 & 18.5 & 18.5 & 16 & 9 & 136.5 & 189.4$^{\dagger}$ \\
\midrule
$1000$ & restarting UP & 0.00 & 0.92 & 90.5 & 99.5 & \textbf{66.5} & \textbf{40.5} & 818 & 360.5$^{\dagger}$ \\
 & conditional CTM (ONS) & 0.00 & 0.93 & 130.5 & 133 & 113.5 & 72 & 480 & 327.3$^{\dagger}$ \\
 & standard CTM & 0.00 & 1.00 & \textbf{80.5} & \textbf{83.5} & 76 & 60.5 & \textbf{114} & \textbf{96.8} \\
 & best CRP after shift (oracle) & 0.00 & 0.93 & 18.5 & 18.5 & 16 & 9 & 136.5 & 189.4$^{\dagger}$ \\
\midrule
$8000$ & restarting UP & 0.00 & 0.92 & \textbf{114.5} & \textbf{119} & \textbf{84} & \textbf{47} & 1064.5 & 397.6$^{\dagger}$ \\
 & conditional CTM (ONS) & 0.00 & 0.91 & 388 & 398.5 & 318.5 & 215.5 & 1364 & 621.1$^{\dagger}$ \\
 & standard CTM & 0.00 & 1.00 & 250.5 & 265.5 & 240.5 & 191 & \textbf{383.5} & \textbf{296.6} \\
 & best CRP after shift (oracle) & 0.00 & 0.93 & 18.5 & 18.5 & 16 & 9 & 136.5 & 189.4$^{\dagger}$ \\
\bottomrule
\end{tabular}
\caption{\textbf{ImageNet-C, $n = 100$.} 15 severity-5 corruptions $\times$ 10 seeds after $t_0$ clean images; every method runs $H = 2000$ rounds past the change. \emph{False al.}: rejection at or before $t_0$; \emph{power}: $\mathbb{P}(\tau\leq t_0+H\mid\tau>t_0)$. Delays $\tau-t_0$ (warm-up rounds included), medians over the trials without a false alarm, misses counted as $\infty$; the mean counts misses as $H+1$ ($^{\dagger}$: some runs missed, so the mean is a lower bound). The restarting portfolio rejects on $M_t$. Groups as in~\citet{shaer2026testing}, Figure~4. \textbf{Bold}: best implementable strategy.}
\label{tab:imagenet-n100}
\end{table}

\begin{table}[htbp]
\centering
\small
\setlength{\tabcolsep}{3pt}
\begin{tabular}{rlrrrrrrrr}
\toprule
& & & & \multicolumn{5}{c}{median delay} & mean \\
\cmidrule(lr){5-9}
$t_0$ & strategy & false al. & power & all & noise & blur & weather & digital & delay \\
\midrule
$0$ & restarting UP & 0.00 & 0.98 & \textbf{21} & \textbf{21} & \textbf{19} & \textbf{14} & \textbf{67} & 118.0$^{\dagger}$ \\
 & conditional CTM (ONS) & 0.00 & 0.99 & 73 & 75 & 68.5 & 63 & 102 & 151.9$^{\dagger}$ \\
 & standard CTM & 0.00 & 1.00 & 79 & 84 & 75.5 & 69 & 102.5 & \textbf{92.7} \\
 & best CRP after shift (oracle) & 0.00 & 0.99 & 11 & 11 & 11 & 7 & 29 & 78.5$^{\dagger}$ \\
\midrule
$1000$ & restarting UP & 0.00 & 0.96 & \textbf{60.5} & \textbf{65.5} & \textbf{49} & \textbf{31.5} & 185 & 194.2$^{\dagger}$ \\
 & conditional CTM (ONS) & 0.00 & 0.99 & 103.5 & 105 & 88 & 63 & 213.5 & 197.1$^{\dagger}$ \\
 & standard CTM & 0.00 & 1.00 & 83.5 & 82.5 & 75.5 & 63 & \textbf{125.5} & \textbf{101.5} \\
 & best CRP after shift (oracle) & 0.00 & 0.99 & 11 & 11 & 11 & 7 & 29 & 78.5$^{\dagger}$ \\
\midrule
$8000$ & restarting UP & 0.00 & 0.95 & \textbf{71.5} & \textbf{78} & \textbf{59.5} & \textbf{37.5} & \textbf{238.5} & \textbf{218.8$^{\dagger}$} \\
 & conditional CTM (ONS) & 0.00 & 0.94 & 295.5 & 301.5 & 247.5 & 184.5 & 612.5 & 432.2$^{\dagger}$ \\
 & standard CTM & 0.00 & 1.00 & 272 & 283.5 & 249 & 196.5 & 438.5 & 313.0 \\
 & best CRP after shift (oracle) & 0.00 & 0.99 & 11 & 11 & 11 & 7 & 29 & 78.5$^{\dagger}$ \\
\bottomrule
\end{tabular}
\caption{\textbf{ImageNet-C, $n = 250$.} 15 severity-5 corruptions $\times$ 10 seeds after $t_0$ clean images; every method runs $H = 2000$ rounds past the change. \emph{False al.}: rejection at or before $t_0$; \emph{power}: $\mathbb{P}(\tau\leq t_0+H\mid\tau>t_0)$. Delays $\tau-t_0$ (warm-up rounds included), medians over the trials without a false alarm, misses counted as $\infty$; the mean counts misses as $H+1$ ($^{\dagger}$: some runs missed, so the mean is a lower bound). The restarting portfolio rejects on $M_t$. Groups as in~\citet{shaer2026testing}, Figure~4. \textbf{Bold}: best implementable strategy.}
\label{tab:imagenet-n250}
\end{table}

\begin{table}[htbp]
\centering
\small
\setlength{\tabcolsep}{3pt}
\begin{tabular}{rlrrrrrrrr}
\toprule
& & & & \multicolumn{5}{c}{median delay} & mean \\
\cmidrule(lr){5-9}
$t_0$ & strategy & false al. & power & all & noise & blur & weather & digital & delay \\
\midrule
$0$ & restarting UP & 0.00 & 1.00 & \textbf{19} & \textbf{20} & \textbf{17.5} & \textbf{13} & \textbf{29.5} & \textbf{48.3} \\
 & conditional CTM (ONS) & 0.00 & 1.00 & 70 & 71 & 67 & 61.5 & 82 & 91.5 \\
 & standard CTM & 0.00 & 1.00 & 76 & 78.5 & 74 & 68 & 93 & 84.3 \\
 & best CRP after shift (oracle) & 0.00 & 1.00 & 9 & 8 & 9 & 6.5 & 20 & 29.3 \\
\midrule
$1000$ & restarting UP & 0.00 & 1.00 & \textbf{46.5} & \textbf{44} & \textbf{41.5} & \textbf{29} & 134.5 & 106.9 \\
 & conditional CTM (ONS) & 0.00 & 1.00 & 90 & 91 & 80.5 & 59.5 & 163 & 128.0 \\
 & standard CTM & 0.00 & 1.00 & 83 & 85 & 78 & 64.5 & \textbf{120.5} & \textbf{97.1} \\
 & best CRP after shift (oracle) & 0.00 & 1.00 & 9 & 8 & 9 & 6.5 & 20 & 29.3 \\
\midrule
$8000$ & restarting UP & 0.00 & 1.00 & \textbf{62} & \textbf{65} & \textbf{51.5} & \textbf{35} & \textbf{161} & \textbf{129.2} \\
 & conditional CTM (ONS) & 0.00 & 1.00 & 263.5 & 271 & 226.5 & 176 & 488 & 347.2 \\
 & standard CTM & 0.00 & 1.00 & 261 & 287 & 241 & 193 & 355.5 & 297.0 \\
 & best CRP after shift (oracle) & 0.00 & 1.00 & 9 & 8 & 9 & 6.5 & 20 & 29.3 \\
\bottomrule
\end{tabular}
\caption{\textbf{ImageNet-C, $n = 500$.} 15 severity-5 corruptions $\times$ 10 seeds after $t_0$ clean images; every method runs $H = 2000$ rounds past the change. \emph{False al.}: rejection at or before $t_0$; \emph{power}: $\mathbb{P}(\tau\leq t_0+H\mid\tau>t_0)$. Delays $\tau-t_0$ (warm-up rounds included), medians over the trials without a false alarm, misses counted as $\infty$; the mean counts misses as $H+1$. The restarting portfolio rejects on $M_t$. Groups as in~\citet{shaer2026testing}, Figure~4. \textbf{Bold}: best implementable strategy.}
\label{tab:imagenet-n500}
\end{table}

\begin{table}[htbp]
\centering
\small
\setlength{\tabcolsep}{3pt}
\begin{tabular}{rlrrrrrrrr}
\toprule
& & & & \multicolumn{5}{c}{median delay} & mean \\
\cmidrule(lr){5-9}
$t_0$ & strategy & false al. & power & all & noise & blur & weather & digital & delay \\
\midrule
$0$ & restarting UP & 0.00 & 1.00 & \textbf{17} & \textbf{18} & \textbf{16.5} & \textbf{12} & \textbf{27.5} & \textbf{31.8} \\
 & conditional CTM (ONS) & 0.00 & 1.00 & 67 & 69 & 65.5 & 61 & 77.5 & 80.3 \\
 & standard CTM & 0.00 & 1.00 & 75 & 76 & 72.5 & 66 & 85 & 81.3 \\
 & best CRP after shift (oracle) & 0.00 & 1.00 & 8 & 7 & 8.5 & 6 & 19 & 17.3 \\
\midrule
$1000$ & restarting UP & 0.00 & 1.00 & \textbf{42} & \textbf{41.5} & \textbf{38} & \textbf{27.5} & \textbf{88.5} & \textbf{74.1} \\
 & conditional CTM (ONS) & 0.00 & 1.00 & 84.5 & 85 & 74.5 & 59.5 & 143 & 106.1 \\
 & standard CTM & 0.00 & 1.00 & 78 & 82 & 74 & 58 & 106.5 & 87.9 \\
 & best CRP after shift (oracle) & 0.00 & 1.00 & 8 & 7 & 8.5 & 6 & 19 & 17.3 \\
\midrule
$8000$ & restarting UP & 0.00 & 1.00 & \textbf{50.5} & \textbf{54.5} & \textbf{45} & \textbf{31} & \textbf{130.5} & \textbf{91.4} \\
 & conditional CTM (ONS) & 0.00 & 1.00 & 244.5 & 250 & 220.5 & 165 & 425.5 & 301.9 \\
 & standard CTM & 0.00 & 1.00 & 264 & 270 & 241 & 192.5 & 390.5 & 301.5 \\
 & best CRP after shift (oracle) & 0.00 & 1.00 & 8 & 7 & 8.5 & 6 & 19 & 17.3 \\
\bottomrule
\end{tabular}
\caption{\textbf{ImageNet-C, $n = 1000$.} 15 severity-5 corruptions $\times$ 10 seeds after $t_0$ clean images; every method runs $H = 2000$ rounds past the change. \emph{False al.}: rejection at or before $t_0$; \emph{power}: $\mathbb{P}(\tau\leq t_0+H\mid\tau>t_0)$. Delays $\tau-t_0$ (warm-up rounds included), medians over the trials without a false alarm, misses counted as $\infty$; the mean counts misses as $H+1$. The restarting portfolio rejects on $M_t$. Groups as in~\citet{shaer2026testing}, Figure~4. \textbf{Bold}: best implementable strategy.}
\label{tab:imagenet-n1000}
\end{table}

\begin{table}[htbp]
\centering
\small
\setlength{\tabcolsep}{3pt}
\begin{tabular}{rlrrrrrrrr}
\toprule
& & & & \multicolumn{5}{c}{median delay} & mean \\
\cmidrule(lr){5-9}
$t_0$ & strategy & false al. & power & all & noise & blur & weather & digital & delay \\
\midrule
$0$ & restarting UP & 0.00 & 1.00 & \textbf{16} & \textbf{16} & \textbf{16} & \textbf{12} & \textbf{25} & \textbf{23.9} \\
 & conditional CTM (ONS) & 0.00 & 1.00 & 65 & 66 & 64 & 60.5 & 73.5 & 73.8 \\
 & standard CTM & 0.00 & 1.00 & 74 & 76 & 71.5 & 66 & 83.5 & 79.9 \\
 & best CRP after shift (oracle) & 0.00 & 1.00 & 7 & 7 & 7 & 6 & 17.5 & 13.9 \\
\midrule
$1000$ & restarting UP & 0.00 & 1.00 & \textbf{39.5} & \textbf{41.5} & \textbf{36} & \textbf{26} & \textbf{73.5} & \textbf{58.0} \\
 & conditional CTM (ONS) & 0.00 & 1.00 & 82.5 & 85.5 & 74.5 & 59.5 & 133 & 101.7 \\
 & standard CTM & 0.10 & 1.00 & 105 & 109 & 94 & 76 & 155 & 116.6 \\
 & best CRP after shift (oracle) & 0.00 & 1.00 & 7 & 7 & 7 & 6 & 17.5 & 13.9 \\
\midrule
$8000$ & restarting UP & 0.00 & 1.00 & \textbf{43.5} & \textbf{47} & \textbf{38} & \textbf{28.5} & \textbf{99.5} & \textbf{67.8} \\
 & conditional CTM (ONS) & 0.00 & 1.00 & 229 & 237 & 209.5 & 165.5 & 398.5 & 274.7 \\
 & standard CTM & 0.10 & 1.00 & 229 & 249 & 217 & 181 & 342 & 273.7 \\
 & best CRP after shift (oracle) & 0.00 & 1.00 & 7 & 7 & 7 & 6 & 17.5 & 13.9 \\
\bottomrule
\end{tabular}
\caption{\textbf{ImageNet-C, $n = 4000$.} 15 severity-5 corruptions $\times$ 10 seeds after $t_0$ clean images; every method runs $H = 2000$ rounds past the change. \emph{False al.}: rejection at or before $t_0$; \emph{power}: $\mathbb{P}(\tau\leq t_0+H\mid\tau>t_0)$. Delays $\tau-t_0$ (warm-up rounds included), medians over the trials without a false alarm, misses counted as $\infty$; the mean counts misses as $H+1$. The restarting portfolio rejects on $M_t$. Groups as in~\citet{shaer2026testing}, Figure~4. \textbf{Bold}: best implementable strategy.}
\label{tab:imagenet-n4000}
\end{table}

\begin{table}[htbp]
\centering
\small
\begin{tabular}{rrrrrrr}
\toprule
 & \multicolumn{2}{c}{restarting UP} & \multicolumn{2}{c}{conditional CTM} & \multicolumn{2}{c}{standard CTM} \\
\cmidrule(lr){2-3}\cmidrule(lr){4-5}\cmidrule(lr){6-7}
$n$ & rejected & med.\ final & rejected & med.\ final & rejected & med.\ final \\
\midrule
$100$ & 0.00 & -0.56 & 0.00 & -0.89 & 0.03 & -0.80 \\
$500$ & 0.00 & -0.79 & 0.00 & -1.54 & 0.02 & -0.94 \\
$4000$ & 0.01 & -1.07 & 0.00 & -2.33 & 0.01 & -0.86 \\
\bottomrule
\end{tabular}
\caption{\textbf{ImageNet null check.} The stream is the rest of the clean holdout ($T = 12400$ for $n = 100$, $T = 12000$ for $n = 500$, $T = 8500$ for $n = 4000$), 100 seeds per $n$: fraction of streams whose wealth ever reaches $1/\alpha$ ($M_t$ for the restarting portfolio, the martingale itself for the CTMs), and the median final log-wealth.}
\label{tab:imagenet-null}
\end{table}

\end{document}